\documentclass[12pt]{article}

\usepackage{booktabs}       % professional-quality tables
\usepackage{nicefrac}       % compact symbols for 1/2, etc.
\usepackage{microtype}      % microtypography
\usepackage{smile}
\usepackage{algorithm}
\usepackage{algorithmic}
\usepackage{lipsum}
\usepackage{ragged2e}
\usepackage{geometry}
\usepackage{setspace}   
\usepackage[symbol]{footmisc}
\renewcommand*{\thefootnote}{\fnsymbol{footnote}}

\usepackage[title,toc]{appendix}
\usepackage{minitoc}
\noptcrule
\usepackage{natbib}
\usepackage[utf8]{inputenc} % allow utf-8 input
\usepackage[T1]{fontenc}    % use 8-bit T1 fonts
\usepackage{hyperref}   
\hypersetup{
	colorlinks=true,       % false: boxed links; true: colored links
	linkcolor=blue,        % color of internal links
	citecolor=blue,        % color of links to bibliography
	filecolor=magenta,     % color of file links
	urlcolor=blue
}% hyperlinks
\usepackage{url}            % simple URL typesetting
\usepackage{booktabs}       % professional-quality tables
\usepackage{amsfonts}       % blackboard math symbols
\usepackage{nicefrac}       % compact symbols for 1/2, etc.
\usepackage{microtype}      % microtypography
\usepackage[table,xcdraw]{xcolor}
\usepackage{smile}
\usepackage{algorithm}
\usepackage{algorithmic}
\usepackage[]{graphicx,float,latexsym}
\usepackage{amsfonts,amstext,amsmath,amssymb,amsthm}
\usepackage{wrapfig}
\usepackage{lipsum}
\usepackage{gensymb}

\usepackage{xcolor}

\newcommand{\myt}{\texttt{t}}
\newcommand{\mys}{\texttt{s}}
\def\T{{ \mathrm{\scriptscriptstyle T} }}

\def\mytitle{Transfer Learning for Causal Effect Estimation}

\newcommand{\independent}{\perp\!\!\!\!\perp} 

\title{\mytitle}

\author{\quad \\Song Wei\textsuperscript{$\dagger$}, \ Hanyu Zhang\textsuperscript{$\dagger$}, \ Ronald Moore\textsuperscript{$\ddagger$},\\Rishikesan Kamaleswaran\textsuperscript{$\ddagger$$\dagger$}, \ Yao Xie\textsuperscript{$\dagger$}\footnote{Author correspondence to Yao Xie (e-mail: \texttt{yao.xie@isye.gatech.edu}). This work is partially supported by an NSF CAREER CCF-1650913, NSF DMS-2134037, CMMI-2015787, CMMI-2112533, DMS-1938106, DMS-1830210, the Coca-Cola Foundation, and an Emory Hospital grant. The implementation using benchmark dataset is available online at \url{https://github.com/SongWei-GT/L1-TCL}.} \\
  \small{\textsuperscript{$\dagger$}Georgia Institute of Technology, \quad \textsuperscript{$\ddagger$}Emory University.} \\ \quad
}
\date{\vspace{-20pt}}

\begin{document}

\maketitle

\begin{abstract}
We present a Transfer Causal Learning (TCL) framework when target and source domains share the same covariate/feature spaces, aiming to improve causal effect estimation accuracy in limited data. Limited data is very common in medical applications, where some rare medical conditions, such as sepsis, are of interest. Our proposed method, named \texttt{$\ell_1$-TCL}, incorporates $\ell_1$ regularized TL for nuisance models (e.g., propensity score model); the TL estimator of the nuisance parameters is plugged into downstream average causal/treatment effect estimators (e.g., inverse probability weighted estimator). We establish non-asymptotic recovery guarantees for the \texttt{$\ell_1$-TCL} with generalized linear model (GLM) under the sparsity assumption in the high-dimensional setting, and demonstrate the empirical benefits of \texttt{$\ell_1$-TCL} through extensive numerical simulation for GLM and recent neural network nuisance models. Our method is subsequently extended to real data and generates meaningful insights consistent with medical literature, a case where all baseline methods fail. 
\end{abstract}

% key words: healthcare application, causal inference, average causal effect, transfer learning, generalized linear models, neural networks

\doparttoc % Tell to minitoc to generate a toc for the parts
\faketableofcontents % Run a fake tableofcontents command for the partocs

\part{} % Start the document part
%\parttoc % Insert the document TOC

%\tableofcontents*

\renewcommand*{\thefootnote}{\arabic{footnote}}

\vspace{-0.45in}

\section{Introduction}

Causal effect estimation from observational data has attracted much attention in many fields since it is crucial for informed decision-making and effective intervention design. Several unbiased estimators for the Average Causal Effect (ACE) have been proposed, e.g., the inverse probability weighted (IPW) estimator, outcome regression (OR) estimator, and doubly robust (DR) estimator, which have shown good empirical performances and strong theoretical guarantees; see, e.g., \citet{yao2021survey}, for a survey of those estimators.
However, there is less study for the situations with limited data present in the target study. In modern applications, advanced data acquisition techniques make it possible to collect datasets from other domains, referred to as the source domains, that are related to (but different from) that of the target study. Transfer Learning (TL), aiming to boost performance in the target domain with knowledge gained from the source domain, has shown promise in practice \citep{torrey2010transfer}.

\subsection{Motivating application and dataset}\label{sec:motivation}

Our work is motivated by a real study on the treatment effect (or rather, ACE) of vasopressor therapy for septic patients; we handle a dataset containing
in-hospital Electronic Medical Records (EMRs) derived from an academic level 1 trauma center in 2018 (i.e., the target domain). The resulting summary statistics of the \hypertarget{demographics}{patient demographics} are reported in Table~\ref{tab:sum_stats_2018}. 

\begin{table*}[!htb]
    \centering
    \caption{Summary statistics of the patient demographics in the selected cohort in the real-data example. Q1 and Q3 stand for the $25\%$ and $75\%$ quantiles (also gives the interquartile range).}\label{tab:sum_stats_2018}
    \vspace{.1in}
    \resizebox{.95\textwidth}{!}{
        \begin{tabular}{lcccc}
        \toprule
        &  \multicolumn{2}{c}{Source} & \multicolumn{2}{c}{Target} \\
        \cmidrule(l){2-5}
        & Treatment & Control & Treatment & Control \\
        Number of patients & $207$ & $1249$ & $58$ & $700$ \\
        Age, median and [Q1,Q3]  & $63.0_{\ [51.0, 70.5]}$ & $64.0_{\ [53.0, 74.0]}$ & $55.5_{\ [37.25, 67.5]}$ & $58.0_{\ [41.0, 68.0]}$ \\
        Male, number and percentage & $131_{\ (63.3 \%)}$ & $652_{\ (52.2 \%)}$ & $38_{\ (65.5 \%)}$ & $449_{\ (64.1 \%)}$ \\
        28D-Mortality, number and percentage & $43_{\ (20.8 \%)}$ & $117_{\ (9.4 \%)}$ & $20_{\ (34.5 \%)}$ & $71_{\ (10.1 \%)}$ \\
        Total hospital days, median and [Q1,Q3] & $22.0_{\ [12.5, 33.5]}$ & $13.0_{\ [8.0, 21.0]}$ & $25.5_{\ [13.0, 45.5]}$ & $18.0_{\ [10.0, 31.0]}$ \\
        \bottomrule
        \end{tabular}
    } 
\end{table*}

Due to {\it limited} data in the target domain, our preliminary analysis using IPW estimator shows that vasopressor therapy increases the risk of 28-day mortality, which contradicts established result in medical literature \citep{avni_vasopressors_2015} (see Section~\ref{sec:real_exp} for complete results). We aim to correct this using the available source domain dataset, which is derived from one geographically adjacent academic level 1 trauma center (patient demographics also reported in Table~\ref{tab:sum_stats_2018}). Note that, however, naive integration of both datasets is impractical since the patients may not only differ in the treatment assignment mechanism (see Figure~\ref{fig:support} for empirical evidence) but also in the way they respond to treatment. Therefore, the current objective is to find a principled TL approach to integrate {\it abundant} data from the source domain to improve the estimation accuracy of the target domain causal effect.

\subsection{Literature}

TL has been applied to causal inference in a distinct yet more straightforward manner, owing to the unique treatment-and-control structure. For instance, \citet{shalit2017estimating,shi2019adapting} proposed a novel NN architecture tailored to causal effect estimation by considering shared and group-specific layers in the potential outcome models for treatment and control groups.
However, adapting TL techniques from the supervised learning setting to handle data integration for causal effect estimation is non-trivial, as it requires counterfactual information.
In causal inference, this problem is solved by the aforementioned plug-in estimators (e.g., IPW, OR, and DR estimators), which involve preliminary stage nuisance parameter estimation for the propensity score (PS) and/or OR models.
Hence, a natural solution is to apply data-integrative TL to the supervised nuisance parameter estimation problem and subsequently evaluate the plug-in estimators for ACE using target domain data, where the hope is to improve ACE estimation accuracy by enhancing the quality of estimated nuisance parameters, regardless of whether the ground truth ACEs are the same across both domains.

While there has been increased interest in applying data-integrative TL techniques to causal inference in the presence of heterogeneous covariate spaces \citep{yang2020combining,wu2022transfer,hatt2022combining,bica2022transfer}, these methods typically fail to handle the same covariate space setting, known as the inductive multi-task transfer learning according to \citet{pan2010survey}. This limitation arises from their algorithm designs, which mostly rely on domain-specific covariate spaces. 
To the best of our knowledge, \citet{kunzel2018transfer} is the first and only work studying data-integrative TL for causal effect estimation under the inductive multi-task setting, referred to as the {\it Transfer Causal Learning} (TCL) problem. They proposed to transfer knowledge by using neural network (NN) weights estimated from the source domain as the warm start of the subsequent target domain NN training. 
Despite its improved empirical performance, the theoretically grounded approach for the TCL problem is still largely missing. For other related works on applying TL in causal inference, we refer readers to an extended literature review in Appendix~\ref{sec:literature} and a nice survey by \citet{yao2021survey}.

\subsection{Contribution}

In this work, we address the gap in the literature by presenting a generic framework for the Transfer Causal Learning problem, called \texttt{$\ell_1$-TCL} framework. It entails data-integrative transfer learning of the nuisance parameter and plug-in estimation for causal effect in the target domain. The transfer learning stage comprises two steps: (i) rough estimation using abundant source domain data, and (ii) bias correction step via $\ell_1$ regularized estimation of the difference between the target and source domain nuisance parameters using target domain data. Subsequently, the estimated nuisance parameters are plugged into the unbiased causal effect estimators, including IPW, OR, and DR estimators. Most importantly, as shown in \citet{bastani2021predicting}, with the help of Lasso in the high-dimensional setting, we can establish non-asymptotic recovery guarantees for the causal effect estimators when the nuisance models (i.e., PS and OR models) are parameterized as generalized linear models (GLMs) and under the sparsity assumption on the target and source nuisance parameters' difference. Furthermore, we incorporate recently developed non-parametric PS and OR models in \texttt{$\ell_1$-TCL} to improve robustness to model misspecification, and we show improved performance of our \texttt{$\ell_1$-TCL} framework using NN-based approaches \citep{shalit2017estimating,shi2019adapting} by comparing with existing TL approaches \citep{kunzel2018transfer} for ACE estimation on a benchmark pseudo-real-dataset \citep{brooks1992effects}. The \texttt{$\ell_1$-TCL} framework is subsequently applied to a real study and reveals that vasopressor therapy could prevent mortality within septic patients \citep{avni_vasopressors_2015}, which all baseline approaches fail to show.

\subsection{Notations}

%The notations used here follow standard conventions.
Superscript $^\T$ denotes vector or matrix transpose, and $\norm{\cdot}_p$ denotes the vector $\ell_p$ norm. We use upper case letters (e.g., $\boldsymbol{X}$) to denote random variables (r.v.s) and the corresponding lower case letters with additional subscripts (e.g., $\boldsymbol{x}_1, \boldsymbol{x}_2, \dots$) to denote their realizations.
For asymptotic notations: 
$f(n) = o(g(n))$ or $g(n) \gg f(n)$ means for all $c > 0$ there exists $k > 0$ such that $0 \leq f(n) < cg(n)$ for all $n \geq k$;
$f(n) =  \cO(g(n))$ means there exist positive constants $c$ and $k$, such that $0 \leq f(n) \leq cg(n)$ for all $n \geq k$.

%\vspace{-0.1in}
\section{Methodology}
%\vspace{-0.1in}
We study the causal inference under Neyman–Rubin potential outcome framework \citep{rubin1974estimating,splawa1990application}. 
This section will introduce our \texttt{$\ell_1$-TCL} where the nuisance models are parameterized as the Generalized Linear Models. We will briefly review the IPW estimator for causal effect estimation, introduce the formal TCL problem setup, and present our proposed adaption of $\ell_1$ regularized TL techniques in the IPW estimator. Lastly, we will extend those contents above to OR and DR estimators.

%\vspace{-0.1in}

%\vspace{-0.1in}
\subsection{Background}
%\vspace{-0.05in}

\paragraph{Potential outcome framework.}
Consider the tuple $(\boldsymbol{X}, {Z}, {Y})$ in the target study, where random vector $\boldsymbol{X} \in \cX \subset \RR^d$ represents covariates measured before receipt of treatment, r.v. ${Z}\in \{0,1\}$ is treatment indicator (${Z}= 1$ if treated and $0$ otherwise) and r.v. $Y$ is the {\it observed outcome}:
\[{Y} = {Y}_{1} {Z}+ (1-{Z}) {Y}_{0}.\]
Here, ${Y}_{0}$ and ${Y}_{1}$, referred to as {\it potential outcomes}, are the values of the outcome that would be seen if the subject were to receive control or treatment. Throughout this work, we are interested in estimating the ACE or average treatment effect, which is formally defined as:
\[\tau = \EE[{Y}_{1}] - \EE[{Y}_{0}].\]
In an observational study, the treatment ${Z}$ is typically not statistically independent from $({Y}_{0}, {Y}_{1})$ since the characteristics that determine the treatment assignment may also be correlated, or ``confounded'', with the potential outcome. To handle this problem, a common practice is to assume there are ``no unmeasured confounders'' (also known as the Ignorability Assumption):
\[({Y}_{0}, {Y}_{1}) \independent {Z} \mid \boldsymbol{X}.\]
In the following, we shall continue our study under the above assumption. 

%\vspace{-0.1in}

% meaning of unbiased, as the word is overloaded and means differently in causal inference and in ML fairness.

\paragraph{Inverse probability weighting.}
The propensity score $e(\boldsymbol{X}) = \PP({Z} = 1|\boldsymbol{X})$ is the probability of treatment given covariates and specifies the treatment assignment mechanism. \citet{rosenbaum1983central} showed:
\[({Y}_{0}, {Y}_{1}) \independent {Z} \mid e(\boldsymbol{X}),\]
which leads to an unbiased estimator for ACE through inverse probability weighting. Consider $n$ samples from the target domain:
\begin{equation}\label{eq:obs_target}
    \cD_{i} = (\boldsymbol{x}_{i}, z_{i}, y_{i}), \quad i = 1,\dots, n,
\end{equation}
and let $\hat e(\boldsymbol{x}_{i})$ be the estimated propensity score for $i$-th subject, the IPW estimator for ACE is:
\begin{equation}\label{eq:ipw}
    \hat \tau_{\rm IPW} = \frac{1}{n} \sum_{i = 1}^{n} \frac{z_{i} y_{i}}{\hat e(\boldsymbol{x}_{i})} - \frac{(1-z_{i}) y_{i}}{1-\hat e(\boldsymbol{x}_{i})}.
\end{equation}
For further background knowledge on the causal inference, such as a brief introduction of OR and DR estimators and why those estimators are unbiased, we refer readers to Appendix~\ref{appendix:causal_background} and some nice survey studies \citep{lunceford2004stratification,bang2005doubly,yao2021survey}.

%\vspace{-0.1in}
\subsection{Problem setup}\label{sec:setup}
%\vspace{-0.05in}

\paragraph{Inductive multi-task transfer learning for causal inference.}
As mentioned earlier, our TCL essentially aims to conduct causal effect estimation in the inductive multi-task TL setup \citep{pan2010survey}, where the feature spaces across both domains are the same, but the feature distribution, treatment assignment rule, and underlying causal effect can be different.

From now on, we will refer to \eqref{eq:obs_target} as samples from the target domain.
Assume we additionally observe $n_{\mys}$ samples of the covariates, treatment and outcome tuple $(\boldsymbol{X}_{\mys}, Z_{\mys}, Y_{\mys})$ from the source domain:
\begin{equation*}
    \cD_{i,\mys} = (\boldsymbol{x}_{i, \mys}, z_{i, \mys}, y_{i, \mys}), \quad i = 1,\dots, n_{\mys}.
\end{equation*}

In our motivating real example, $n \ll n_{\mys}$, rendering it difficult to get an accurate ACE estimate by solely using target domain data and necessitating the use of source domain data. 
However, a practical issue often arises that neither the nuisance models (e.g., the PS model) nor the ground truth ACEs are the same between both domains, making naively merging two datasets impractical. To be precise, consider that the PS model takes the following form:
\begin{equation}\label{eq:general_propensityscore}
\begin{split}
    \PP(Z = 1 | \boldsymbol{X}) = e(\boldsymbol{X}; \beta_\myt), \quad \PP(Z_\mys = 1 | \boldsymbol{X}_\mys) = e_{\mys}(\boldsymbol{X}_\mys; \beta_\mys),
\end{split} 
\end{equation}
where functions $e(\cdot)$, $e_{\mys}(\cdot)$ have known form with unknown $d_1$-dimensional nuisance parameters, i.e., $\beta_\myt, \beta_\mys \in \RR^{d_1}$. In our TCL problem, we aim to develop a principled method to integrate data from both domains to help estimate the ACE in the target domain. 

\paragraph{An illustrative example.}
To help readers understand this TCL setup and elucidate why the TCL problem is non-trivial, we present a toy example as follows: 
\begin{equation*}
    \begin{split}
        \underline{\text{\rm Treatment assignment}}: \quad & \PP(Z = 1 | X_1, X_2) = g(\beta_1 X_1 + \beta_2 X_2), \\
    \underline{\text{\rm Causal relationship}}: \quad & Y = \tau Z +  \alpha X_2 + \epsilon,
    \end{split}
\end{equation*}
\begin{wrapfigure}{r}{0.5\textwidth}
\centering
\vspace{-0.27in}
\subfigure{\includegraphics[width = 0.5\textwidth]{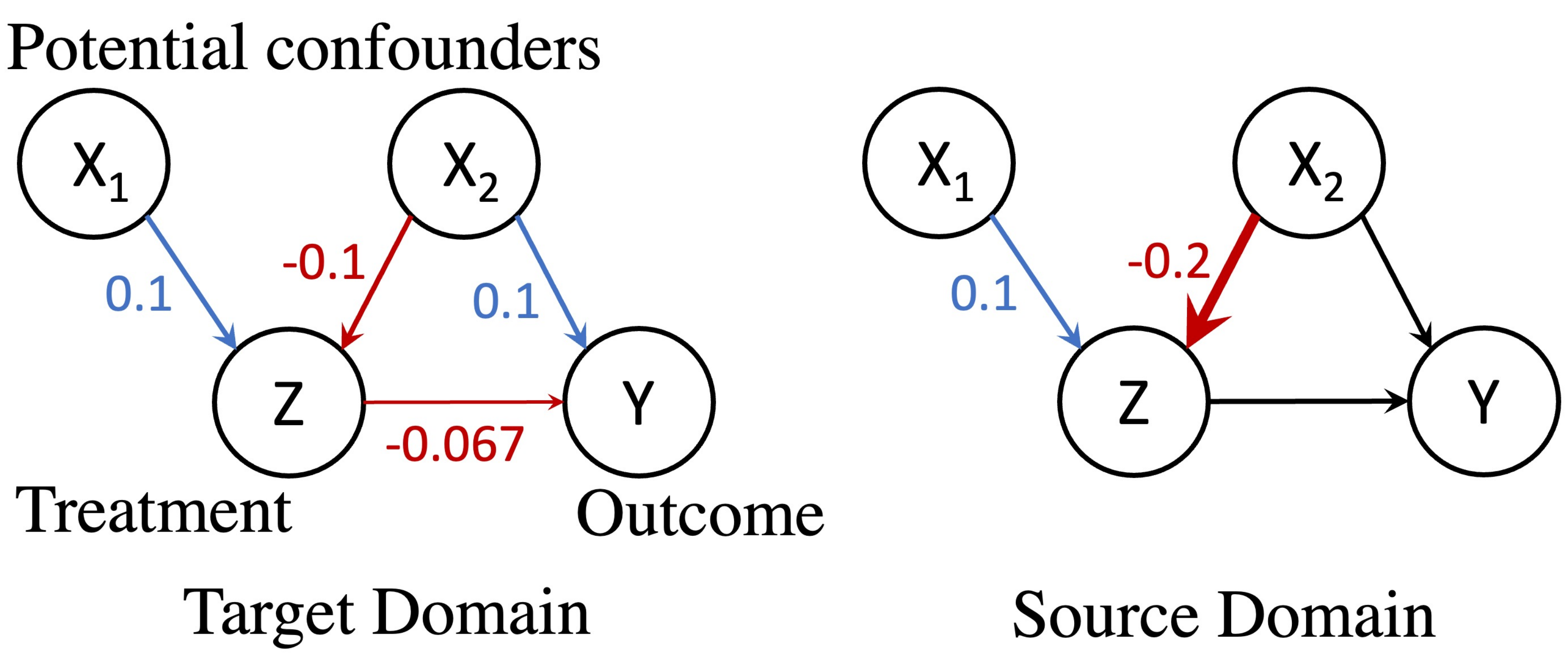}}
\vspace{-0.27in}
\caption{In our toy example, the treatment assignments differ between target and source domains in that the effects from covariate $X_2$ are different. We do not impose assumptions on whether or not the ACEs are the same for both domains.}
\label{fig:illus} 
\vspace{-.1in}
\end{wrapfigure}
where $g(x) = 1/(1+e^x)$ is the sigmoid function. 
The goal is to infer the causal effect from treatment $Z$ to outcome $Y$, given potential confounding variables $X_1$ and $X_2$; the additive noise $\epsilon$ is independent of the aforementioned r.v.s. The treatment assignment mechanism and the causal relationship are visualized in Figure~\ref{fig:illus}; further experimental details, such as the configurations, can be found in Appendix~\ref{appendix:syn_exp}.

Although IPW is consistent \citep{wooldridge2002inverse,wooldridge2007inverse}, making inferences from the limited amount of target domain data leads to an estimate of the ACE with large bias, as verified by the estimate using target-only causal learning (\texttt{TO-CL}) in Table~\ref{tab:illus}, and this necessitates the use of source domain data. 
One naive way is to integrate (or merge) both datasets in the PS model nuisance parameter estimation stage for causal learning, i.e., \texttt{Merge-CL}. 
However, due to different treatment assignments, \texttt{Merge-CL} will not help correct the bias. To make things even worse, since we have $n_\mys \gg n$, this \texttt{Merge-CL} estimate will bias towards the source domain, leading to a potentially worse downstream plug-in IPW estimator, as verified in Table~\ref{tab:illus}. 

\begin{table}[!htp]
\centering
    %%\vspace{-0.12in}
    \caption{Comparison of ACE estimation accuracy: the truth is $\tau = -0.067$. Our proposed \texttt{$\ell_1$-TCL} yields the most accurate one, which correctly recovers the {\it inhibiting} effect.}
    \label{tab:illus}
    \vspace{.1in}
    \resizebox{.6\textwidth}{!}{
    \begin{tabular}{lccc}
    \toprule
    Data used & Target only & \multicolumn{2}{c}{Both domains} \\
    Learning framework & \texttt{TO-CL} & \texttt{Merge-CL} & \texttt{$\ell_1$-TCL} \\
    \cmidrule(l){2-4}
    IPW estimate & 0.0002 & 0.0441 & $-$0.0013  \\
    \bottomrule
    \end{tabular}
    %%\vspace{-0.3in}
    }
\end{table}

%\vspace{-0.1in}
\subsection{Proposed method}
%\vspace{-0.05in}

We begin with a simple yet popular Generalized Linear Model (GLM) \citep{nelder1972generalized} parameterization of the {\it nuisance models} (i.e., PS \eqref{eq:general_propensityscore} and OR \eqref{eq:general_outcomeregression} models). A GLM for r.v. $\tilde Z$ with parameter $\beta$ and predictor ${\boldsymbol{\tilde X}}$ is: 
\begin{equation*}%\label{eq:GLM_general}
    \tilde Z \mid \boldsymbol{\tilde X} \sim \PP(\tilde Z | \boldsymbol{\tilde X}) = F(\tilde Z) \exp\{\tilde Z {\boldsymbol{\tilde X}}^\T \beta - G({\boldsymbol{\tilde X}}^\T \beta) \}, \ \ \text{\rm which satisfies } \ \ \EE[\tilde Z|{\boldsymbol{\tilde X}}] = G'({\boldsymbol{\tilde X}}^\T \beta).
\end{equation*}
Here, $G'(\cdot)$, known as the (inverse) link function, is the derivative of $G(\cdot)$; common non-linear choices include sigmoid link function $G'(x) = 1/(1+e^{-x})$ on a domain $x \in \RR$ and exponential link function $G'(x) = 1 - e^{-x}$ on a domain $x \in [0, \infty)$. The function $F(\cdot)$ normalizes, ensuring a valid probability distribution. Given samples $(\boldsymbol{\tilde x_i},\tilde z_i), \ i = 1,\dots,n$, the maximum likelihood estimation (MLE) of the GLM model parameter is given by: 
\[\hat{\beta}_{\rm MLE} = \arg \min _b \sum_{i=1}^n -\tilde z_i \boldsymbol{\tilde x_i}^{\T} b+G\left(\boldsymbol{\tilde x_i}^{\T} b\right) .\]

As the treatment indicator is binary, the GLM parameterization with link function $G'(\cdot) = g(\cdot)$ can be expressed as follows:
\begin{equation}\label{eq:GLM_propensityscore}
\begin{split}
     \EE[Z|\boldsymbol{X}] = \PP(Z = 1 | \boldsymbol{X}) = g(\boldsymbol{X}^\T \beta_\myt), \quad \EE[Z_\mys|\boldsymbol{X}_\mys] = \PP(Z_\mys = 1 | \boldsymbol{X}_\mys) = g(\boldsymbol{X}_\mys^\T \beta_\mys).
\end{split} 
\end{equation}
Here, the nuisance parameters $\beta_\myt, \beta_\mys$ have dimensionality $d_1 = d$. 
Without loss of generality, we consider the same link functions in both domains for simplicity; however, the success of the knowledge transfer does not require the ``same link function'' as long as the link functions are known.

%\vspace{-0.1in}

\paragraph{Assumption to guarantee knowledge transferability.} The key assumption guaranteeing the success of the knowledge transfer is the sparsity of the nuisance parameter difference $\Delta_\beta$, which is defined as: 
\begin{equation}\label{eq:beta_diff}
    \Delta_\beta = \beta_\myt - \beta_\mys.
\end{equation}

%\vspace{-0.05in}

\begin{definition}[$s$-sparse vector]
A vector $v \in \RR^d$ is said to be $s$-sparse (with $0 \leq s \leq d$) if this vector has at most $s$ non-zero elements, i.e., $\norm{v}_0 \leq s.$
\end{definition}

\begin{figure}[!htp]
\centerline{
\includegraphics[width = .75\textwidth]{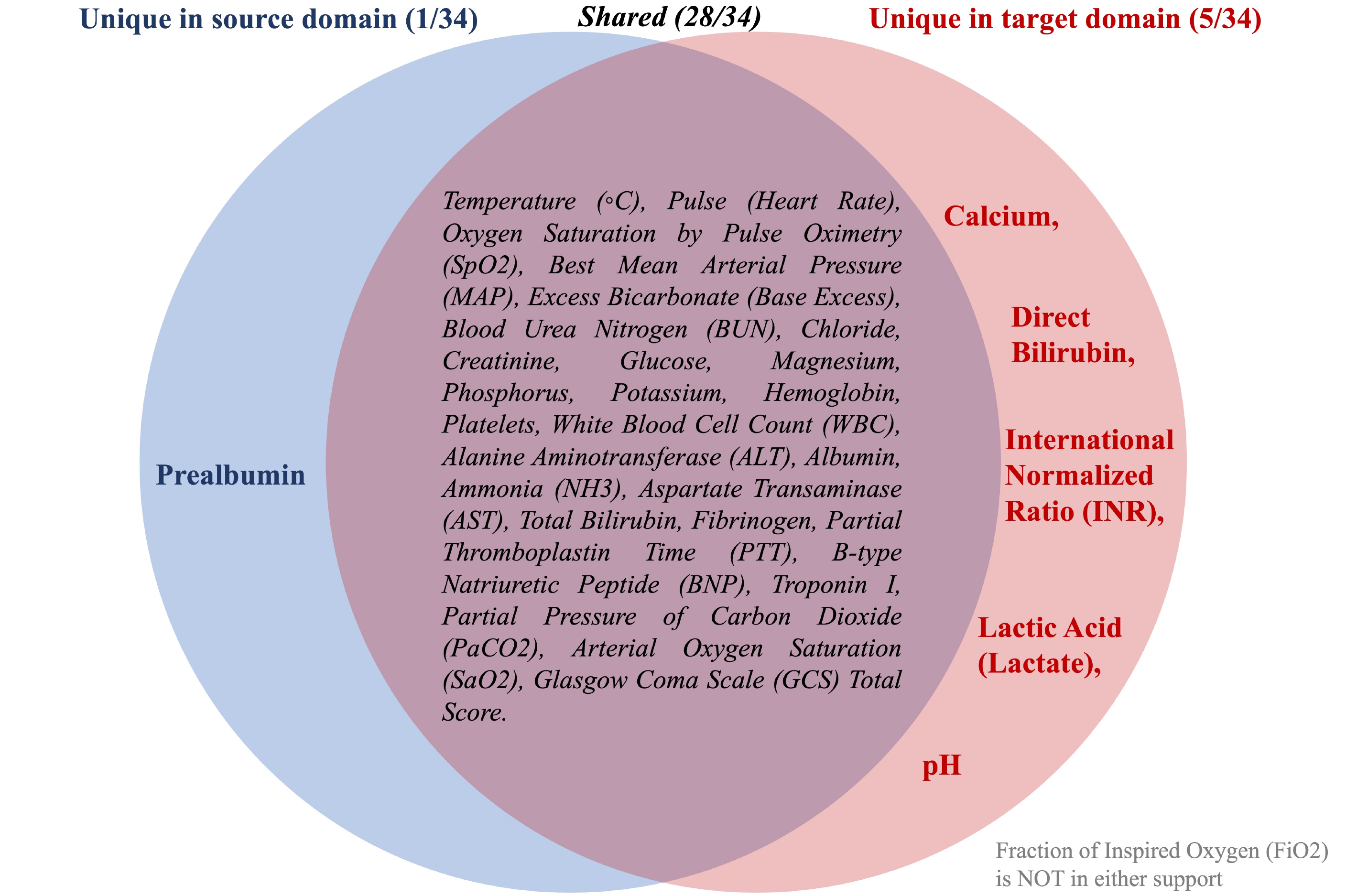} 
}
\vspace{-0.05in}
\caption{Support of the PS model parameters via $\ell_1$ regularized logistic regression in both domains in the motivating real example. We can see that they share very similar supports, and their difference is only supported on 6 out of a total of 34 features (listed in Table~\ref{table:real_data_features} in Section~\ref{sec:real_exp}). Since the PS model parameter essentially shows how the clinicians assign treatment based on collected EMR data, the ``similar support'' observation could be evidence that $\Delta_\beta$ in our real example is sparse.}
\label{fig:support}
\vspace{-0.05in}
\end{figure}

We posit that the mechanisms guiding treatment assignments should exhibit substantial similarity across both domains, a concept encapsulated by the {\it $s$-sparse $\Delta_\beta$ assumption}. This assumption is supported by the fact that the two trauma centers involved in our study are geographically adjacent and share a common set of clinicians. These shared clinicians likely apply consistent decision-making criteria when treating patients, as inferred from the supports of the fitted PS model parameters shown in Figure~\ref{fig:support}. %Therefore, the resemblance in clinical decision-making patterns across these centers supports our assumption of a sparse difference $\Delta_\beta$ between domains, reflecting minor variations in treatment assignments rather than fundamental differences.

\paragraph{$\ell_1$ regularized data-integrative TL for PS model.}
The first stage involves two steps: (i) leveraging abundant source domain data to estimate the source parameter $\beta_\mys$, which serves as a rough estimator of $\beta_\myt$ due to their sparse difference, and (ii) using $\ell_1$ regularization to learn the difference $\Delta_\beta$ from target domain data, which corrects the bias of the first-step rough estimator, i.e., %Formally, the TL estimate of $\beta_\myt$ is:
\begin{equation}\label{eq:step2}
    \begin{split}
         \underline{\text{\rm Rough estimation for PS model}}&:  \hat \beta_\mys = \arg \min_b \frac{1}{n_\mys} \sum_{i=1}^{n_\mys} -z_{i, \mys} \boldsymbol{x}_{i, \mys}^{\T} b+G\left(\boldsymbol{x}_{i, \mys}^{\T} b\right),  \\ \noalign{\vskip-2.5pt}
    \underline{\text{\rm Bias correction for PS model}}&:  \hat \beta_\myt = \arg \min_b \frac{1}{n} \sum_{i=1}^{n} -z_{i} \boldsymbol{x}_{i}^{\T} b+G\left(\boldsymbol{x}_{i}^{\T} b\right) + \lambda_{\rm PS} \norm{b - \hat \beta_\mys}_1. 
    \end{split}
\end{equation}
Here, $\lambda_{\rm PS} > 0$ is a tunable regularization strength hyperparameter and will be selected via cross-validation (CV) in practice.
Equivalently, the bias correction step can be expressed as: 
\begin{equation}\label{eq:re-formulation}
      \hat \beta_\myt = \hat{\Delta}_\beta + \hat \beta_\mys, \quad  \hat{\Delta}_\beta = \arg \min _\Delta \sum_{i=1}^{n} -z_{i} \boldsymbol{x}_{i}^{\T} (\Delta + \hat \beta_\mys)+G\left(\boldsymbol{x}_{i}^{\T} (\Delta + \hat \beta_\mys)\right)  + \lambda_{\rm PS} \norm{\Delta}_1.
\end{equation}
Later in Section~\ref{sec:theory}, we will show that, even when $n \ll d$ in the bias correction step, with the help of high-dimensional Lasso, $\Delta_\beta$ can be faithfully recovered with theoretical guarantees. This is quite intuitive: source domain nuisance parameters can be faithfully recovered using a large amount of source domain data, whereas the sparsity assumption guarantees valid inference of the difference using target domain data via $\ell_1$ regularization.

\paragraph{Plug-in estimation for the Average Causal Effect.}
%\vspace{-0.05in}
In the second stage, the above fitted PS model parameters via TL techniques are plugged into the IPW \eqref{eq:ipw} estimator to get the GLM-based \texttt{$\ell_1$-TCL} estimate of the ACE, called the TLIPW estimate, as follows:
\begin{equation}\label{eq:TLIPW}
    \begin{split}
        \hat \tau_{\rm TLIPW} & = \frac{1}{n} \sum_{i = 1}^{n} \frac{z_{i} y_{i}}{g(\boldsymbol{x}_{i}^{\T}\hat \beta_\myt )} - \frac{(1-z_{i}) y_{i}}{1-g(\boldsymbol{x}_{i}^{\T}\hat \beta_\myt )}. 
    \end{split}
\end{equation}

\subsection{Extension to OR and DR estimators}
Alternatively, depending on the confidence in the PS and/or OR model specification, one could apply similar TL techniques to OR or DR estimators (reviewed and defined in \eqref{eq:or} and \eqref{eq:dr} in Appendix~\ref{appendix:causal_background}, respectively) to get the GLM-based \texttt{$\ell_1$-TCL} estimate of the ACE, which we will call TLOR and TLDR estimators.

Following our GLM parameterization, in the TCL problem setup, the OR model has the following form: for $z \in \{0,1\}$,
\begin{equation}\label{eq:general_outcomeregression}
    \begin{split}
        \EE[Y_z | \boldsymbol{X}] = m_z(\boldsymbol{X}; \alpha_{z,\myt}), \quad  \EE[Y_{z,\mys} | \boldsymbol{X}_\mys] = m_{z,\mys}( \boldsymbol{X}_\mys; \alpha_{z,\mys}),
    \end{split}
\end{equation}
where functions $m_z(\cdot), m_{z,\mys}(\cdot)$ have known form with unknown nuisance parameters $\alpha_{z,\myt}, \alpha_{z,\mys} \in \RR^{d_2}$. For simplicity, we parameterize the OR model via linear regression from now on; however, our method and theory (to be presented) can be extended to handle GLM parameterization for the OR model. Specifically, for $z \in \{0,1\}$, let
\begin{equation}\label{eq:LM_outcomeregression}
    \begin{split}
         \EE[Y_z | \boldsymbol{X}] = \boldsymbol{X}^\T \alpha_{z,\myt}, \quad \EE[Y_{z,\mys} | \boldsymbol{X}_\mys] = \boldsymbol{X}_\mys^\T \alpha_{z,\mys},
    \end{split}
\end{equation}
where the OR model nuisance parameters have dimensionality $d_2 = d$. 

Similarly, the transferability guarantee comes from the assumption of the following differences:
\begin{equation}\label{eq:alpha_diff}
    \Delta_{\alpha,z} = \alpha_{z,\myt} - \alpha_{z,\mys}, \quad z \in \{0,1\},
\end{equation}
are $s$-sparse, i.e., $\norm{\Delta_{\alpha,0}}_0, \norm{\Delta_{\alpha,1}}_0 \leq s$. This enables us to apply those above $\ell_1$ regularized TL techniques to estimate the OR model parameters in the target domain with the help of source domain data: For $z \in \{0,1\}$, denote $n_{z,\mys} = \#\{i: z_{i, \mys} = z\}$ ($\#$ represents the cardinality of a set), and let $\lambda_{\rm OR} > 0$ be the tunable regularization strength hyperparameter:
\begin{equation}\label{eq:step4_DR}
    \begin{split}
        \underline{\text{\rm Rough estimation for OR model}}&:  \hat \alpha_{z,\mys} = \arg \min_\alpha \frac{1}{n_{z,\mys}} \sum_{z_{i, \mys} = z} \big(y_{i, \mys} - \boldsymbol x_{i,\mys}^\T \alpha \big)^2,  \\ \noalign{\vskip-2.5pt}
        \underline{\text{\rm Bias correction for OR model}}&:  \hat \alpha_{z,\myt}  = \arg \min_\alpha \frac{1}{n_z} \sum_{z_{i} = z} \big(y_{i} - \boldsymbol x_{i}^\T \alpha \big)^2 + \lambda_{\rm OR} \norm{\alpha - \hat \alpha_{z,\mys}}_1. %\nonumber
    \end{split}
\end{equation}

Finally, we are ready to plug the TL estimates of the nuisance model parameters to get the \texttt{$\ell_1$-TCL} estimates of the ACE:
\begin{align}
        \hat \tau_{\rm TLOR} & = \frac{1}{n_1} \sum_{z_i = 1} \boldsymbol{x}_{i}^\T \hat \alpha_{1,\myt} - \frac{1}{n_0} \sum_{z_i = 0} \boldsymbol{x}_{i}^\T \hat \alpha_{0,\myt},  \label{eq:TLOR}\\
    \hat \tau_{\rm TLDR} & = \frac{1}{n} \sum_{i = 1}^{n}  \frac{z_{i} y_{i} - \boldsymbol{x}_{i}^{\T}\hat \alpha_{1,\myt}(z_i - g(\boldsymbol{x}_{i}^{\T}\hat \beta_\myt ))}{g(\boldsymbol{x}_{i}^{\T}\hat \beta_\myt )}  - \frac{(1-z_{i}) y_{i} + \boldsymbol{x}_{i}^{\T}\hat \alpha_{0,\myt}(z_i - g(\boldsymbol{x}_{i}^{\T}\hat \beta_\myt ))}{1-g(\boldsymbol{x}_{i}^{\T}\hat \beta_\myt )}. \label{eq:TLDR}
\end{align}

%\vspace{-0.1in}
\section{Theoretical Analysis}\label{sec:theory}
%\vspace{-0.05in}

Typically, to make valid inferences by solely using target domain data, we need a sufficiently large amount of target domain data such that $n \gg d$. However, in our setting, such an assumption does not hold; to make things even worse, we may encounter $n < d$ case. Fortunately, with the help of Lasso for high-dimensional regression, recovery guarantees can still be established when we have abundant source domain data, which only require target domain sample size $n$ to be on the order of $\log d$. In this section, we present the main results and their interpretations; complete details, including the technical assumptions and proofs, can be found in Appendices~\ref{appendix:TLIPW}, \ref{appendix:TLOR}, and \ref{appendix:TLDR}.

%\vspace{-0.1in}

\paragraph{Main theoretical results.}
When the PS model is correctly specified, and the difference is $s$-sparse, i.e., $\norm{\Delta_{\beta}}_0 \leq s$, in the large sample limit $n, n_\mys \rightarrow \infty$, consider the following {\it regime}:
\begin{equation}\label{eq:regime}
    n  \gg  s^2 \log d, \quad n_\mys  \gg  n d^2,
\end{equation}
By taking
\[\lambda_{\rm PS}= \cO\left(\sqrt{\log d} \left(\frac{1}{\sqrt{n}} + \frac{d}{\sqrt{n_\mys}}\right)\right),\]
we can show that, with probability at least $1- 1/n$, the absolute estimation error is upper bounded as: %of both $\hat \tau_{\rm TLIPW}$ and $\hat \tau_{\rm TLDR}$ is bounded as follows:
\begin{equation*}
    \begin{split}
        \left| \hat \tau_{\rm TL} - \tau \right| \ = \cO \bigg( \underbrace{s \sqrt{\frac{\log d}{n}}}_{\text{\rm{\it bias correction error}}} + \underbrace{s d \sqrt{\frac{\log d}{n_\mys}}}_{\text{\rm{\it rough estimation error}}}\bigg), 
    \end{split}
\end{equation*}
where $\hat \tau_{\rm TL}$ can be either the TLIPW estimator $\hat \tau_{\rm TLIPW}$ \eqref{eq:TLIPW} or the TLDR estimator $\hat \tau_{\rm TLDR}$ \eqref{eq:TLDR}.

%\vspace{-0.1in}

\paragraph{Interpretations.}
Similar to the two-stage estimation, i.e., nuisance parameter recovery and plug-in estimation for ACE, the proofs are done by plugging the non-asymptotic upper bound on the vector $\ell_1$-norm of the nuisance parameter to the absolute error bound of the downstream plug-in estimators, resulting in {\it the error bound decomposition as above}. In particular, the bias correction term is $\cO(s \sqrt{\log d/n})$ (which aligns with that of the classic Lasso estimator, cf. Theorem 7.1 \citep{bickel2009simultaneous}) and dominates the rough estimation error term due to $n_\mys  \gg  n d^2$ \eqref{eq:regime}; however, according to the above error upper bound, the condition on source domain sample size can be relaxed to $n_\mys \gg s^2 d^2 \log d$ to achieve consistency. Without the help of the source domain, the overall error rate will be similar to that of the rough estimation, which requires $n \gg d^2$ target domain samples to achieve a satisfying error bound (cf. Theorem 1 \citep{bastani2021predicting}). In contrast, the abundant source domain data, characterized by $n_\mys  \gg  n d^2$ in the considered regime \eqref{eq:regime}, relaxes the requirement on target domain sample size to $n \gg s^2 \log d$ to achieve the same satisfying error bound.

In our proof, we invoke the Compatibility Condition \citep{bastani2021predicting} for the sample covariance matrix, which is standard in high-dimensional Lasso literature; alternatively, as suggested in Remark 1 \citep{bastani2021predicting}, if we consider the classic Restricted Eigenvalue Condition \citep{bickel2009simultaneous,meinshausen2009lasso,van2009conditions}, we can prove $\ell_2$ error bound that scales as $\sqrt{s}$ instead of $s$; see Remark~\ref{remark:l1_over_l2} in Appendix~\ref{appendix:TLIPW} on why we consider $\ell_1$ error bound for nuisance parameter estimation over the $\ell_2$ bound. 
Lastly, our non-asymptotic analysis shows that the error upper bound with probability at least $1 - \varepsilon$ (for any $\varepsilon \in (0,1)$) will have a $\cO(s\sqrt{\log(1/\epsilon) / n})$ term. When we consider the probability converging to one at a polynomial rate, i.e., $\varepsilon = 1/n^\kappa$ for positive integer $\kappa$, this term will be $\cO(s\sqrt{\log n^\kappa / n})$ and dominated by the $\cO(s\sqrt{\log d / n})$ term in the above bound under our considered regime \eqref{eq:regime}. The above result corresponds to the $\kappa = 1$ case.

%\vspace{-0.1in}

\paragraph{Additional results for correctly specified OR model.}
When the OR model specification is correct with $s$-sparse differences $\norm{\Delta_{\alpha,z}}_0 \leq s$ ($z \in \{0,1\}$), if the samples in the treatment and control groups are ``balanced'' in the sense that there exists a constant $r \in (0,1)$ such that, for $z \in \{0,1\}$,
\begin{equation}\label{Asampleratio}
    \liminf_{n \rightarrow \infty} \frac{n_z}{n} \geq r, \quad \liminf_{n_\mys \rightarrow \infty} \frac{n_{z,\mys}}{n_\mys} \geq r,
\end{equation}
where $n_z = \#\{i: z_{i} = z\}$ and $n_{z,\mys} = \#\{i: z_{i, \mys} = z\}$, then, by taking
\begin{equation*}
    \lambda_{\rm OR}= \cO \left(\sqrt{\log d} \left(\frac{1}{\sqrt{rn}} + \frac{d}{\sqrt{rn_\mys}}\right)\right),
\end{equation*}
for $\hat \tau_{\rm TL} = \hat \tau_{\rm TLOR}$ or $\hat \tau_{\rm TLDR}$, we can show that with probability at least $1 - 1/n$, the absolute estimation error can be upper bounded as follows:
\begin{equation*}
    \begin{split}
        \left| \hat \tau_{\rm TL} - \tau \right| = \cO \bigg( s \sqrt{\log d} \left(\frac{1}{\sqrt{rn}} + \frac{d}{\sqrt{rn_\mys}}\right) \bigg).
    \end{split}
\end{equation*}

Due to space limitations, complete details are deferred to the Appendix, including the assumptions, lemmas, formal statements of the non-asymptotic theoretical guarantees, and all proofs. To help readers find the results, we provide a summary of the locations of our theories in the Appendix (Appx.); see Table~\ref{tab:theory_table}. Furthermore, the superior empirical performance of the above GLM parametric approach is verified via numerical simulation in Appendix~\ref{appendix:syn_exp}.

\begin{table}[!htp]
\centering
%\vspace{-.05in}
\caption{Locations of all non-asymptotic results.}
\label{tab:theory_table}
\vspace{.1in}
    \resizebox{.7\textwidth}{!}{
    \begin{tabular}{lcc}
\toprule
        ~ & Nuisance parameter estimation & Plug-in ACE estimation \\
        \cmidrule(l){2-3}
        {TLIPW} & {Lemma~\ref{lma:TL-beta} (Appx.~\ref{appendix:TLIPW})} & {Theorem~\ref{thm} (Appx.~\ref{appendix:TLIPW})} \\ 
        {TLOR} & {Lemma~\ref{lma:TL-alpha} (Appx.~\ref{appendix:TLOR})} & {Theorem~\ref{thm:OR_0} (Appx.~\ref{appendix:TLOR})} \\ 
        {TLDR} & {Lemma~\ref{lma:TL-beta}, Lemma~\ref{lma:TL-alpha}}  & {Theorem~\ref{thm:OR} (Appx.~\ref{appendix:TLDR})} \\ 
\bottomrule
    \end{tabular}
%\vspace{-.1in}
}
\end{table}

%\vspace{-0.15in}

\section{Real-Data Example}\label{sec:real_exp}

%%\vspace{-0.12in}

In this real experiment, we aim to investigate whether vasopressor therapy can {\it prevent} mortality within sepsis patients. Vasopressor therapy for septic patients has been shown to decrease the risk of 28-day mortality \citep{avni_vasopressors_2015}. Therefore, it is worth observing the underlying causal structure of this treatment. Indeed, \citet{wei2022granger} showed that vasopressor therapy may have an inhibiting effect on sepsis, which suggests its inhibiting effect on mortality as well.

Baseline approaches that only use the target domain data or naively merge both domains' data all indicate statistically significant {\it promoting} effect from treatment (verified by the $90\%$ confidence intervals (CI) of the ACE estimates), which clearly violates common sense. Fortunately, by leveraging our \texttt{$\ell_1$-TCL} framework, we can reach a reasonable conclusion that vasopressor therapy does {\it prevent} mortality within sepsis patients. Due to space limitations, complete details, such as training details, are deferred to Appendix~\ref{appendix:add_real_result}.

%\vspace{-0.1in}

\paragraph{Data description.}
We construct a retrospective cohort of patients using in-hospital data from two adjacent academic, level 1 trauma centers located in the South Eastern United States in 2018 (see patient demographics in Table~\ref{tab:sum_stats_2018} in Section~\ref{sec:motivation}). The data was collected and analyzed in accordance with an institutional review board and relevant ethics approval information will be provided if the paper is accepted. A total of 34 patient covariates comprised of vital signs and laboratory (Lab) results are examined in this study. Patients are considered to be treated if they received vasopressor therapy, which is defined as receiving norepinephrine, epinephrine, dobutamine, dopamine, phenylephrine, or vasopressin, at any time within the 12-hour window before sepsis onset. The outcome variable is the 28-day mortality, which is a common metric used by clinicians performing observational studies on sepsis patients \citep{stevenson_two_2014}.

In our study, all patient data for each encounter is binned into hourly windows that begin with hospital admission and end with discharge. If more than one measurement occurs in an hour, then the average of the values is recorded. 
To ensure that causal effect estimation is performed on data series of similar lengths (which could help control potentially unobserved confounding), patients are included in the cohort if they meet the Sepsis-3 criteria during the hospital stay, and we examine exactly 12 hours of data before and after sepsis onset, resulting in a 25-hour subset of the full patient encounter (i.e., the hour of sepsis onset as well as 12 hours before and after this time). 

The 28-day mortality, i.e., the binary outcome variables, is defined as the patient's death within 28 days or less after the time of admission. Covariates from the EMR data include: 
\begin{itemize}
    \item Vital Signs — in the ICU environment, these are usually recorded at hourly intervals. However, patients on the floor may only have  measurements for every 8 hours.
    \item Laboratory Results — the Lab tests are most commonly ordered on a daily basis. However, the collection frequency may change based on the severity of a patient’s illness and the clinician's request.
\end{itemize}
Our study considers in total 4 vital signs and 30 Lab results, as presented in Table \ref{table:real_data_features}. Since the covariate names explain themselves, we omit further descriptions of those covariates.

\begin{table*}[!htp]
    %\tiny
    \centering
    \caption{List of covariates, i.e., vital signs and Lab results, included in the real-data example.}\label{table:real_data_features}
    \vspace{.1in}
    \resizebox{.95\textwidth}{!}{
        \begin{tabular}{cl}
         \toprule
          {Type} & {Name}   \\
          \cmidrule(l){1-2}
          \multirow{2}{*}{Vital sign} & Temperature ($^{\circ}{\rm C}$), Pulse (Heart Rate),  Oxygen Saturation by Pulse Oximetry (SpO2), \\
           & Best Mean Arterial Pressure (MAP).  \\
           \cmidrule(l){2-2}
          \multirow{8}{*}{Lab result} & Excess Bicarbonate (Base Excess), Blood Urea Nitrogen (BUN), Calcium, Chloride, \\
           &  Creatinine,
          Glucose,
          Magnesium, Phosphorus,
          Potassium,
          Hemoglobin,
          Platelets, \\
           & 
          White Blood Cell Count (WBC), Alanine Aminotransferase (ALT),
          Albumin, \\
           & 
          Ammonia (NH3),
          Aspartate Transaminase (AST), Direct Bilirubin,
          Total Bilirubin,
          \\
           & 
          Fibrinogen, International Normalized Ratio (INR), Lactic Acid (Lactate), pH, \\
           & 
          Partial Thromboplastin Time (PTT),
          Prealbumin,  B-type Natriuretic Peptide (BNP), \\
           & 
          Troponin I,
          Fraction of Inspired Oxygen (FiO2), Arterial Oxygen Saturation (SaO2),
          \\
           &  Partial Pressure of Carbon Dioxide (PaCO2),
            Glasgow Coma Scale (GCS) Total Score.\\
           
          \bottomrule
        \end{tabular}
        }
\end{table*}

It is common for vital signs and laboratory results to be missing due to the recording irregularity issues listed earlier. To handle this problem, we impute missing values using the fill-forward method, where any missing hourly values are replaced with the most recent value from the preceding hours. Any remaining missing values are then imputed using the population median. Lastly, for each patient, we take the first data point after the time of the sepsis onset time for our experiments.

%\vspace{-0.1in}

\paragraph{Baseline approaches.}
We choose the PS model parameterized by GLM  \eqref{eq:GLM_propensityscore} with sigmoid link function and IPW estimator for ACE estimation.
We begin with solely using target domain data to estimate ACE, i.e., the \texttt{TO-CL} mentioned in our toy example. By considering both datasets as the target domain dataset, \texttt{TO-CL} yields ACEs $0.12$ in the target domain and $0.057$ in the source domain.
Those results are counterintuitive and clearly contradict established results in the medical literature \citep{avni_vasopressors_2015,wei2022granger}. Indeed, the estimate in the source domain is almost zero, which is closer to the established inhibiting treatment effect \citep{avni_vasopressors_2015} than that of the target domain, potentially due to its larger sample size.
Naively merging two domains' data, which we call \texttt{Merge-CL} framework, is a tempting choice, given that two studied trauma centers sometimes share clinicians; it leads to a point estimate of $0.082$, which aligns with the intuition that \texttt{Merge-CL} ``drags'' the \texttt{TO-CL} estimate of the target domain towards that of the source domain, as the source domain has more samples.

%\vspace{-0.1in}
\begin{table}[!htp]
    \centering
    %\vspace{-0.25in}
    \caption{Comparison of estimated ACEs in the real-data example: the only reasonable result is given by our proposed \texttt{$\ell_1$-TCL}, which indicates {\it inhibiting} causal effect from the vasopressor therapy to 28-day mortality in sepsis patients.}
    \label{tab:real_data}
    \vspace{.1in}
    \resizebox{.75\textwidth}{!}{
    \begin{tabular}{lccc}
    \toprule
    {Data used} & {Target domain only} & \multicolumn{2}{c}{{Both domains}} \\
    {Framework} & {\texttt{TO-CL}} & {\texttt{Merge-CL}} & {\texttt{$\ell_1$-TCL}} \\
    \cmidrule(l){2-4}
    {Point estimate} & {0.120} & {0.082} & {$-$0.011}  \\
    {Bootstrap mean} & {0.072} & {0.130} & {$-$0.853}  \\
    {Bootstrap median} & {0.072} & {0.120} & {$-$0.067}  \\
    {Bootstrap 90\% CI} & {[0.015,  0.134]} & {[0.016,   0.275]} & {[$-$7.257,   1.951]}  \\
    \bottomrule
    \end{tabular}
    %\vspace{-0.3in}
    }
\end{table}

\paragraph{Estimated treatment effects and uncertainty quantification.}
Now, we apply the TLIPW estimator \eqref{eq:TLIPW} in our \texttt{$\ell_1$-TCL}, yielding a point estimate of $-0.011$; this gives a more reasonable conclusion that vasopressor therapy has an inhibiting causal effect on mortality in sepsis patients, agreeing with existing medical literature. Additionally, we perform bootstrap uncertainty quantification (UQ) with $200$ bootstrap trials, each with $700$ random samples (with replacement) from the target domain.
The baseline frameworks (i.e., \texttt{TO-CL} and \texttt{Merge-CL}) all show statistically significant promoting causal effects, verified by the $90\%$ bootstrap CI, which again violates established medical findings.
In contrast, despite the $90\%$ CI containing zero, the mean and median of bootstrap \texttt{$\ell_1$-TCL} causal effect estimates all suggest that vasopressor therapy can prevent 28-day mortality within sepsis patients.

Reliable decision-making is essential in healthcare, a major application of our \texttt{$\ell_1$-TCL}. One common approach is UQ; however, as reflected by the wider bootstrap CI for our \texttt{$\ell_1$-TCL} (compared to that of the baseline approaches), the performance of our $\ell_1$-TCL is sensitive to the choice of hyperparameters ---  oftentimes there exist bootstrap samples where the pre-selected grid does not cover the empirical optimal choice, leading to unreasonably large or small ACE estimates.
It poses a practical challenge in that it requires large computational resources to perform a grid search for hyperparameter selection in each bootstrap trial, rendering vanilla bootstrap impractical.
Currently, the most reliable estimate for drawing causal conclusions in \texttt{$\ell_1$-TCL} framework would be the bootstrap median, which still indicates inhibiting causal effects from the treatment. 
Indeed, this highlights an important future direction, i.e., the development of a principled approach for UQ in the TCL problem. For example, \citet{juditsky2023generalized} recently introduced a new CI construction approach for GLM using a relatively novel concentration result of vector fields. This may facilitate the construction of CI of the nuisance parameters and, hence, the causal effect through the unbiased plug-in estimators. This topic is outside the scope of this work, and we leave it for future study.

\section{A Generic Framework for Transfer Causal Learning}
%\vspace{-0.1in}

Inspired by the superior performance of the GLM-based parametric approach, we now extend our method into a generic framework for the TCL problem by considering arbitrary parameterization of the {\it nuisance model} (i.e., PS \eqref{eq:general_propensityscore} and/or OR \eqref{eq:general_outcomeregression} models), which is called \texttt{$\ell_1$-TCL} framework. This extension can benefit from improved robustness to model misspecification, and it is motivated by a well-known observation \citep{tibshirani1996regression,fan2001variable,zou2005regularization} that, in the presence of the sparsity, $\ell_1$ regularization does not only help establish theoretical guarantee but also improves the estimation accuracy when only limited data is available. Most importantly, \texttt{$\ell_1$-TCL} can be applied to the Conditional Average Causal Effect estimation in the presence of a heterogeneous causal effect. We will formally present the \texttt{$\ell_1$-TCL} framework.

%\vspace{-0.12in}

\subsection{\texttt{$\ell_1$-TCL} framework}\label{sec:part}
\begin{figure}[!htp]
%%\vspace{-0.1in}
\centerline{
\includegraphics[width = \textwidth]{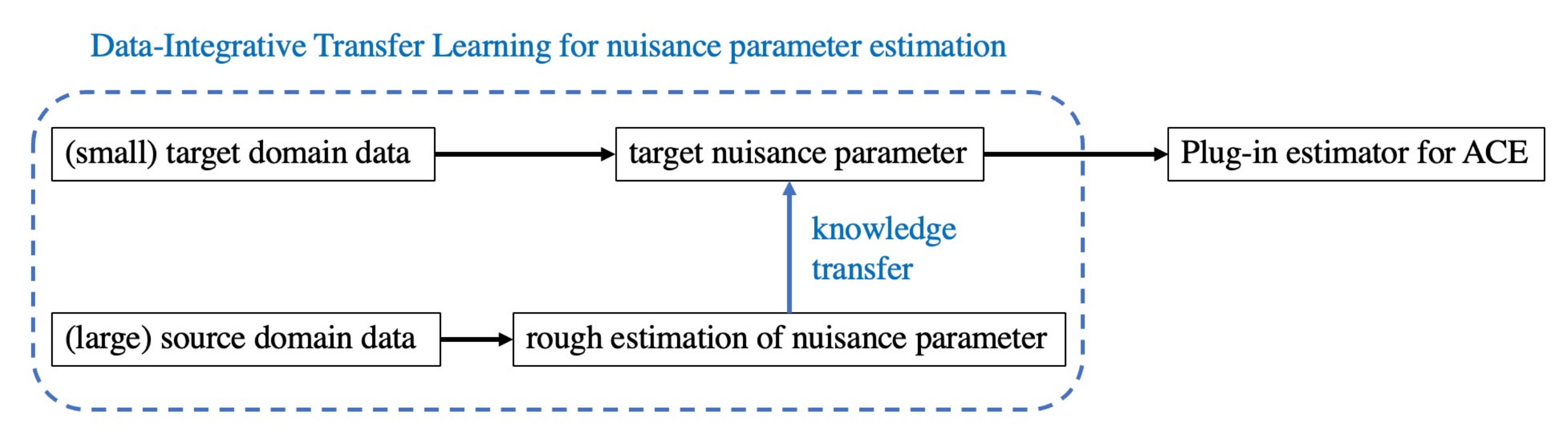} 
}
%\vspace{-0.1in}
\caption{Illustration of the general approach for TCL problem. In our proposed \texttt{$\ell_1$-TCL} framework, the nuisance parameter estimation stage leverages $\ell_1$ regularized TL, and the plug-in estimation stage considers IPW, OR and DR estimators.}
\label{fig:method_illus}
%%\vspace{-0.05in}
\end{figure}

Consider arbitrary parameterization of the nuisance model with finite-dimensional nuisance parameter $\theta \in \Theta$. Given dataset $\cD$, suppose the estimator for nuisance parameter can be obtained as:
$\hat \theta = \arg \min_{\theta \in \Theta} \cL(\theta;\cD),$ where $\cL$ is the loss function.
In our setup, the ground truth nuisance parameters differ across both domains, i.e., $\theta_\myt \not= \theta_\mys$. We assume their difference $\theta_\myt - \theta_\mys$ is sparse such that this difference can be estimated from the target domain using $\ell_1$ regularization to correct the bias of the rough estimator obtained from the source domain. 
Formally, the {\it nuisance parameter estimation stage} of our proposed \texttt{$\ell_1$-TCL} is given by:

%\vspace{-0.07in}
\begin{changemargin}{2.6cm}{.5cm}
\underline{{Rough estimation}}: $
\hat \theta_\mys = \arg \min_{\theta \in \Theta} \ \cL(\theta;\cD_\mys),$ 
\end{changemargin}
%\vspace{-0.15in}
\begin{changemargin}{2.8cm}{.5cm} 
$\ $\underline{{Bias correction}}: $ \
\hat \theta_\myt = \arg \min_{\theta \in \Theta} \ \cL(\theta;\cD_\myt) + \lambda \norm{\theta - \hat \theta_\mys}_1,$
\end{changemargin}
where $\cD_\mys = \{\cD_{i,\mys}, i = 1,\dots,n_\mys\}$ and $\cD_\myt = \{\cD_{i}, i = 1,\dots,n\}$ are the collections of source and target domain samples respectively, and $\lambda > 0$ is a tunable hyperparameter.
In the subsequent {\it plug-in estimation stage}, the IPW estimator \eqref{eq:ipw}, OR estimator \eqref{eq:or}, and/or DR estimator \eqref{eq:dr} are evaluated using the estimated nuisance parameters above to get the \texttt{$\ell_1$-TCL} estimate of the ACE. Please see a graphical illustration in Figure~\ref{fig:method_illus}.

\paragraph{Non-parametric approach based on neural networks.}
While there exist many recent efforts on improving robustness in causal inference, such as meta-learning \citep{westreich2010propensity} (notably, super learning \citep{pirracchio2015improving}), using NN to parameterize the nuisance models \citep{keller2015neural} is the most straightforward approach due to NN's superior model expressiveness. In the following, we will consider two recently developed NN architectures: Treatment-Agnostic Representation Network (TARNet) \citep{shalit2017estimating} and Dragonnet \citep{shi2019adapting}; we defer further details, such as their loss functions, to Appendix~\ref{appendix:NN_causal_model}. The implementation of the nuisance parameter estimation stage in our NN-based \texttt{$\ell_1$-TCL} is straightforward: the rough estimation step follows standard NN training using source domain data; in the bias correction step, similar to eq.~\eqref{eq:re-formulation} for GLM, we will estimate the sparse difference between the target and source NN weights with zero initialization. Complete details of our NN-based \texttt{$\ell_1$-TCL} can be found in Appendix~\ref{appendix:NN_training_detail}.

\paragraph{Application.} Heterogeneous causal effect has recently drawn increasing attention in causal inference, and there have been many popular machine learning approaches, such as meta-learning \citep{curth2021nonparametric} and heterogeneous transfer learning (i.e., TL under the heterogeneous covariate space setting) \citep{bica2022transfer}, applied to this problem. Typically, this problem is approached via the Conditional Average Treatment (or Causal) Effect (CATE) instead of ACE, i.e.,
\[\tau_{S} = \EE[ Y_1 | \boldsymbol{X} \in S ] - \EE[ Y_0 | \boldsymbol{X} \in S ],\]
which studies the causal effect within a sub-cohort of patients whose covariates lie in a targeted subset of the covariate space, i.e., $S \subset \cX$.

Built on the proposed \texttt{$\ell_1$-TCL}, we propose a \underline{Par}tition-then-\underline{T}ransfer approach, which we call \texttt{ParT}, for CATE estimation. Unlike the heterogeneous transfer learning approach by \citet{bica2022transfer}, which may require an additional dataset with a different covariate space, \texttt{ParT} handles the single dataset (or multiple datasets with the same covariate space) setting. Consider samples from a single dataset as in eq.~\eqref{eq:obs_target}, let $S_\myt \subset \cX$ be the target subset, and the goal is to estimate $\tau_{S_\myt}$. \texttt{ParT} first partitions the covariate space into $\cX = S_\mys \cup S_\myt$, resulting in a source-target domain partition: $\cD_\mys = \{\cD_{i}: \boldsymbol{x}_i \in S_\mys\}$ and $\cD_\myt = \{\cD_{i}: \boldsymbol{x}_i \in S_\myt\}$. Then, \texttt{$\ell_1$-TCL} can be readily applied to leverage knowledge gained from $\cD_\mys$ to help estimate the CATE (or target domain ACE) $\tau_{S_\myt}$.

In practice, the target subset $S_\myt$ is typically defined through a binary (or categorical) covariate, resulting in a natural covariate space partition based on the corresponding labels. As the partitioned domains come from the same dataset, it is reasonable to assume the underlying treatment assignment mechanisms are similar across both domains and therefore, our \texttt{$\ell_1$-TCL} is applicable. Nevertheless, it is important to develop a principled approach to determine whether the knowledge is transferable from the partitioned source domain. In particular, when covariate space is partitioned via a categorical covariate with three or more labels, the problem is cast as a multiple-source TL problem since there are multiple source domains; in this case, it is important to determine which source domain to include in the transfer learning. Indeed, \citet{tian2022transfer} studied this multiple-source TL problem using the $\ell_1$ regularized approach considered in this work, which we believe can help establish theoretical guarantee for \texttt{ParT}; however, this is out of the scope of the current study, and we leave this for future discussion. Next, we use a pseudo-real-data experiment to show the effectiveness of \texttt{ParT}.

\subsection{Pseudo-real-data experiment}\label{sec:nn_exp}

%\vspace{-0.12in}

In this experiment, we aim to show the effectiveness of \texttt{ParT} for CATE estimation, which also demonstrates the good performance of its building block, i.e., our \texttt{$\ell_1$-TCL} framework, by comparing with baseline frameworks.
As ground truth causal effects are inaccessible in most real studies, we consider a commonly used pseudo-real-dataset, i.e., the Infant Health and Development Program (IHDP) dataset \citep{brooks1992effects,hill2011bayesian}. It includes 747 subjects (139 treated and 608 control), with six continuous and 19 categorical covariates (of which eighteen are binary). 
We randomly pick one binary covariate (denoted by $X_{\rm par}$) and assign subjects with labels 0 and 1 to source and target domains, respectively, resulting in $n_\mys = 546, n = 201$. The goal is to study the ACE in the target domain, or the CATE for the subjects with $X_{\rm par} = 1$. Due to space consideration, additional details for the dataset, configurations, training, and results are deferred to Appendix~\ref{appendix:add_ihdp_result}.

%\vspace{-0.1in}

\paragraph{Comparison with baselines.} 
We compare \texttt{$\ell_1$-TCL} framework with two baseline learning frameworks: \texttt{TO-CL} introduced above and the ``warm-start'' TCL baseline (\texttt{WS-TCL}) by \citet{kunzel2018transfer}, which used the estimated NN weights in the source domain as the warm-start of the subsequent target domain NN training. For each framework, the nuisance model for PS and OR is either Dragonnet or TARNet with hyperparameters selected based on minimum average NN regression loss on a randomly selected validation target domain dataset; the estimated nuisance parameters are subsequently plugged into IPW, OR, and DR estimators to get the estimated ACEs. Code is available at \url{https://github.com/SongWei-GT/L1-TCL}.

\begin{table}[!htp]
\centering
%\vspace{-.05in}
\caption{Mean and standard deviation of absolute errors of estimated ACEs over 50 trials using IHDP dataset. The primary goal is to compare three learning frameworks: we can observe that TL can help improve ACE estimation accuracy for all ACE estimators (highlighted in green for each column) and our proposed \texttt{$\ell_1$-TCL} yields the best in-sample and out-of-sample results (highlighted in bold font).}
\label{tab:idx4_1000}
\vspace{.1in}
\resizebox{.85\textwidth}{!}{%
\begin{tabular}{lcccccc}
%\multicolumn{7}{c}{{Hyperparameter selected based on (minimum) NN regression loss on validation dataset}} \\ 
\toprule
\multirow{2}{*}{In-sample}    & \multicolumn{3}{c}{Dragonnet} & \multicolumn{3}{c}{3-Headed Variant of TARNet} \\
    & IPW       & OR      & DR      &      IPW       & OR      & DR\\
\cmidrule(l){2-4}
\cmidrule(l){5-7}
\texttt{TO-CL}  & $6.668_{(11.234)}$ & $0.573_{(0.579)}$ & $0.612_{(0.711)}$ & $4.307_{(4.408)}$ & $0.5_{(0.437)}$ & $0.415_{(0.349)}$ \\ 
\texttt{WS-TCL} & 
  $5.28_{(12.068)}$  &  $0.49_{(0.445)}$  &  $0.503_{(0.7)}$  & $2.815_{(3.203)}$ &  $0.41_{(0.373)}$ & $0.337_{(0.273)}$ \\ 
\texttt{$\ell_1$-TCL}  & 
\cellcolor[HTML]{C6EFCE}{\color[HTML]{006100} $4.724_{(11.038)}$ }
& \cellcolor[HTML]{C6EFCE}{\color[HTML]{006100} $0.386_{(0.406)}$ }& \cellcolor[HTML]{C6EFCE}{\color[HTML]{006100} $0.404_{(0.443)}$ }& \cellcolor[HTML]{C6EFCE}{\color[HTML]{006100}  $2.804_{(3.149)}$ } & \cellcolor[HTML]{C6EFCE}{\color[HTML]{006100} $0.394_{(0.434)}$ } & \cellcolor[HTML]{C6EFCE}{\color[HTML]{006100}  $\boldsymbol{0.289}_{(0.238)}$ } \\ 
\cmidrule(l){1-7}
  \multirow{2}{*}{Out-of-sample}    & \multicolumn{3}{c}{Dragonnet} & \multicolumn{3}{c}{3-Headed Variant of TARNet} \\
    & IPW       & OR      & DR      &      IPW       & OR      & DR\\
\cmidrule(l){2-4}
\cmidrule(l){5-7}
\texttt{TO-CL}  & $28.568_{(43.776)}$ & $0.639_{(0.713)}$ & $2.095_{(3.909)}$ & $4.873_{(5.842)}$ & $0.603_{(0.594)}$ & $0.36_{(0.301)}$ \\ 
\texttt{WS-TCL} &  $16.416_{(28.332)}$  &  $0.54_{(0.533)}$  &  $1.203_{(2.063)}$  & \cellcolor[HTML]{C6EFCE}{\color[HTML]{006100}  $4.182_{(5.588)}$ } &   $0.469_{(0.447)}$  &   ${0.324}_{(0.282)}$  \\ 
\texttt{$\ell_1$-TCL}  & \cellcolor[HTML]{C6EFCE}{\color[HTML]{006100} $13.872_{(24.138)}$} & \cellcolor[HTML]{C6EFCE}{\color[HTML]{006100} $0.425_{(0.454)}$} & \cellcolor[HTML]{C6EFCE}{\color[HTML]{006100} $0.747_{(0.94)}$} & $4.193_{(5.502)}$ & \cellcolor[HTML]{C6EFCE}{\color[HTML]{006100} $0.42_{(0.439)}$ }& \cellcolor[HTML]{C6EFCE}{\color[HTML]{006100} $\boldsymbol{0.301}_{(0.259)}$ }\\ 
\bottomrule
\end{tabular}%
%\vspace{-.1in}
}
\end{table}

We report both in-sample (i.e., training and validation target datasets) and out-of-sample (i.e., testing target dataset) absolute estimation errors over 1000 trials in Table~\ref{tab:idx4_1000}, from which we can observe that: (i) transfer learning helps improve estimation accuracy for all {\it ACE estimators} (we will call a specific nuisance model coupled with a specific plug-in estimator as an ACE estimator); (ii) in almost all cases, our proposed \texttt{$\ell_1$-TCL} outperforms the existing \texttt{WS-TCL} approach; (iii) most importantly, the best results (highlighted in bold fonts) are given by our proposed \texttt{$\ell_1$-TCL} framework.
Another interesting finding is that the plug-in estimator based on the OR model typically performs better than the PS model-based IPW estimator, potentially due to severe model misspecification (or poor tuning) of the NN-based PS model. This is consistent with the observation noted by \citet{shi2019adapting}, who only considered OR estimator in their experiments, and may explain why NN classification cross entropy (CE) loss and mean squared error (MSE) do not serve as good hyperparameter selection criteria in our task; those results are presented in Table~\ref{tab:idx4} for completeness. To further validate the effectiveness of our \texttt{$\ell_1$-TCL} (as well as our \texttt{ParT}), we report results for source-target domain partition based on another binary covariate (which yields $n_\mys = 642$ and $n = 105$) in Tables~\ref{tab:idx8} and \ref{tab:idx8_median}.

The results above established the superior performance of the 3-Headed Variant of TARNet coupled with the DR estimator for ACE estimation for the IHDP dataset in our setting; from now on, we will continue our experiments based on this ACE estimator.

\paragraph{An alternative hyperparameter selection criterion: Standard Mean Difference.} 
Recently, \citet{machlanski2023hyperparameter} found that proper hyperparameter tuning can sometimes close the performance gap among different ACE estimators.
The challenge is that a ``golden standard'', such as prediction accuracy in classic supervised learning, is largely missing since we cannot evaluate the causal effect estimation accuracy from data due to unobserved counterfactual outcomes. Unfortunately, there is no good solution beyond the recent empirical analysis of some nuisance model prediction performance metrics (which is what we use in the above experiments) as the selection criteria \citep{athey2016recursive,curth2023search,machlanski2023hyperparameter}. 

Here, we ``propose'' one additional solution by studying the covariate distribution balances via Standard Mean Difference (SMD); indeed, SMD has been used to access the Goodness-of-Fit of PS models in causal inference in various applications, such as \citet{zhang2019balance}. Here, we defer the definition of SMD to Appendix~\ref{appendix:SMD} and report the results on comparison among different learning frameworks with SMD as the hyperparameter selection metric in Table~\ref{table:SMD}.

\begin{table}[!h]
\centering
\resizebox{.5\textwidth}{!}{%
\begin{tabular}{lcccccc} 
\toprule
    &     \texttt{TO-CL}       &  \texttt{WS-TCL}     & \texttt{$\ell_1$-TCL} \\
\cmidrule(l){2-4} 
In-sample  &   $0.415_{(0.385)}$ & $0.337_{(0.273)}$ & $\boldsymbol{0.277}_{(0.222)}$  \\ 
Out-of-sample  &   $0.421_{(0.467)}$  & $0.324_{(0.282)}$ & $\boldsymbol{0.301}_{(0.252)}$  \\ 
\bottomrule
\end{tabular}%
}
%\vspace{-0.1in}
\caption{Absolute error mean and standard deviation for SMD as the hyperparameter selection criterion; the ACE estimator is chosen as the 3-Headed Variant of TARNet coupled with the DR estimator, which has the best performance as shown in Table~\ref{tab:idx4_1000}. The above table further establishes the superior performance of our \texttt{$\ell_1$-TCL} framework. Moreover, as the above table shows the overall best performance compared with results in Table~\ref{tab:idx4_1000}, it also established the potential of SMD as the hyperparameter selection criterion or Goodness-of-Fit score.}\label{table:SMD}
%\vspace{-0.15in}
\end{table}

\subsection{$\ell_1$-\texttt{TCL} with NN-based nuisance model on real-data}
The above experiment indeed favors SMD as the Goodness-of-Fit score to help select hyperparameters. In the last experiment, we report SMD for the real-data experimental results in Table~\ref{tab:real_data}; we also apply the NN-based methods on real-data, and report the point estimates of the ACE and the SMD; the aforementioned results are summarized in Table~\ref{table:real_SMD_NN}, reaffirming that $\ell_1$-\texttt{TCL} can help obtain a reasonable causal effect estimate (i.e., with the best Goodness-of-Fit score) from limited target data. As shown in this table, the results selected by SMD support inhibiting treatment effect, i.e., vasopressor therapy
does decrease the risk of 28-day mortality in sepsis patients, agreeing with established results in medical literature \citep{avni_vasopressors_2015}.

\begin{table}[!h]
\centering
\resizebox{0.75\textwidth}{!}{%
\begin{tabular}{lccccccc} 
\toprule 
& \multicolumn{3}{c}{GLM-Sigmoid-IPW} &  & \multicolumn{3}{c}{3 Headed-TARNet-SMD-DR} \\
    &       \texttt{TO-CL}       &     \texttt{Merge-CL}  & \texttt{$\ell_1$-TCL} &  \quad  &   \texttt{TO-CL}       &  \texttt{WS-TCL}     & \texttt{$\ell_1$-TCL} \\
\cmidrule(l){2-4} \cmidrule(l){6-8}
Point estimate  &   $0.12$ & $0.082$ & $-0.011$  &  \quad\quad\quad & $-12.612$ & $-11.851$ & $-9.676$ \\ 
SMD  &   $0.242$ & $0.268$ & $\boldsymbol{0.225}$  &  \quad &   ${0.1731}$ & $0.1721$ & $\boldsymbol{0.1720}$\\ 
%\cmidrule(l){1-4} \cmidrule(l){5-8}
\bottomrule
%\rule{0pt}{0.5ex} \\ 
\end{tabular}%
}
\caption{SMDs of the point estimates in Table~\ref{tab:real_data} (left), and the results using the NN-based method (right). In the first row, we specify nuisance model parameterization (with additional details such as link function and hyperparameter selection criterion) and the plug-in estimator. The results selected by SMD indicate an inhibiting effect from treatment to mortality, agreeing with both common sense and existing medical literature.}\label{table:real_SMD_NN} 
\end{table}

\section{Conclusion and Discussion}

This work proposes a framework to adapt transfer learning to causal inference, which can address the limited data issue faced in many real applications. The proposed \texttt{$\ell_1$-TCL} not only enjoys strong theoretical guarantees but also can be used along with many recently developed non-parametric ACE estimators. Most importantly, it shows great promise in estimating the treatment effect in real applications, especially in the medical setting, as evidenced by our experimental results. Moreover, the successful application of $\ell_1$ regularized TL in causal inference could inspire a potential research direction: Recently, statistics literature has witnessed a surge of theoretically grounded TL approaches due to their empirical success, and these principled approaches could be readily adapted to the novel TCL problem; for example, TL for non-parametric regression \citep{cai2022transfer,lin2023source} and high-dimensional Gaussian graphical models \citep{li2022transfer} might be applied to causal effect estimation and causal graph discovery \citep{spirtes2000constructing,pearl2009causality}, respectively. In addition, the $\ell_1$ regularized TL discussed in this work might be extended to convex causal graph discovery using GLMs \citep{10366499}.
Lastly, it would be rather interesting to extend the \texttt{ParT} framework for CATE estimation (introduced in Section~\ref{sec:part}) to ``Partition-Transfer-Reconcile'': Inspired by the forecasting reconciliation \citep{wickramasuriya2018optimal}, one might be able to develop a reconciliation constraint on the CATEs in the subgroups subject to the ACE of the whole dataset, estimated using some sophisticated models (e.g., NN-based model).

\bibliographystyle{plainnat}  
\bibliography{ref}

\newpage

% \begin{center}
% {\LARGE\bf Supplementary Materials of Transfer Causal Learning: Causal Effect Estimation with Knowledge Transfer}
% \end{center}

\appendices

%\begin{KeepFromToc}
\addcontentsline{toc}{section}{Appendix} % Add the appendix text to the document TOC
\part{\centering \Large Appendix of \mytitle} % Start the appendix part

\topskip0pt
%\vspace*{\fill}

\parttoc % Insert the appendix TOC
%\faketableofcontents
%\tableofcontents
%\end{KeepFromToc}

%\vfill
%\vspace*{\fill}

\section{Extended Literature Survey}\label{sec:literature}

\subsection{Background on transfer learning}
Transfer learning \citep{torrey2010transfer} has received increasing attention due to its empirical success in various fields, ranging from machine learning problems, such as natural language processing \citep{daume2007frustratingly}, recommendation system \cite{pan2013transfer} and computer vision \citep{tzeng2017adversarial}, to science problems, such as predictions of protein localization \citep{mei2011gene}, biological imaging diagnosis \citep{shin2016deep}, integrative analysis of ``multi-omics'' (e.g., genomics) data \citep{sun2016integrative,hu2019statistical,wang2019horizontal}, cancer image classification \citep{hosny2018skin,sevakula2018transfer}, drug sensitivity prediction \citep{turki2017transfer} and discovery \citep{bastani2021predicting}, and so on. 
Based on whether or not the target and source domains, as well as the target and source tasks, are the same, transfer learning problems can be categorized into several sub-problems \citep{pan2010survey}; in this study, we focus on ``Inductive Multi-Task Transfer Learning'' and our $\ell_1$ regularization-based approach can be categorized as ``Transferring Knowledge of Parameters''. Our work only leverages a particular transfer learning technique and we refer readers to \citet{pan2010survey,weiss2016survey,zhuang2020comprehensive} for comprehensive surveys on transfer learning.

\subsection{Developments of $\ell_1$ regularized transfer learning approaches}
The idea of using $\ell_1$ regularization to develop a theoretically grounded TL approach could date back to \citet{evgeniou2004regularized}, who considered a support vector machine with parameter decomposed as a summation of a shared term and a task-specific term and proposed a learning algorithm by imposing $\ell_1$ regularization on the task-specific terms in all domains. Recently, this idea was applied to GLM by \citet{bastani2021predicting}, and this seminal work has motivated several follow-up studies: \citet{tian2022transfer} extended this work to multi-source TL problems,
\citet{li2022transfer} proved minimax optimality under liner regression setting and later on showed minimax rate of convergence for high-dimensional GLM estimation \citep{li2023estimation}, and so on.
Our work follows this line of study and adapts the $\ell_1$ regularized TL approach proposed by \citet{bastani2021predicting} to develop a theoretically grounded method for TCL, but the theoretical results may be strengthened using those aforementioned recent developments; notably, the principled method for multi-source TL problem developed by \citet{tian2022transfer} may help our \texttt{ParT} approach.   
Most importantly, it is important to recognize that our work points out a new direction on leveraging recently developed principled TL methods to contribute to the TCL problem.

\subsection{Connections between transfer learning and causal inference}
There have been works leveraging causal inference to help with TL problems, such as domain adaption, by exploring the invariant causal relationships between both domains, which is referred to as the causal transfer learning problem \citep{rojas2018invariant}.
While the causal transfer learning problem has been studied from both empirical \citep{zhang2015multi,magliacane2018domain,yang2021learning} and theoretical \citep{rojas2018invariant,chen2021domain} perspectives, the reverse study on adapting TL techniques to causal inference (i.e., our proposed TCL problem) only starts to attract some attention recently.
In particular, a line of research \citep{yang2020combining,wu2022transfer,hatt2022combining} has been focused on handling the unmeasured confounding variables in the target observational datasets with the help of unconfounded randomized experimental source domain data, where, in its nature, only the TL approaches for heterogeneous covariate space settings are applicable. 
However, such experimental data is not always available in reality, and the fundamental problem of estimating causal effects under the classic no unmeasured confounding assumption receives little attention; existing works along this direction include the aforementioned ``warm-start'' knowledge transfer approach under our TCL setting \citep{kunzel2018transfer} and a special neural network architecture designed based on the shared covariate space and the domain-specific covariate spaces \citep{bica2022transfer}. Here, we not only provide a theoretically grounded approach for the TCL problem but also use numerical evidence to show our proposed \texttt{$\ell_1$-TCL} outperforms the existing warm-start method.

\section{Additional Background Knowledge}

\subsection{Additional details on the potential outcome framework}\label{appendix:causal_background}
The gold-standard approach to estimating the causal effect is randomized controlled trials (RCT), where subjects are randomized to receive treatment or placebo (i.e., the control group). However, RCT is unethical in most studies, such as the medical study. 
Therefore, the main question is how to estimate causal effect from observational data. 

In this work, we approach this problem under the potential outcome framework \citep{rubin1974estimating}
Let us recall the notations we used: random vector $\boldsymbol{X} \in \RR^d$ represents covariates measured prior to receipt of treatment, r.v. $Z \in \{0,1\}$ is treatment indicator, r.v. $Y$ is the observed outcome, which is defined using potential outcomes $Y_0$ and $Y_1$: $Y = Y_1 Z + (1-Z) Y_0.$ The ACE (or average treatment effect) is the estimand in this study and is defined as: $\tau = \EE[Y_1] - \EE[Y_0].$

Apparently, observing $Y_0$ and $Y_1$ simultaneously is impossible, making it a tempting choice to estimate $\EE[Y_0]$ and $\EE[Y_1]$ using the sample average outcome in the control and treatment group and take their difference as an estimate of the ACE. Unfortunately, the problem is $\EE[Y|{Z} = 0] = \EE[Y_0|{Z} = 0] \not= \EE[Y_0]$ and $\EE[Y|{Z} = 1] = \EE[Y_1|{Z} = 1] \not= \EE[Y_1]$, as the treatment ${Z}$ is typically not statistically independent from $(Y_0, Y_1)$ due to some pre-treatment covariates in $\boldsymbol{X}$. Such a problem is also referred to as ``selection bias'', which is illustrated in Figure~\ref{fig:selection_bias_illus} below.

\begin{figure}[!htp]
%%\vspace{-0.1in}
\centerline{
\includegraphics[width = .6\textwidth]{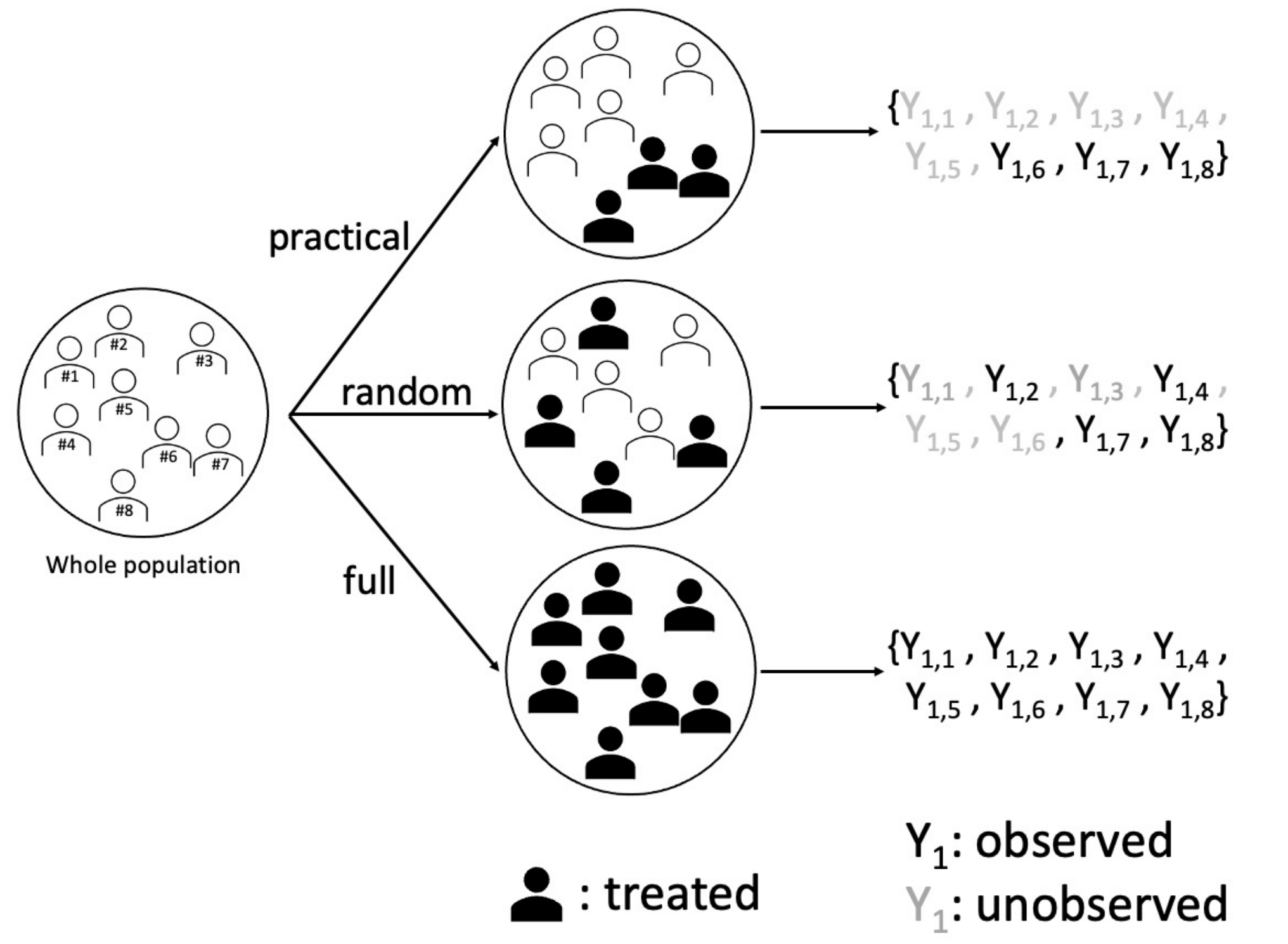} 
}
%\vspace{-0.1in}
\caption{Illustration of the selection bias. In practice (top), especially in the observational study, the treatment assignment is typically dependent on pre-treatment covariates $\boldsymbol{X}$, making the selected (or observed) treatment cohort NOT independent of the outcome variable. As a result, the selected cohort is not ``representative'' of the whole population, and inference based on such a selected cohort will typically be biased.}
\label{fig:selection_bias_illus}
%%\vspace{-0.05in}
\end{figure}

In observational study, although $(Y_0, Y_1) \independent Z$ is unlikely to hold, it may be possible to identify those pre-treatment covariates that related to (or can affect) both potential outcome and treatment, referred to as ``confounders''. If we assume the covariate vector $\boldsymbol{X}$ contains all such confounders, we would have
$(Y_0, Y_1) \independent Z \mid \boldsymbol{X},$
which is referred to as ``no unmeasured confounders'' or ignorability assumption \citep{robins2000marginal}.
Under this assumption, we shall have 
\begin{equation}\label{eq:OR_unbiased}
\begin{split}
    \EE[Y | {Z} = 1] &= \EE\{\EE[Y | {Z} = 1, \boldsymbol{X}]\}=\EE\left\{\EE\left[Y_1 | {Z} = 1, \boldsymbol{X}\right]\right\}\\ \noalign{\vskip1pt}
    &=\EE\left\{\EE\left[Y_1 | \boldsymbol{X}\right]\right\}=\EE\left[Y_1\right].
\end{split}
\end{equation}
Similarly, \[\EE[Y | {Z} = 0] = \EE\{\EE[Y | {Z} = 0, \boldsymbol{X}]\}=\EE\left[Y_0\right].\] The above observations actually motivate the unbiased estimator using the outcome regression model, i.e., the OR estimator \eqref{eq:or}. Under the no unmeasured confounding assumption, the ACE $\tau$ is identifiable from observational data.

The propensity score $e(\boldsymbol{X}) = \PP({Z} = 1 | \boldsymbol{X})$ is the probability of treatment given covariates, which specifies the treatment assignment mechanism. \citet{rosenbaum1983central} showed that
$(Y_0, Y_1) \independent Z \mid e(\boldsymbol{X}),$
which implies that $\EE\left[I({Z} = 1) | Y_1, \boldsymbol{X}\right] = e(\boldsymbol{X})$ (here, $I(A) = 1$ if event $A$ is true and 0 otherwise). Therefore, we will have
\begin{equation}\label{eq:IPW_unbiased}
\begin{split}
        \EE\left[\frac{Z Y}{e(\boldsymbol{X})}\right]
    &= \EE\left\{\EE\left[\frac{I({Z} = 1) Y_1}{e(\boldsymbol{X})} \ \bigg| \ Y_1, \boldsymbol{X}\right]\right\} \\
    &= \EE\left\{\frac{Y_1}{e(\boldsymbol{X})} \EE\left[I({Z} = 1) | Y_1, \boldsymbol{X}\right]\right\}
    =\EE[Y_1].
\end{split}
\end{equation}
Similarly, \[\EE\left[\frac{(1-Z) Y}{1-e(\boldsymbol{X})}\right] =\EE[Y_0].\]
The above observations actually motivate the application of IPW \citep{horvitz1952generalization} for ACE estimation and show that IPW estimator \eqref{eq:ipw} is unbiased under correct PS model specification.

An alternative unbiased estimator uses the (potential) outcome regression model: \[m_{z}(\boldsymbol{X}) = \EE[{Y}_{z}| \boldsymbol{X}], \quad z \in \{0,1\}.\]
Given samples \eqref{eq:obs_target}, for $z \in \{0,1\}$, let $n_{z} = \#\{i: z_{i} = z\}$ and $\hat m_{z}(\boldsymbol{x}_{i})$ be the fitted potential outcome for $i$-th subject, the OR estimator for ACE is given by:
\begin{equation}\label{eq:or}
    \hat \tau_{\rm OR} = \frac{1}{n_{1}} \sum_{z_{i} = 1} {\hat m_1(\boldsymbol{x}_{i})} - \frac{1}{n_{0}} \sum_{z_{i} = 0} {\hat m_0(\boldsymbol{x}_{i})}.
\end{equation}

One common drawback of both IPW and OR estimators is that they require correct specification of the PS and OR models respectively, which is challenging in practice. To fix this issue, an augmented IPW estimator (also known as DR estimator) is proposed \citep{robins1994estimation,rotnitzky1998semiparametric,scharfstein1999adjusting} --- 
The main idea is that by incorporating an augmented term (which is related to the OR model) in IPW, the estimator will be doubly robust. To be precise, given samples \eqref{eq:obs_target}, the DR estimator for ACE is defined as:
\begin{equation}\label{eq:dr}
    \begin{split}
\hat{\tau}_{\rm DR} & =\frac{1}{n} \sum_{i=1}^n\left[\frac{z_i y_i}{\hat{e}\left(\boldsymbol{x}_i\right)}-\frac{z_i-\hat{e}\left(\boldsymbol{x}_i\right)}{\hat{e}\left(\boldsymbol{x}_i\right)} \hat{m}_1\left(\boldsymbol{x}_i\right)\right]-\frac{1}{n} \sum_{i=1}^n\left[\frac{\left(1-z_i\right) y_i}{1-\hat{e}\left(\boldsymbol{x}_i\right)}+\frac{z_i-\hat{e}\left(\boldsymbol{x}_i\right)}{1-\hat{e}\left(\boldsymbol{x}_i\right)} \hat{m}_0\left(\boldsymbol{x}_i\right)\right] \\ \noalign{\vskip1pt}
& =\frac{1}{n} \sum_{i=1}^n\left[\hat{m}_1\left(\boldsymbol{x}_i\right)+\frac{z_i\left(y_i-\hat{m}_1\left(\boldsymbol{x}_i\right)\right)}{\hat{e}\left(\boldsymbol{x}_i\right)}\right]-\frac{1}{n} \sum_{i=1}^n\left[\hat{m}_0\left(\boldsymbol{x}_i\right)+\frac{\left(1-z_i\right)\left(y_i-\hat{m}_0\left(\boldsymbol{x}_i\right)\right)}{1-\hat{e}\left(\boldsymbol{x}_i\right)}\right].
    \end{split}
\end{equation}
Notice that:
\begin{equation*}
    \begin{split}
        \EE[Y_1] & =\mathbb{E}\left[\frac{Z Y}{e(\boldsymbol{X})}-\frac{Z-e(\boldsymbol{X})}{e(\boldsymbol{X})} m_1(\boldsymbol{X})\right] =\mathbb{E}\left[m_1\left(\boldsymbol{X}\right)+\frac{Z\left(Y-m_1\left(\boldsymbol{X}\right)\right\}}{e\left(\boldsymbol{X}\right)}\right], \\ \noalign{\vskip1pt}
\EE[Y_0] & = \mathbb{E}\left[\frac{(1-Z) Y}{1-e(\boldsymbol{X})}+\frac{Z-e(\boldsymbol{X})}{1-e(\boldsymbol{X})} m_0(\boldsymbol{X})\right] = \mathbb{E}\left[m_0\left(\boldsymbol{X}\right)+\frac{\left(1-Z\right)\left(Y-m_0\left(\boldsymbol{X}\right)\right)}{1-e\left(\boldsymbol{X}\right)}\right].
    \end{split}
\end{equation*}
Therefore, the DR estimator is unbiased when either the PS model or the OR model is correctly specified, i.e., it is doubly robust.

Lastly, those aforementioned estimators for ACE all have nice theoretical properties; see, e.g., \citet{wooldridge2002inverse,wooldridge2007inverse} for the theory of IPW estimator and \citet{robins1994estimation,bang2005doubly} for the theory of DR estimator. There are also other approaches to estimate causal effects using propensity score, such as matching; see \citet{lunceford2004stratification} for a nice survey on the use of propensity scores in causal inference and \citet{yao2021survey} for a recent comprehensive survey on causal inference.

\subsection{Neural network-based nuisance models for causal effect estimation}\label{appendix:NN_causal_model}

In this part, we briefly review the aforementioned NN-based approaches for ACE estimation: TARNet \citep{shalit2017estimating} and Dragonnet \citep{shi2019adapting}, which can both be categorized as representation learning methods according to \citet{yao2021survey}.

\paragraph{TARNet.}
Consider covariate vector, treatment and observed outcome tuple $(\boldsymbol{X}, {Z}, {Y})$ tuple with realizations $\cD = \{(\boldsymbol{x}_{i}, z_{i}, y_{i}), \ i = 1,\dots,n\}$.
TARNet finds a representation of the covariates, denoted by $\Phi(\boldsymbol{x}_{i})$, which maps the covariate vector onto a representation space, and hypothesis of the potential outcome variable, denoted by $m_{z_i}(\Phi(\boldsymbol{x}_{i}))$, simultaneously by minimizing  the following regularized objective function: 
\begin{equation*}
\begin{split}
            \min _{\substack{m_0, m_1, \Phi}} \ \cL_{\rm TAR}(m_0, m_1, \Phi; & \ \cD) =  \frac{1}{n} \sum_{i=1}^n w_i \tilde L\left(m_{z_i}(\Phi(\boldsymbol{x}_{i})); y_i\right)    \\
        & + \lambda_{\rm CPLX} {\rm R}(m_0,m_1) + \lambda_{\rm BAL} \operatorname{IPM}\left(\left\{\Phi\left(\boldsymbol{x}_{i}\right)\right\}_{i: z_i=0},\left\{\Phi\left(\boldsymbol{x}_{i}\right)\right\}_{i: z_i=1}\right),
\end{split}
\end{equation*}
where ${\rm R}(\cdot,\cdot)$ represents the model complexity, $\operatorname{IPM}(\cdot,\cdot)$ represents the Integral Probability Metric (IPM) \citep{sriperumbudur2012empirical}, such as the Maximum Mean Discrepancy and the Wasserstein Distance, evaluated on two empirical distributions defined by two collections of data-points on the representation space, and weights $w_i$'s compensate for the difference in treatment group size and are defined as follows:
\[w_i = \frac{z_i}{2 u}+\frac{1-z_i}{2(1-u)}, \quad i = 1,\dots,n, \quad u = \frac{1}{n} \sum_{i=1}^n z_i.\]
The loss function $\tilde L$ for the network training is decomposed into two terms, i.e., $\tilde L\left(m_{z}(\Phi(\cdot)); \cdot\right),$ $z \in \{0,1\}$, which correspond to the control and treatment groups, respectively. The weights for the treatment and control functions are updated only if the sample belongs to that group. Either the MSE or log-loss can be used as $\tilde L$, depending on whether the outcome variable is continuous or binary. Most importantly, to handle the problem of variance arising from treatment imbalance, the TARNet objective includes the empirical IPM to upper bound this variance. Hyperparameter $\lambda_{\rm BAL} > 0$ controls the trade-off between outcome regression model fitting and the treatment-and-control distribution balances. When $\lambda_{\rm BAL} = 0$, it corresponds to the TARNet; otherwise, it corresponds to the Counterfactual Regression.

\paragraph{Dragonnet.}
Similarly, Dragonnet creates a shared representation of the covariates can be used to predict the treatment and potential outcomes. It uses an NN for the shared representation followed by two NNs used for predicting potential outcomes of the treatment and control groups, respectively. However, instead of using an IPM layer, \citet{shi2019adapting} incorporated a mapping layer for the propensity score, which is named the ``propensity score head'' and denoted by $e(\cdot)$, to connect the shared representation of the covariates with the estimated propensity scores. To be precise, the objective function is:
\begin{equation}\label{eq:dragon_obj}
\begin{split}
            \min _{\substack{\theta}} \ \cL_{\rm Dragon}(\theta; \cD) = \frac{1}{n} \sum_{i=1}^n \underbrace{\left(m_{z_i}(\theta; \boldsymbol{x}_{i}) - y_i\right)^2}_{\text{\rm {NN regression loss}}}   +  \lambda_{\rm BAL} \underbrace{\operatorname{CE}\left(e(\theta; \boldsymbol{x}_{i}),z_i\right)}_{\text{\rm {NN classification CE loss}}},
\end{split}
\end{equation}
where $\operatorname{CE}(\cdot,\cdot)$ is the binary classification cross-entropy loss, and $\lambda_{\rm BAL} > 0$ is a tunable hyperparameter controlling the trade-off between outcome regression model fitting and the treatment-and-control distribution balances.

For further details of TARNet and Dragonnet, we refer readers to the original papers. In our numerical experiments, we use the \hypertarget{gitlink}{open source implementation}\footnote{The are two implementations on GitHub, one is from the Dragonnet paper author: \url{https://github.com/claudiashi57/dragonnet}, and the other is a reproduction of the results using PyTorch: \url{https://github.com/alecmn/dragonnet-reproduced}.} of TARNet and Dragonnet on the IHDP dataset and readers can find further implementation details therein.

\section{Non-Asymptotic Recovery Guarantee for TLIPW}\label{appendix:TLIPW}

We begin our theoretical analysis with the TLIPW estimator. We will first prove the non-asymptotic upper bound on the $\ell_1$ regularized TL estimator for PS model \eqref{eq:step2} and then plug it into the error bound for unbiased IPW estimator \eqref{eq:ipw} to get the final recovery guarantee for the TLIPW estimator.

\subsection{Guarantee for PS model estimation with knowledge transfer}
Let us begin with the necessary assumptions:
\begin{assumption}\label{A0}
    The covariates in both target and source domains are uniformly bounded, i.e., there exists $M_X> 0$ such that $\norm{\boldsymbol x_i}_{\infty} \leq M_X,  i = 1,\dots,n,$ and $\norm{\boldsymbol x_{i,\mys}}_{\infty} \leq M_X,  i = 1,\dots,n_\mys$. 
\end{assumption}

The above assumption is a slightly different from the ``standardized design matrix'' assumption in \citet{bastani2021predicting}, which requires the squared matrix $F$-norms of design matrices $(\boldsymbol x_{1},\dots,\boldsymbol x_{n})^\T$ and $(\boldsymbol x_{1, \mys},\dots,\boldsymbol x_{n_{\mys}, \mys})^\T$ to be $n$ and $n_{\mys}$, respectively. However, we will see they serve the same purpose when proving Lemma~\ref{lma:TL-beta} (to be presented).
Denote the sample covariance matrices as follows:
\begin{equation}\label{eq:cov}
    \Sigma = \frac{1}{n} \sum_{i = 1}^n \boldsymbol x_{i} \boldsymbol x_{i}^\T \in \RR^{n \times n}, \quad \Sigma_\mys = \frac{1}{n_\mys} \sum_{i = 1}^{n_\mys} \boldsymbol x_{i,\mys} \boldsymbol x_{i,\mys}^\T \in \RR^{n_\mys \times n_\mys}.
\end{equation}

\begin{assumption}\label{A1}
The source domain sample covariance matrix $\Sigma_\mys$ is positive-definite (PD); in particular, we assume that $\Sigma_\mys$ has minimum eigenvalue $\psi > 0$.
\end{assumption}

Here, Assumption~\ref{A1} ensures we can faithfully recover $\beta_\mys$ using MLE from the source domain data, and this assumption is mild when $n_\mys > d$, which is satisfied under our considered regime \eqref{eq:regime}.

\begin{definition}[Compatibility Condition \citep{bastani2021predicting}] The compatibility condition with constant $\phi > 0$ is met for the index set $\cI \subset \{1,\dots,d\}$ and the matrix $\Sigma \in \RR^{d \times d}$, if for all $u \in \RR^d$ satisfying $\norm{u_{\cI^{\rm c}}}_1 \leq 3 \norm{u_{\cI}}_1$, the following condition holds:
\[ u^\T \Sigma u \geq  \frac{\phi^2}{\# \cI} \norm{u_{\cI}}_1^2,\]
where (recall that) $\#$ represents the cardinality of a set, and $u_{\cI}$ is a vector with $j$-th elements being $u_j$, i.e., $j$-th element in vector $u$, if $j$ belongs to index set $\cI$ and zero otherwise.
\end{definition}

A standard assumption in high-dimensional Lasso literature is:

\begin{assumption}\label{A2}
The index set $\cI = \operatorname{supp}(\Delta_\beta)$ \eqref{eq:beta_diff} and target domain sample covariance matrix $\Sigma$ \eqref{eq:cov} meet the above compatibility condition with constant $\phi > 0$.
\end{assumption}

This assumption guarantees the identifiability of $\Delta_\beta$, and it holds automatically when the target domain sample covariance is PD. However, when $n < d$, the target domain sample covariance is rank-deficient, and Assumption~\ref{A2} is crucial for the identifiability of $\Delta_\beta$.

\begin{assumption}\label{A3}
    The function $G(\cdot)$ is strongly convex with $\gamma > 0$, i.e., for all $w_1,w_2$ in its domain, the following holds:
    \[G(w_1) - G(w_2) \geq G'(w_2) (w_1 - w_2) + \gamma \frac{(w_1 - w_2)^2}{2}.\]
\end{assumption}

Assumption~\ref{A3} is standard in GLM literature, and it is automatically satisfied when the link function $G'(\cdot) = g(\cdot)$ \eqref{eq:GLM_propensityscore} is linear, i.e., $g(x) = x$ with domain $x \in [0,1]$. Now, we are ready to present the recovery guarantee for the $\ell_1$ regularized TL for the PS model nuisance parameters.

\begin{lemma}[Transferable guarantee for PS model]\label{lma:TL-beta}
Under Assumptions~\ref{A0}, \ref{A1}, \ref{A2} and \ref{A3}, when the PS model \eqref{eq:GLM_propensityscore} is correctly specified and the difference $\Delta_\beta$ \eqref{eq:beta_diff} is $s$-sparse, the following holds for the estimator $\hat \beta_\myt$ \eqref{eq:step2} with regularization strength parameter $\lambda_{\rm PS} > 0$:
\begin{equation}\label{eq:upper_hatbeta}
\begin{split}
    &\PP\left(\left\|\hat{\beta}_\myt -\beta_\myt \right\|_1 \geq \frac{5 \lambda_{\rm PS}}{\gamma}\left(\frac{1}{8 \psi^2}+\frac{1}{\psi}+\frac{s}{\phi^2}\right)\right) \leq \\
    & \quad \quad \quad \quad \quad \quad \quad \quad \quad \quad \quad \quad \quad \quad \quad \quad  2 d \exp \left(-\frac{2\lambda_{\rm PS}^2 n}{125 M_X^2}\right)+2 d \exp \left(-\frac{2\lambda_{\rm PS}^2 n_\mys}{ 5 d^2 M_X^2}\right).
\end{split}
\end{equation}
\end{lemma}

\begin{remark}\label{remark:l1_over_l2}
    As one will see later in the next subsection, the above error bound is invoked when we upper bound the error for the estimated propensity scores, i.e., $|g(\boldsymbol x_i^\T \beta_\myt) - g(\boldsymbol x_i^\T \hat \beta_\myt)|$, which involves applying Hölder's inequality to get 
    \[\boldsymbol x_i^\T (\beta_\myt - \hat \beta_\myt) \leq |\boldsymbol x_i^\T (\beta_\myt - \hat \beta_\myt)| \leq \|\boldsymbol x_i\|_{p_1} \|\beta_\myt - \hat \beta_\myt\|_{p_2},\]
    with $1/p_1+1/p_2 = 1, \ p_1, p_2 \geq 1$. Notice that common choices include $(p_1,p_2) = (2,2)$ and $(\infty,1)$. As mentioned earlier, we can invoke Restricted Eigenvalue Condition \citep{candes2007dantzig,bickel2009simultaneous,meinshausen2009lasso,van2009conditions} to upper bound $\|\beta_\myt - \hat \beta_\myt\|_{2}$, which scales as $\sqrt{s}$ instead of $s$; however $\|\boldsymbol x_i\|_{2}$ will scale as $\sqrt{n}$ under Assumption~\ref{A0}, which typically dominates the sparsity term in our regime \eqref{eq:regime}. Therefore, the overall error upper bound on ACE estimate will deteriorate to $\cO(\sqrt{sn\log d/n})$, compared with $\cO(s \sqrt{\log d/n})$ (to be presented below). This explains why we use the Compatibility Condition to obtain the $\ell_1$ error bound for the estimated nuisance parameters instead of using the Restricted Eigenvalue Condition to get the $\ell_2$ error bound.
\end{remark}

\subsection{Guarantee for plug-in TLIPW estimator}
To bound the absolute estimation error $|\hat \tau_{\rm TLIPW} - \tau|$, we additionally need some (mild) technical assumptions:

\begin{assumption}\label{A4}
    The target domain outcomes are uniformly bounded, i.e., there exists $M_Y > 0$ such that  $|y_i| \leq M_Y,  i = 1,\dots,n$. 
\end{assumption}

This technical assumption helps simplify the analysis; however, our following theoretical analysis will also hold for sub-Gaussian (see Definition~\ref{def:subG}) outcome random variables as shown by the techniques used in the proof of Theorem~\ref{thm:OR}, case (I).

\begin{assumption}\label{A5}
    The propensity scores evaluated on the target domain data are bounded away from zero and one, i.e., there exists $0 < m_g < 1/2$ such that \[m_g \leq e(\boldsymbol x_i) = g(\boldsymbol x_i^\T \beta_\myt) \leq 1 - m_g, \quad i = 1,\dots,n.\]
\end{assumption}

Assumption~\ref{A5} is standard for proving the theoretical guarantee of
IPW estimator, see \citet{wooldridge2002inverse,wooldridge2007inverse} for classic asymptotic analysis for the IPW estimator's $\sqrt{n}$-consistency and asymptotic normality (cf. Theorems 3.1 and 4.1 \citep{wooldridge2002inverse} respectively). Now,
by leveraging Hoeffding's inequality, we can establish the following concentration result:

\begin{lemma}\label{lma:IPW}
Under Assumptions~\ref{A4} and \ref{A5}, for any $t > 0$, we have:
\begin{equation}\label{eq:upper_tau_1}
\begin{split}
    &\PP\left(\left|\frac{1}{n} \sum_{i = 1}^{n} \frac{z_{i} y_{i}}{g(\boldsymbol{x}_{i}^{\T} \beta_\myt )} - \frac{(1-z_{i}) y_{i}}{1-g(\boldsymbol{x}_{i}^{\T} \beta_\myt )} - \tau \right| \geq t\right) \leq 4  \exp \left(-\frac{m_g^2 t^2 n}{8 M_Y^2}\right).
\end{split}
\end{equation}
\end{lemma}

Before presenting the non-asymptotic guarantee for the TLIPW estimator, we additionally impose the following technical assumption for simplicity:
\begin{assumption}\label{Alip}
    The link function $g(\cdot)$ is $L$-Lipschitz with constant $L > 0$, i.e., for $x_1, x_2$ in its domain we have $|g(x_1) - g(x_2)| \leq L|x_1 - x_2|$.
\end{assumption}

Finally, with the help of the above lemmas, we can establish the non-asymptotic upper bound on the absolute estimation error of $\hat \tau_{\rm TLIPW}$ as follows:
\begin{theorem}[Non-asymptotic recovery guarantee for $\hat \tau_{\rm TLIPW}$ \eqref{eq:TLIPW}]\label{thm}
Under Assumptions~\ref{A0}, \ref{A1}, \ref{A2}, \ref{A3}, \ref{A4}, \ref{A5} and \ref{Alip}, for any constant $\delta > 0$, if the PS model \eqref{eq:GLM_propensityscore} is correctly specified and the difference $\Delta_\beta$ \eqref{eq:beta_diff} is $s$-sparse, as $n, n_\mys \rightarrow \infty$, suppose \eqref{eq:regime} holds, i.e.,
\begin{equation*}%\label{eq:regime}
    s \sqrt{\frac{\log d}{n}}= o(1), \quad d \sqrt{\frac{n}{n_\mys}} = \cO(1),
\end{equation*}
we take $\ell_1$ regularization strength parameter to be
\begin{equation}\label{eq:lambda_1}
    \lambda_{\rm PS} = \sqrt{\frac{5 M_X^2 \log(6nd)}{2n}\max\left\{25, \frac{n d^2}{n_\mys}\right\}},
\end{equation}
and we will have
\begin{equation}\label{eq:main}
    \begin{split}
        \PP\Bigg(\left| \hat \tau_{\rm TLIPW} - \tau \right| \leq (1+\delta) \left( C_1 s \sqrt{\frac{\log n + \log d}{n}\max\left\{1, \frac{n d^2}{25 n_\mys}\right\}} + \frac{2 M_Y}{m_g} \sqrt{\frac{\log n}{n}}\right)&\Bigg) \\
        & \geq 1 - \frac{1}{n},
    \end{split}
\end{equation}
where constant $C_1 = C_1(M_X,M_Y,\psi,\phi;\gamma,m_g,L)$ is defined as:
\begin{equation*}
    C_1 = \frac{100\sqrt{5}M_X^2 M_Y L}{\sqrt{2}m_g^2 \gamma}\left(\frac{1}{8\psi^2} + \frac{1}{\psi} + \frac{1}{\phi^2}\right).
\end{equation*}
\end{theorem}

\subsection{Proofs}\label{appendix:proof_TLIPW}

\begin{proof}[Proof outline of Lemma~\ref{lma:TL-beta}]
This proof mostly follows the proof of Theorem 6 in \citet{bastani2021predicting}. The differences in our setting come from: {\it (i)} The Bernoulli r.v.s are sub-Gaussian with variance bounded by $1/4$, which implies
\[\EE[Z - g(\boldsymbol X^\T \beta_\myt)] = 0, \quad \Var(Z - g(\boldsymbol X^\T \beta_\myt)) \leq 1/4 + 1 = 5/4.\]
We need to substitute the variance terms with this upper bound (i.e., $5/4$). 

{\it (ii)} By Assumption~\ref{A0}, we have
\[\sum_{i=1}^n (\boldsymbol x_i)_j^2 \leq n M_X^2,\]
where $(\boldsymbol x_i)_j$ denotes the $j$-th element in the vector $\boldsymbol x_i$.
This implies that $\sum_{i=1}^n (z_i - g(\boldsymbol x_i^\T \beta_\myt))(\boldsymbol x_i)_j$ is ($\sqrt{5n}M_X/2$)-sub-Gaussian (cf. Lemma 16 in \citet{bastani2021predicting}).
Notice that this is different from ``$\sum_{i=1}^n (\boldsymbol x_i)_j^2 = n$'' due to the ``normalized feature assumption'' in the proof of Lemma 4 \citet{bastani2021predicting}. Therefore, in addition to substituting the variance terms as mentioned in (i), we need to include the additional $M_X$ term due to different model assumptions. 
Lastly, we perform the same modification to Lemma 5 and its proof in \citet{bastani2021predicting}, and these lead to \eqref{eq:upper_hatbeta}. For complete details of the proof, we refer readers to Appendix C in \citet{bastani2021predicting}.
\end{proof}

\begin{proof}[Proof of Lemma~\ref{lma:IPW}]
For correctly specified propensity score model \eqref{eq:GLM_propensityscore}, the IPW estimator is unbiased as shown in eq.~\eqref{eq:IPW_unbiased}.
Notice that Assumptions~\ref{A4} and \ref{A5} ensures 
\[\left |\frac{z_{i} y_{i}}{g(\boldsymbol{x}_{i}^{\T} \beta_\myt )}\right| \leq \frac{M_Y}{m_g}.\]
By Hoeffding's inequality, we have
\begin{equation*}
\begin{split}
    &\PP\left(\left|\frac{1}{n} \sum_{i = 1}^{n} \frac{z_{i} y_{i}}{g(\boldsymbol{x}_{i}^{\T} \beta_\myt )}  - \EE[Y_1] \right| \geq t\right) \leq 2  \exp \left(-\frac{m_g^2 t^2 n}{2 M_Y^2}\right).
\end{split}
\end{equation*}

Similarly, we have $
\EE\left[\frac{(1-Z) Y}{1-e(\boldsymbol{X})}\right]=E[Y_0]$, and we can show
\begin{equation*}
\begin{split}
    &\PP\left(\left|\frac{1}{n} \sum_{i = 1}^{n} \frac{(1-z_{i}) y_{i}}{1 - g(\boldsymbol{x}_{i}^{\T} \beta_\myt )}  - \EE[Y_0] \right| \geq t\right) \leq 2  \exp \left(-\frac{m_g^2 t^2 n}{2 M_Y^2}\right).
\end{split}
\end{equation*}

Recall that $\tau = \EE[Y_1] - \EE[Y_0]$, we have 
\begin{align*}
    &\PP\left(\left|\frac{1}{n} \sum_{i = 1}^{n} \frac{z_{i} y_{i}}{g(\boldsymbol{x}_{i}^{\T} \beta_\myt )} - \frac{(1-z_{i}) y_{i}}{1-g(\boldsymbol{x}_{i}^{\T} \beta_\myt )} - \tau \right| \geq t\right) \\
    \leq &\PP\left(\left|\frac{1}{n} \sum_{i = 1}^{n} \frac{z_{i} y_{i}}{g(\boldsymbol{x}_{i}^{\T} \beta_\myt )}  - \EE[Y_1] \right| \geq t/2\right) + \PP\left(\left|\frac{1}{n} \sum_{i = 1}^{n} \frac{(1-z_{i}) y_{i}}{1 - g(\boldsymbol{x}_{i}^{\T} \beta_\myt )}  - \EE[Y_0] \right| \geq t/2\right)\\
 \leq &4  \exp \left(-\frac{m_g^2 t^2 n}{8 M_Y^2}\right).
\end{align*}
We complete the proof.
\end{proof}
\begin{proof}[Proof of Theorem~\ref{thm}]
One one hand, plugging the regularization parameter choice \eqref{eq:lambda_1} into \eqref{eq:upper_hatbeta} yields:
\begin{equation}\label{eq:pf1}
    \PP\left(\left\|\hat{\beta}_\myt -\beta_\myt \right\|_1 \geq \frac{5 \lambda_{\rm PS}}{\gamma}\left(\frac{1}{8 \psi^2}+\frac{1}{\psi}+\frac{s}{\phi^2}\right)\right) \leq \frac{2}{3n}.
\end{equation}
On the other hand, by setting $t = \frac{2 M_Y}{m_g} \sqrt{\frac{\log(12n)}{n}}$ in eq.~\eqref{eq:upper_tau_1} we have
\begin{equation}\label{eq:pf2}
    \PP\left(\left|\frac{1}{n} \sum_{i = 1}^{n} \frac{z_{i} y_{i}}{g(\boldsymbol{x}_{i}^{\T} \beta_\myt )} - \frac{(1-z_{i}) y_{i}}{1-g(\boldsymbol{x}_{i}^{\T} \beta_\myt )} - \tau \right| \geq \frac{2 M_Y}{m_g} \sqrt{\frac{\log(12n)}{n}}\right) \leq \frac{1}{3n}.
\end{equation}

Due to Assumptions~\ref{A4} and \ref{A5}, we have
\begin{equation}
    \begin{split}\label{eq:pf3}
        \left|\frac{1}{n} \sum_{i = 1}^{n} \frac{z_{i} y_{i}}{g(\boldsymbol{x}_{i}^{\T} \beta_\myt )} - \frac{z_{i} y_{i}}{g(\boldsymbol{x}_{i}^{\T} \hat \beta_\myt )} \right| \leq \frac{1}{n} \sum_{i = 1}^{n} \frac{ M_Y |g(\boldsymbol{x}_{i}^{\T} \beta_\myt ) - g(\boldsymbol{x}_{i}^{\T} \hat \beta_\myt )|}{m_g\left(m_g - |g(\boldsymbol{x}_{i}^{\T} \beta_\myt ) - g(\boldsymbol{x}_{i}^{\T} \hat \beta_\myt )|\right)}.
    \end{split}
\end{equation}
Since $g(\cdot)$ is $L$-Lipschitz, we have
\begin{equation*}
    |g(\boldsymbol{x}_{i}^{\T} \beta_\myt ) - g(\boldsymbol{x}_{i}^{\T} \hat \beta_\myt )| \leq L |\boldsymbol{x}_{i}^{\T} (\beta_\myt -  \hat \beta_\myt)| \leq L \norm{\boldsymbol{x}_{i}}_\infty \left\|\hat{\beta}_\myt -\beta_\myt \right\|_1,
\end{equation*}
where the last inequality comes from Hölder's inequality. Due to Assumption~\ref{A4} and the fact that $f(x) = {x}/{(m_g - x)}$ monotonically increase on a domain $0 \leq x < m_g$, we can further bound the right-hand side (RHS) of \eqref{eq:pf3} as follows:
\begin{equation}
    \begin{split}\label{eq:pf4}
        \left|\frac{1}{n} \sum_{i = 1}^{n} \frac{z_{i} y_{i}}{g(\boldsymbol{x}_{i}^{\T} \beta_\myt )} - \frac{z_{i} y_{i}}{g(\boldsymbol{x}_{i}^{\T} \hat \beta_\myt )} \right| & \leq \frac{1}{n} \sum_{i = 1}^{n} \frac{M_X M_Y L\left\|\hat{\beta}_\myt -\beta_\myt \right\|_1}{m_g\left(m_g - M_X L\left\|\hat{\beta}_\myt -\beta_\myt \right\|_1\right)}\\
        & \leq \frac{1}{n} \sum_{i = 1}^{n} \frac{M_X M_Y L\left\|\hat{\beta}_\myt -\beta_\myt \right\|_1}{m_g^2/2}.
    \end{split}
\end{equation}
The above inequality will hold since, for large enough $n, n_\mys$ and in the regime \eqref{eq:regime}, Lemma~\ref{lma:TL-beta} guarantees $M_X L\left\|\hat{\beta}_\myt -\beta_\myt \right\|_1 \rightarrow 0,$ and therefore we will have $M_X L\left\|\hat{\beta}_\myt -\beta_\myt \right\|_1 \leq m_g/2$.
Similarly, we can obtain
\begin{equation}
    \begin{split}\label{eq:pf5}
        \left|\frac{1}{n} \sum_{i = 1}^{n} \frac{(1-z_{i}) y_{i}}{1-g(\boldsymbol{x}_{i}^{\T} \beta_\myt )} - \frac{(1-z_{i}) y_{i}}{1 - g(\boldsymbol{x}_{i}^{\T} \hat \beta_\myt )} \right| \leq \frac{1}{n} \sum_{i = 1}^{n} \frac{M_X M_Y L\left\|\hat{\beta}_\myt -\beta_\myt \right\|_1}{m_g^2/2}.
    \end{split}
\end{equation}

Now, \eqref{eq:pf1} and \eqref{eq:pf2} tell us that, with probability as least $1 - 1/n$,
\begin{equation*}
    \begin{split}
        & \left| \hat \tau_{\rm TLIPW} - \tau \right| \\
         \leq & \left| \hat \tau_{\rm TLIPW} - \frac{1}{n} \sum_{i = 1}^{n} \frac{z_{i} y_{i}}{g(\boldsymbol{x}_{i}^{\T} \beta_\myt )} - \frac{(1-z_{i}) y_{i}}{1-g(\boldsymbol{x}_{i}^{\T} \beta_\myt )} \right| +\left| \frac{1}{n} \sum_{i = 1}^{n} \frac{z_{i} y_{i}}{g(\boldsymbol{x}_{i}^{\T} \beta_\myt )} - \frac{(1-z_{i}) y_{i}}{1-g(\boldsymbol{x}_{i}^{\T} \beta_\myt )} - \tau \right| \\
         <& \frac{20 M_X M_Y L \lambda_{\rm PS}}{m_g^2 \gamma}\left(\frac{1}{8\psi^2} + \frac{1}{\psi} + \frac{s}{\phi^2}\right) + \frac{2 M_Y}{m_g} \sqrt{\frac{\log(12n)}{n}} \\
         \leq & \frac{20 M_X M_Y L \lambda_{\rm PS}}{m_g^2 \gamma}\left(\frac{1}{8\psi^2} + \frac{1}{\psi} + \frac{1}{\phi^2}\right) s + \frac{2 M_Y}{m_g} \sqrt{\frac{\log(12n)}{n}}.
    \end{split}
\end{equation*}
Here, the last inequity above assumes $s \geq 1$; however, this inequality is merely to make the result look more concise, and we can handle the $s=0$ case in a similar manner.
Plugging the $\lambda_{\rm PS}$ choice \eqref{eq:lambda_1} into the above equation, and notice that, for any constant $\delta > 0$, for large enough $n$ the following holds:
\[ \sqrt{\log (12n)} = \sqrt{\log 12 + \log n} \leq (1 + \delta) \sqrt{\log n}, \quad \sqrt{\log(6nd)} \leq (1 + \delta) \sqrt{\log(nd)}.\]
We can obtain the non-asymptotic result in eq.~\eqref{eq:main}.
Now we complete the proof.

\end{proof}

\section{Non-Asymptotic  Recovery Guarantee for TLOR}\label{appendix:TLOR}

\subsection{Guarantee for OR model estimation with knowledge transfer}
Now we prove the non-asymptotic guarantee for our proposed TLOR estimator.

\begin{definition}\label{def:subG}
A random variable $Z \in \RR$ is $\sigma$-sub-Gaussian if $
\mathbb{E}\left[e^{t z}\right] \leq e^{\sigma^2 t^2 / 2}
$ for all $t \in \RR$.
\end{definition}

Many classical distributions are subgaussian; typical examples include any bounded, centered distribution, or the normal distribution.
For $z \in \{0,1\}$, we denote the ``noise terms'' in the OR models \eqref{eq:LM_outcomeregression} as follows:
\begin{equation*}%\label{eq:LM_outcomeregression_noise}
    \begin{split}
        & \epsilon_{z} = Y_z - \EE[Y_z | \boldsymbol{X}] = Y_z - \boldsymbol{X}^\T \alpha_{z,\myt}, \\
        & \epsilon_{z,\mys} = Y_{z,\mys} - \EE[Y_{z,\mys} | \boldsymbol{X}_\mys] = Y_{z,\mys} -\boldsymbol{X}_\mys^\T \alpha_{z,\mys}.
    \end{split}
\end{equation*}

\begin{assumption}\label{AsubG}
     The noise terms are sub-Gaussian, i.e., for $z \in \{0,1\}$, there exist constants $\sigma, \sigma_\mys > 0$ such that $\epsilon_{z}$ is $\sigma$-sub-Gaussian, and $\epsilon_{z,\mys}$ is $\sigma_\mys$-sub-Gaussian. 
\end{assumption}

\begin{assumption}\label{A2-OR}
For $z \in \{0,1\}$, the index set $\cI = \operatorname{supp}(\Delta_{\alpha,z})$ \eqref{eq:alpha_diff} and target domain sample covariance matrix $\Sigma$ \eqref{eq:cov} meet the compatibility condition with $\phi_z > 0$.
\end{assumption}

\begin{lemma}[Transferable guarantee for OR model, cf. Theorem 5 \citep{bastani2021predicting}]\label{lma:TL-alpha}
Under Assumptions~\ref{A0}, \ref{A1}, \eqref{AsubG} and \ref{A2-OR}, assume the sample balanceness condition \eqref{Asampleratio} holds, for $z \in \{0,1\}$, when the OR model \eqref{eq:LM_outcomeregression} is correctly specified and the difference $\Delta_{\alpha, z}$ \eqref{eq:alpha_diff} is $s_z$-sparse, i.e., 
\[\norm{\Delta_{\alpha,0}}_0 \leq s_0, \quad \norm{\Delta_{\alpha,1}}_0 \leq s_1,\]
the following holds for the estimator $\hat{\alpha}_{z,\myt}$ \eqref{eq:step4_DR} with regularization strength parameter $\lambda_{\rm OR} > 0$:
\begin{equation}\label{eq:upper_hatalpha}
\begin{split}
    &\PP\left(\left\|\hat{\alpha}_{z,\myt} -\alpha_{z,\myt} \right\|_1 \geq 5 \lambda_{\rm OR}\left(\frac{1}{4 \psi^2}+\frac{1}{\psi}+\frac{s_z}{2\phi_z^2}\right)\right) \leq \\
    &  \quad \quad \quad \quad \quad \quad \quad \quad \quad \quad \quad \quad \quad \quad  \quad \quad 2 d \exp \left(-\frac{r n \lambda_{\rm OR}^2}{200 \sigma^2 M_X^2}\right)+2 d \exp \left(-\frac{r n_\mys \lambda_{\rm OR}^2 }{2 d^2 \sigma_\mys^2 M_X^2}\right).
\end{split}
\end{equation}
\end{lemma}

\subsection{Guarantee for plug-in TLOR estimator}
Here, we impose an additional technical assumption that the covariates in the target domain are sub-Gaussian such that there exists constant $\sigma_{X} > 0$:
\begin{equation}\label{eq:subG_X}
    \PP \left( \left|\frac{1}{n} \sum_{i = 1}^{n} \boldsymbol{x}_{i}^{\T} \alpha_{z,\myt} -  \EE[Y_z]\right|>t \right) \leq 2 \exp \left(-\frac{nt^2}{2\sigma_X^2}\right), \quad z \in \{0,1\}.
\end{equation}

\begin{theorem}[Non-asymptotic recovery guarantee for $\hat \tau_{\rm TLDR}$ \eqref{eq:TLDR}]\label{thm:OR_0}
Under Assumptions~\ref{A0}, \ref{A1}, \ref{AsubG} and \ref{A2-OR}, suppose the sample balances condition \eqref{Asampleratio} holds and the covariates in target domain follow a sub-Gaussian distribution \eqref{eq:subG_X},
as $n, n_\mys \rightarrow \infty$, suppose \eqref{eq:regime} holds, for any constant $\delta > 0$, when the OR model \eqref{eq:LM_outcomeregression} is correctly specified with $s_z$-sparse $\Delta_{z,\alpha}$ \eqref{eq:alpha_diff} (for $z \in \{0,1\}$), i.e., 
\[\norm{\Delta_{\alpha,0}}_0 \leq s_0, \quad \norm{\Delta_{\alpha,1}}_0 \leq s_1,\] 
then by taking
\begin{equation}\label{eq:lambda_OR_0}
    \lambda_{\rm OR} = \sqrt{2 M_X^2 \log (12nd) \max\left\{\frac{100 \sigma^2}{rn},\frac{d^2 \sigma_\mys^2}{r n_\mys}\right\}},
\end{equation}
we will have
\begin{equation}\label{eq:main_OR}
    \begin{split}
        &\PP\Bigg(\left| \hat \tau_{\rm TLOR} - \tau \right| \leq (1 + \delta) \Bigg( C_2 \sigma (s_0+s_1) \sqrt{ \frac{\log n + \log d}{rn} \max\left\{100 ,\frac{n  \sigma_\mys^2 d^2}{ n_\mys \sigma^2}\right\}}  \\
        & \quad \quad \quad \quad \quad \quad \quad \quad \quad \quad \quad \quad \quad \quad \quad \quad \quad \quad \quad \quad \quad \quad \quad \quad \quad \ + 2 \sigma_X \sqrt{\frac{2\log n}{rn}} \Bigg) \Bigg) \geq 1 - \frac{1}{n}.
    \end{split}
\end{equation}
where constant $C_2 = C_2(M_X,\psi,\phi_0,\phi_1;m_g)$ is defined as follows:
\begin{align}\label{eq:C2}
    C_2  = 5 \sqrt{2} M_X^2  \left( \frac{1}{m_g} - 1 \right) \left(\frac{1}{2 \psi^2}+\frac{2}{\psi}+\frac{1}{2\phi_0^2} + \frac{ 1}{2\phi_1^2}\right). 
\end{align}
\end{theorem}

\subsection{Proof}

The proof of Lemma~\ref{lma:TL-alpha} mostly follows that of Theorem 5 in \citet{bastani2021predicting} and we omit it here. We only give detailed proof of the main theorem.

\begin{proof}[Proof of Theorem~\ref{thm:OR_0}]
The absolute error of the TLOR estimator \eqref{eq:TLOR} can be decomposed as follows:
\begin{equation*}
    \begin{split}
         \left| \hat \tau_{\rm TLOR} - \tau \right| \leq & \left|\frac{1}{n_1} \sum_{z_i = 1} \boldsymbol{x}_{i}^{\T} (\hat \alpha_{1,\myt} - \alpha_{1,\myt}) \right| + \left|\frac{1}{n_0} \sum_{z_i = 0} \boldsymbol{x}_{i}^{\T} (\hat \alpha_{0,\myt}  - \alpha_{0,\myt}) \right|\\
         & + \left|\frac{1}{n_1} \sum_{z_i = 1} \boldsymbol{x}_{i}^{\T} \alpha_{1,\myt} - \frac{1}{n_0} \sum_{z_i = 0} \boldsymbol{x}_{i}^{\T} \alpha_{0,\myt}  - \tau \right|
         \\
         \leq & \ M_X \big(\norm{\hat \alpha_{1,\myt} - \alpha_{1,\myt}}_1 + \norm{\hat \alpha_{0,\myt} - \alpha_{0,\myt}}_1 \big) \\
         & +  \left|\frac{1}{n_1} \sum_{z_i = 1} \boldsymbol{x}_{i}^{\T} \alpha_{1,\myt} - \frac{1}{n_0} \sum_{z_i = 0} \boldsymbol{x}_{i}^{\T} \alpha_{0,\myt}  - \tau \right|.
    \end{split}
\end{equation*}

On one hand, by setting the RHS of \eqref{eq:subG_X} as $1/(6n)$, we have that, with probability at least $1 - 1/(3n)$, the following holds:
\[\left|\frac{1}{n} \sum_{i = 1}^{n} \boldsymbol{x}_{i}^{\T} \alpha_{1,\myt} - \boldsymbol{x}_{i}^{\T} \alpha_{0,\myt}  - \tau \right| \leq 2 \sqrt{\frac{2\sigma_X^2 \log(12n) }{n}}.\]

One the other hand, following Lemma~\ref{lma:TL-alpha} and taking $\lambda_{\rm OR}$ as in eq.~\eqref{eq:lambda_OR_0}, we will have that, with probability at least $1 - 2/(3n)$, 
\[\norm{\hat \alpha_{1,\myt} - \alpha_{1,\myt}}_1 + \norm{\hat \alpha_{0,\myt} - \alpha_{0,\myt}}_1 \leq 5 \lambda_{\rm OR}\left(\frac{1}{2 \psi^2}+\frac{2}{\psi}+\frac{s_0}{2\phi_0^2}+ \frac{ s_1}{2\phi_1^2}\right).\]

Now, combing the above inequalities, we will have that, with probability at least $1-1/n$, the following holds:
\begin{equation*}
    \begin{split}
           \left| \hat \tau_{\rm TLOR} - \tau \right|\leq  & \ 2 \sqrt{\frac{2\sigma_X^2 \log(12n) }{n}} + 5 M_X \left( \frac{1}{m_g} - 1\right) \lambda_{\rm OR}\left(\frac{1}{2 \psi^2}+\frac{2}{\psi}+\frac{s_0}{2\phi_0^2}+ \frac{ s_1}{2\phi_1^2}\right)  \\
         = & \ \cO\left( (s_0 + s_1) \sqrt{\log d} \left(\frac{1}{\sqrt{rn}} + \frac{d}{\sqrt{rn_\mys}}\right)\right).
    \end{split}
\end{equation*}
To get \eqref{eq:main_OR}, we only need to notice that, for any constant $\delta > 0$, for large enough $n$ the following holds:
\[ \sqrt{\log (12n)} = \sqrt{\log 12 + \log n} \leq (1 + \delta) \sqrt{\log n}.\]
We can handle the $\log(12nd)$ term in $\lambda_{\rm OR}$ \eqref{eq:lambda_OR_0} in a similar manner and obtain $\sqrt{\log(12nd)} \leq (1 + \delta) \sqrt{\log(nd)}$. Now, we obtain the non-asymptotic result in eq.~\eqref{eq:main_OR} and complete the proof.
\end{proof}

\section{Non-Asymptotic  Recovery Guarantee for TLDR}\label{appendix:TLDR}

\subsection{Guarantee for plug-in TLDR estimator}
We impose another technical assumption as \citet{bastani2021predicting} did (see Assumption 1 therein):
\begin{assumption}\label{AMalpha}
    The ground truth target domain OR model parameters are bounded, i.e., for $z \in \{0,1\}$, there exists $M_\alpha > 0$ such that $\|\alpha_{z,\myt}\|_1 < M_\alpha$.
\end{assumption}

Similarly, due to the doubly robustness of the DR estimator, we have the following non-asymptotic recovery guarantee:
\begin{theorem}[Non-asymptotic recovery guarantee for $\hat \tau_{\rm TLDR}$ \eqref{eq:TLDR}]\label{thm:OR}
Under Assumptions~\ref{A0} and \ref{A1}, as $n, n_\mys \rightarrow \infty$, suppose \eqref{eq:regime} holds. For any constant $\delta > 0$:

\noindent
(I) When the PS model \eqref{eq:GLM_propensityscore} is correctly specified and $\Delta_\beta$ \eqref{eq:beta_diff} is $s$-sparse, if we additionally assume Assumptions \ref{A2}, \ref{A3}, \ref{A4}, \ref{A5}, \ref{Alip} and \ref{AMalpha} hold, then by taking
\begin{equation}\label{eq:lambda_PS}
    \lambda_{\rm PS} = \sqrt{\frac{5 M_X^2 \log(10nd)}{2n}\max\left\{25, \frac{n d^2}{n_\mys}\right\}}, 
\end{equation}
we will have
\begin{equation}\label{eq:main_DR_PS}
    \begin{split}
        &\PP\Bigg(\left| \hat \tau_{\rm TLDR} - \tau \right| \leq (1 + \delta) \left( C_4 s \sqrt{\frac{ \log n + \log d}{n}\max\left\{1, \frac{25 n d^2}{n_\mys}\right\}} + C_5\sqrt{\frac{\log n}{n}} \right)\Bigg) \geq 1 - \frac{1}{n}.
    \end{split}
\end{equation}
where constants $C_4 = C_4(M_X,M_Y,\psi,\phi;\gamma,m_g,L;M_\alpha)$ and $C_5 = C_5(M_X,M_Y;m_g;M_\alpha)$ are defined as follows:
\begin{align*}
    C_4 = \frac{100\sqrt{5}  M_X^2 \left(  M_X M_\alpha + {M_Y}/{m_g} \right)  L }{\sqrt{2} m_g\gamma}  \left(\frac{1}{8\psi^2} + \frac{1}{\psi} + \frac{1}{\phi^2}\right), \quad C_5 = 2\frac{ M_Y  +  \sqrt{10}M_X M_\alpha }{m_g }. 
\end{align*}

\noindent
(II) When the OR model \eqref{eq:LM_outcomeregression} is correctly specified and $\Delta_{z,\alpha}$ \eqref{eq:alpha_diff} is $s_z$-sparse (for $z \in \{0,1\}$), i.e., 
\[\norm{\Delta_{\alpha,0}}_0 \leq s_0, \quad \norm{\Delta_{\alpha,1}}_0 \leq s_1,\]
under Assumptions \ref{AsubG}, \ref{A2-OR}, and we further assume the sample balances condition \eqref{Asampleratio} holds and the covariates in the target domain are sub-Gaussian \eqref{eq:subG_X}, we strengthen Assumption~\ref{A5} by assuming the link function $g(\cdot)$ \eqref{eq:GLM_propensityscore} takes value on $[m_g,1-m_g]$, then by taking
\begin{equation}\label{eq:lambda_OR}
    \lambda_{\rm OR} = \sqrt{2 M_X^2 \log (16nd) \max\left\{\frac{100 \sigma^2}{rn},\frac{d^2 \sigma_\mys^2}{r n_\mys}\right\}},
\end{equation}
we will have
\begin{equation}\label{eq:main_DR_OR}
    \begin{split}
        &\PP\Bigg(\left| \hat \tau_{\rm TLDR} - \tau \right| \leq (1 + \delta) \Bigg( C_2 \sigma (s_0+s_1) \sqrt{ \frac{\log n + \log d}{rn} \max\left\{100 ,\frac{n  \sigma_\mys^2 d^2}{ n_\mys \sigma^2}\right\}} \\
        & \quad \quad \quad \quad \quad \quad \quad \quad \quad \quad \quad \quad \quad \quad \quad \quad \quad  \quad \quad  \   +  C_3 \sigma \sqrt{\frac{\log n}{n}} + 2 \sigma_X \sqrt{\frac{2\log n}{rn}} \Bigg) \Bigg) \geq 1 - \frac{1}{n}.
    \end{split}
\end{equation}
where constant $C_2 = C_2(M_X,\psi,\phi_0,\phi_1;m_g)$ is defined in eq.~\eqref{eq:C2},
and constant $C_3 = C_3(m_g) = 2\sqrt{2/m_g}$.
\end{theorem}

\subsection{Proof}

\begin{proof}[Proof of Theorem~\ref{thm:OR}]
    We consider two cases: (i) the PS model is correctly specified and (ii) the OR model is correctly specified.

\paragraph{Case (I): the PS model is correctly specified.} This proof closely resembles that of Theorem~\ref{thm}. We only need to show the augmented terms in the DR estimator, i.e., 
\begin{equation*}
    \frac{1}{n} \sum_{i = 1}^{n}  \frac{ \boldsymbol{x}_{i}^{\T}\hat \alpha_{1,\myt}(z_i - g(\boldsymbol{x}_{i}^{\T}\hat \beta_\myt ))}{g(\boldsymbol{x}_{i}^{\T}\hat \beta_\myt )}, \quad \frac{1}{n} \sum_{i = 1}^{n} \frac{\boldsymbol{x}_{i}^{\T}\hat \alpha_{0,\myt}(z_i - g(\boldsymbol{x}_{i}^{\T}\hat \beta_\myt ))}{1-g(\boldsymbol{x}_{i}^{\T}\hat \beta_\myt )},
\end{equation*}
are very close to zero with high probability, which is straightforward since they both have zero means under the correct PS model specification assumption. 

To simplify our proof below, we impose Assumption~\ref{AMalpha},
which implies that, when $n, n_\mys \rightarrow \infty$, in regime \eqref{eq:regime}, we will have 
$\|\hat \alpha_{z,\myt}\|_1 \leq M_\alpha$ due to Lemma~\ref{lma:TL-alpha}. Similarly, Assumption~\ref{A5} ensures that $m_g/2 \leq g(\boldsymbol{x}_{i}^{\T}\hat \beta_\myt ) \leq 1 - m_g/2$ for sufficiently large $n, n_\mys$. Now, we can show that 
\[\sum_{i = 1}^{n} \frac{ \boldsymbol{x}_{i}^{\T}\hat \alpha_{1,\myt}(z_i - g(\boldsymbol{x}_{i}^{\T} \beta_\myt ))}{g(\boldsymbol{x}_{i}^{\T}\hat \beta_\myt )}\]
is $(\sqrt{5n} M_X M_\alpha / m_g)$-sub-Gaussian. Notice that Bernoulli r.v. has variance bounded by $1/4$ and
\[\left|\frac{ \boldsymbol{x}_{i}^{\T}\hat \alpha_{1,\myt}}{g(\boldsymbol{x}_{i}^{\T}\hat \beta_\myt )}\right| \leq \frac{2M_X M_\alpha}{m_g}.\]
Thus, we have
\begin{equation}\label{eq1:pfthm2}
    \PP \left(\left|\frac{1}{n} \sum_{i = 1}^{n} \frac{ \boldsymbol{x}_{i}^{\T}\hat \alpha_{1,\myt}(z_i - g(\boldsymbol{x}_{i}^{\T} \beta_\myt ))}{g(\boldsymbol{x}_{i}^{\T}\hat \beta_\myt )}\right| > t \right) \leq 2 \exp \left(-\frac{ n m_g^2 t^2}{10 M_X^2 M_\alpha^2}\right).
\end{equation}
Similarly, we can show that
\begin{equation}\label{eq2:pfthm2}
    \PP \left(\left|\frac{1}{n} \sum_{i = 1}^{n} \frac{\boldsymbol{x}_{i}^{\T}\hat \alpha_{0,\myt}(z_i - g(\boldsymbol{x}_{i}^{\T} \beta_\myt ))}{1-g(\boldsymbol{x}_{i}^{\T}\hat \beta_\myt )}\right| > t \right) \leq 2 \exp \left(-\frac{ n m_g^2 t^2}{10 M_X^2 M_\alpha^2}\right).
\end{equation}
We take RHS of Equations~\ref{eq1:pfthm2} and \ref{eq2:pfthm2} to be $1/(5n)$, and therefore with probability at least $1 - 2/(5n)$ we have:
\begin{equation}
    \begin{split}
            & \left|- \frac{1}{n} \sum_{i = 1}^{n}  \frac{ \boldsymbol{x}_{i}^{\T}\hat \alpha_{1,\myt}(z_i - g(\boldsymbol{x}_{i}^{\T}\hat \beta_\myt ))}{g(\boldsymbol{x}_{i}^{\T}\hat \beta_\myt )} + \frac{1}{n} \sum_{i = 1}^{n} \frac{\boldsymbol{x}_{i}^{\T}\hat \alpha_{0,\myt}(z_i - g(\boldsymbol{x}_{i}^{\T}\hat \beta_\myt ))}{1-g(\boldsymbol{x}_{i}^{\T}\hat \beta_\myt )}\right| \\
            \leq & \frac{2 M_X M_\alpha \sqrt{10 \log (10 n)}}{m_g \sqrt{n}} + \frac{4 L M_X^2 M_\alpha \norm{\beta_\myt - \hat \beta_\myt}_1}{m_g}.
    \end{split}
\end{equation}

By taking $\lambda_{\rm PS}$ as in eq.~\eqref{eq:lambda_PS} and following the proof of Theorem~\ref{thm}, we have that with probability at least $1 - 1/n$:
\begin{equation*}
    \begin{split}
        & \left| \hat \tau_{\rm TLDR} - \tau \right| \\
         \leq & \left| \hat \tau_{\rm TLIPW} - \frac{1}{n} \sum_{i = 1}^{n} \frac{z_{i} y_{i}}{g(\boldsymbol{x}_{i}^{\T} \beta_\myt )} - \frac{(1-z_{i}) y_{i}}{1-g(\boldsymbol{x}_{i}^{\T} \beta_\myt )} \right| 
         + \left| \frac{1}{n} \sum_{i = 1}^{n} \frac{z_{i} y_{i}}{g(\boldsymbol{x}_{i}^{\T} \beta_\myt )} - \frac{(1-z_{i}) y_{i}}{1-g(\boldsymbol{x}_{i}^{\T} \beta_\myt )} - \tau \right| \\
         & +  \left|- \frac{1}{n} \sum_{i = 1}^{n}  \frac{ \boldsymbol{x}_{i}^{\T}\hat \alpha_{1,\myt}(z_i - g(\boldsymbol{x}_{i}^{\T}\hat \beta_\myt ))}{g(\boldsymbol{x}_{i}^{\T}\hat \beta_\myt )} + \frac{1}{n} \sum_{i = 1}^{n} \frac{\boldsymbol{x}_{i}^{\T}\hat \alpha_{0,\myt}(z_i - g(\boldsymbol{x}_{i}^{\T}\hat \beta_\myt ))}{1-g(\boldsymbol{x}_{i}^{\T}\hat \beta_\myt )}\right| \\
         \leq & \frac{20  M_X \left( M_\alpha M_X + \frac{M_Y}{m_g} \right)  L \lambda_{\rm PS}}{m_g\gamma}  \left(\frac{1}{8\psi^2} + \frac{1}{\psi} + \frac{s}{\phi^2}\right)+  \frac{2 M_Y\sqrt{{\log(20 n)}}  + 2 M_X M_\alpha \sqrt{10 \log (10 n)}}{m_g \sqrt{n}}\\
         = & \cO\left(s \sqrt{\log d} \left(\frac{1}{\sqrt{n}} + \frac{d}{\sqrt{n_\mys}}\right)\right).
    \end{split}
\end{equation*}
To get \eqref{eq:main_DR_PS}, we only need to notice that, for any constant $\delta > 0$, for large enough $n$ the following holds:
\[ \sqrt{\log (10n)} < \sqrt{\log (20n)} = \sqrt{\log 20 + \log n} \leq (1 + \delta) \sqrt{\log n}.\]
We can handle the $\log(10nd)$ term in $\lambda_{\rm PS}$ \eqref{eq:lambda_PS} in a similar manner and obtain $\sqrt{\log(10nd)} \leq (1 + \delta) \sqrt{\log(nd)}$. Now, we obtain the non-asymptotic result in eq.~\eqref{eq:main_DR_PS} and complete the proof.

\paragraph{Case (II): the OR model is correctly specified.} We rewrite our TLDR estimator \eqref{eq:TLDR} as follows:
\begin{equation*}
\begin{split}
    \hat \tau_{\rm TLDR} = \frac{1}{n} \sum_{i = 1}^{n} & \boldsymbol{x}_{i}^{\T}\hat \alpha_{1,\myt} + \frac{z_{i} (y_{i} - \boldsymbol{x}_{i}^{\T}\hat \alpha_{1,\myt})}{g(\boldsymbol{x}_{i}^{\T}\hat \beta_\myt )}  - \boldsymbol{x}_{i}^{\T}\hat \alpha_{0,\myt} - \frac{(1-z_{i}) (y_{i} - \boldsymbol{x}_{i}^{\T}\hat \alpha_{0,\myt})}{1-g(\boldsymbol{x}_{i}^{\T}\hat \beta_\myt )}.
\end{split}
\end{equation*}
We decompose the estimator error as follows:
\begin{equation*}
    \begin{split}
         \left| \hat \tau_{\rm TLDR} - \tau \right| \leq & \left|\frac{1}{n} \sum_{i = 1}^{n} \boldsymbol{x}_{i}^{\T} \alpha_{1,\myt} - \boldsymbol{x}_{i}^{\T} \alpha_{0,\myt}  - \tau \right| +    \Bigg| \frac{1}{n} \sum_{i = 1}^{n} \left(1 - \frac{z_i}{g(\boldsymbol{x}_{i}^{\T}\hat \beta_\myt )}\right)  \boldsymbol{x}_{i}^{\T}(\hat \alpha_{1,\myt} - \alpha_{1,\myt})\Bigg|  \\
         +& \Bigg| \frac{1}{n} \sum_{i = 1}^{n} \left(1 - \frac{1-z_i}{1-g(\boldsymbol{x}_{i}^{\T}\hat \beta_\myt )}\right) \boldsymbol{x}_{i}^{\T}(\hat \alpha_{0,\myt} - \alpha_{0,\myt}) \Bigg| + \left| \frac{1}{n} \sum_{i = 1}^{n} \frac{z_{i} (y_{i} - \boldsymbol{x}_{i}^{\T} \alpha_{1,\myt})}{g(\boldsymbol{x}_{i}^{\T}\hat \beta_\myt )} \right|  \\
        + & \left| \frac{1}{n} \sum_{i = 1}^{n}\frac{(1-z_{i}) (y_{i} - \boldsymbol{x}_{i}^{\T} \alpha_{0,\myt})}{1-g(\boldsymbol{x}_{i}^{\T}\hat \beta_\myt )} \right|.
    \end{split}
\end{equation*}
Notice that we strengthen the Assumption~\ref{A5} that link function $g(\cdot)$ only takes value on $[m_g,1-m_g]$, which gives us
\begin{equation*}
    \begin{split}
        &\Bigg| \frac{1}{n} \sum_{i = 1}^{n} \left(1 - \frac{z_i}{g(\boldsymbol{x}_{i}^{\T}\hat \beta_\myt )}\right)  \boldsymbol{x}_{i}^{\T}(\hat \alpha_{1,\myt} - \alpha_{1,\myt})\Bigg| \leq \left( \frac{1}{m_g} - 1\right) \frac{1}{n} \sum_{i = 1}^{n} \left |\boldsymbol{x}_{i}^{\T} (\hat \alpha_{1,\myt} - \alpha_{1,\myt}) \right|,\\
        &\Bigg| \frac{1}{n} \sum_{i = 1}^{n} \left(1 - \frac{1-z_i}{1-g(\boldsymbol{x}_{i}^{\T}\hat \beta_\myt )}\right) \boldsymbol{x}_{i}^{\T}(\hat \alpha_{0,\myt} - \alpha_{0,\myt}) \Bigg| \leq \left( \frac{1}{m_g} - 1\right) \frac{1}{n} \sum_{i = 1}^{n} \left |\boldsymbol{x}_{i}^{\T}(\hat \alpha_{0,\myt} - \alpha_{0,\myt}) \right|.
    \end{split}
\end{equation*}
Combing the above derivations with Assumption~\ref{A0}, we have
\begin{equation*}
    \begin{split}
         \left| \hat \tau_{\rm TLDR} - \tau \right| \leq & \left|\frac{1}{n} \sum_{i = 1}^{n} \boldsymbol{x}_{i}^{\T} \alpha_{1,\myt} - \boldsymbol{x}_{i}^{\T} \alpha_{0,\myt}  - \tau \right| + M_X \left( \frac{1}{m_g} - 1\right) \big(\norm{\hat \alpha_{1,\myt} - \alpha_{1,\myt}}_1 + \norm{\hat \alpha_{0,\myt} - \alpha_{0,\myt}}_1 \big) \\
         + & \left| \frac{1}{n} \sum_{i = 1}^{n} \frac{z_{i} (y_{i} - \boldsymbol{x}_{i}^{\T} \alpha_{1,\myt})}{g(\boldsymbol{x}_{i}^{\T}\hat \beta_\myt )} \right|  + \left| \frac{1}{n} \sum_{i = 1}^{n}\frac{(1-z_{i}) (y_{i} - \boldsymbol{x}_{i}^{\T} \alpha_{0,\myt})}{1-g(\boldsymbol{x}_{i}^{\T}\hat \beta_\myt )} \right|.
    \end{split}
\end{equation*}

{\it (i)} Since the OR model is assumed to be correct, we have $\EE[Z(Y - \boldsymbol X^\T \alpha_{Z,\myt})] = 0$. Under Assumption~\ref{A5}, we can show:
\begin{equation}\label{eq3:pfthm2}
    \begin{split}
        & \PP\left(\left|\frac{1}{n} \sum_{i = 1}^{n} \frac{z_{i} (y_{i} - \boldsymbol{x}_{i}^{\T} \alpha_{1,\myt})}{g(\boldsymbol{x}_{i}^{\T}\hat \beta_\myt )}\right| > t\right) \\
        = & \PP\left(\left|\frac{1}{n} \sum_{z_i = 1} \frac{y_{i} - \boldsymbol{x}_{i}^{\T} \alpha_{1,\myt}}{g(\boldsymbol{x}_{i}^{\T}\hat \beta_\myt )}\right| > t\right) \leq 2 \exp\left(-\frac{r n m_g^2 t^2}{2 \sigma^2}\right).
    \end{split}
\end{equation}
Similarly, we will get:
\begin{equation}\label{eq4:pfthm2}
    \begin{split}
        & \PP\left(\left|\frac{1}{n} \sum_{i = 1}^{n} \frac{(1-z_{i}) (y_{i} - \boldsymbol{x}_{i}^{\T} \alpha_{0,\myt})}{1-g(\boldsymbol{x}_{i}^{\T}\hat \beta_\myt )}\right| > t\right) \\
        = & \PP\left(\left|\frac{1}{n} \sum_{z_i = 0} \frac{y_{i} - \boldsymbol{x}_{i}^{\T} \alpha_{0,\myt}}{1-g(\boldsymbol{x}_{i}^{\T}\hat \beta_\myt )}\right| > t\right) \leq 2 \exp\left(-\frac{r n m_g^2 t^2}{2 \sigma^2}\right).
    \end{split}
\end{equation}
By setting the RHS of the above two inequalities as $1/(8n)$, we have that, with probability at least $1 - 1/(4n)$, the following holds:
\begin{equation*}
    \begin{split}
         & \left| \frac{1}{n} \sum_{i = 1}^{n} \frac{z_{i} (y_{i} - \boldsymbol{x}_{i}^{\T} \alpha_{1,\myt})}{g(\boldsymbol{x}_{i}^{\T}\hat \beta_\myt )} \right|  + \left| \frac{1}{n} \sum_{i = 1}^{n}\frac{(1-z_{i}) (y_{i} - \boldsymbol{x}_{i}^{\T} \alpha_{0,\myt})}{1-g(\boldsymbol{x}_{i}^{\T}\hat \beta_\myt )} \right|  \leq  2 \sqrt{\frac{2\sigma^2 \log (16n)}{r n m_g^2}}.
    \end{split}
\end{equation*}

{\it (ii)} 
Similar to the proof of Theorem~\ref{thm:OR_0}, by setting the RHS of \eqref{eq:subG_X} as $1/(8n)$, we have that, with probability at least $1 - 1/(4n)$, the following holds:
\[\left|\frac{1}{n} \sum_{i = 1}^{n} \boldsymbol{x}_{i}^{\T} \alpha_{1,\myt} - \boldsymbol{x}_{i}^{\T} \alpha_{0,\myt}  - \tau \right| \leq 2 \sqrt{\frac{2\sigma_X^2 \log(16n) }{n}}.\]

{\it (iii)} Finally, by taking $\lambda_{\rm OR}$ as in eq.~\eqref{eq:lambda_OR} as well as following Lemma~\ref{lma:TL-alpha}, we will have that, with probability at least $1 - 1/(2n)$, 
\[\norm{\hat \alpha_{1,\myt} - \alpha_{1,\myt}}_1 + \norm{\hat \alpha_{0,\myt} - \alpha_{0,\myt}}_1 \leq 5 \lambda_{\rm OR}\left(\frac{1}{2 \psi^2}+\frac{2}{\psi}+\frac{s_0}{2\phi_0^2}+ \frac{ s_1}{2\phi_1^2}\right).\]

Now, plugging {\it (i), (iii), (iii)} back to the decomposition we will have that, with probability at least $1-1/n$, the error $\left| \hat \tau_{\rm TLDR} - \tau \right|$ can be (upper) bounded by:
\begin{equation*}
    \begin{split}
            & 2 \sqrt{\frac{2\sigma_X^2 \log(16n) }{n}} + 5 M_X \left( \frac{1}{m_g} - 1\right) \lambda_{\rm OR}\left(\frac{1}{2 \psi^2}+\frac{2}{\psi}+\frac{s_0}{2\phi_0^2}+ \frac{ s_1}{2\phi_1^2}\right) + 2 \sqrt{\frac{2\sigma^2 \log (16n)}{r n m_g^2}} \\
         = & \ \cO\left( (s_0 + s_1) \sqrt{\log d} \left(\frac{1}{\sqrt{rn}} + \frac{d}{\sqrt{rn_\mys}}\right)\right).
    \end{split}
\end{equation*}

To get \eqref{eq:main_DR_OR}, we only need to notice that, for any constant $\delta > 0$, for large enough $n$ the following holds:
\[ \sqrt{\log (16n)} = \sqrt{\log 16 + \log n} \leq (1 + \delta) \sqrt{\log n}.\]
We can handle the $\log(16nd)$ term in $\lambda_{\rm OR}$ \eqref{eq:lambda_OR} in a similar manner and obtain $\sqrt{\log(16nd)} \leq (1 + \delta) \sqrt{\log(nd)}$ when $n$ is large enough. Now, we obtain the non-asymptotic result in eq.~\eqref{eq:main_DR_OR} and complete the proof.

\end{proof}

\section{Additional Details of the GLM-Based Experiments}\label{appendix:syn_exp}

Here, for the GLM-based parametric approach, we consider the IPW estimator for the demonstration purpose since the focus of our numerical simulation is on NN-based non-parametric approaches (see Section~\ref{sec:nn_exp} and Appendix~\ref{appendix:add_ihdp_result}).

\subsection{Motivating toy example}
In the toy example presented in Appendix~\ref{sec:setup}, 
the r.v.s $X_1 \sim N(\mu_1,1), X_2 \sim N(\mu_2,1), \epsilon \sim N(0,1/4)$. We choose $\mu_1 = 0$, $\beta_1 = 0.1$ for both domains; in target domain, we take $\mu_{2} = 2$, $\beta_{2} = -0.1$ and the causal effects are chosen as $\tau = -2/30 \approx -0.067$ and $\alpha = 0.1$; in source domain, we take $\mu_{2, \mys} = 1$, $\beta_{2, \mys} = -0.2$. 

We randomly generate $2000$ samples from the target domain, and the IPW estimate is $-0.0531$, which is pretty close to the ground truth ACE and validates the effectiveness of the IPW estimator. However, in our TCL setup, we consider limited target domain samples in that we can only observe the first $100$ target domain samples, which yields a very biased IPW estimate: $\hat \tau_{\texttt{TO-CL}} = 0.0002$. Additionally, we observe $1000$ randomly generated samples from the source domain, but we ``do not know'' whether or not the ACEs and the treatment assignment mechanisms are the same across both domains; apparently, in our toy example, treatment assignment mechanisms are different. 

In this work, we aim to leverage the abundant source domain data to improve propensity score estimation via TL techniques. Since we do not assume the same ACEs in both domains, we evaluate the IPW estimator only using the target domain data. 
However, fitting the PS model using the naively merged datasets (i.e., without the TL techniques) would fail since the treatment assignment mechanisms across different domains are different: in our numerical example, this naive approach yields an IPW estimate $\hat \tau_{\texttt{Merge-CL}} = 0.0441$, which is even more biased than only using target domain data.
Our proposed \texttt{$\ell_1$-TCL} does help yield a more accurate estimated ACE: $\hat \tau_{\rm TLIPW} = -0.0013$.  
This toy example tells us that additionally incorporating the source domain data ``in a smart way'' by using the TL technique does improve the IPW estimator's accuracy --- we can now at least infer that there is an inhibiting causal effect from treatment $Z$ to outcome $Y$.

\subsection{Synthetic-data experiments}

Next, we consider experiments with the GLM parametric approach under more settings. The goal is to simply demonstrate the effectiveness of the proposed \texttt{$\ell_1$-TCL} framework compared with two baselines: solely using target domain data for causal learning, i.e., \texttt{TO-CL}, and naively merging both domains' datasets for causal learning, i.e., \texttt{Merge-CL}.

\subsubsection{Experimental configurations and training details} We generate synthetic data where the treatment assignment follows the GLM-based PS model \eqref{eq:GLM_propensityscore} with randomly generated $d$-dimensional target domain nuisance parameters as well as $s$-sparse difference (from Gaussian distribution). The settings we consider include: $d \in \{10, 20, 50, 75, 100\}$, $s \in \{1, 3, 5, 7, 10\}$, source domain sample size $n_\mys \in \{2000, 3000, 5000\}$ and target domain sample size $n \in \{100, 200, 500\}$.
For demonstration purposes, the link function is chosen to be a sigmoid function and considered as a prior. This reduces the PS model fitting in the rough estimation step to naive logistic regression; in the $\ell_1$ regularized bias correction step, we use gradient descent to optimize the objective function \eqref{eq:re-formulation} with respect to the (sparse) difference, which has in total $8000$ iterations and initial learning rate $0.02$ (which decays by half every $2000$ iterations). Hyperparameter $\lambda_{\rm PS}$ is chosen to maximize the treatment prediction area under the ROC curve (AUC) on a validation (target domain) dataset with size $50$.

\subsubsection{Results} 
\begin{figure}[!htp]
%%\vspace{-0.1in}
\centerline{
\includegraphics[width = .85\textwidth]{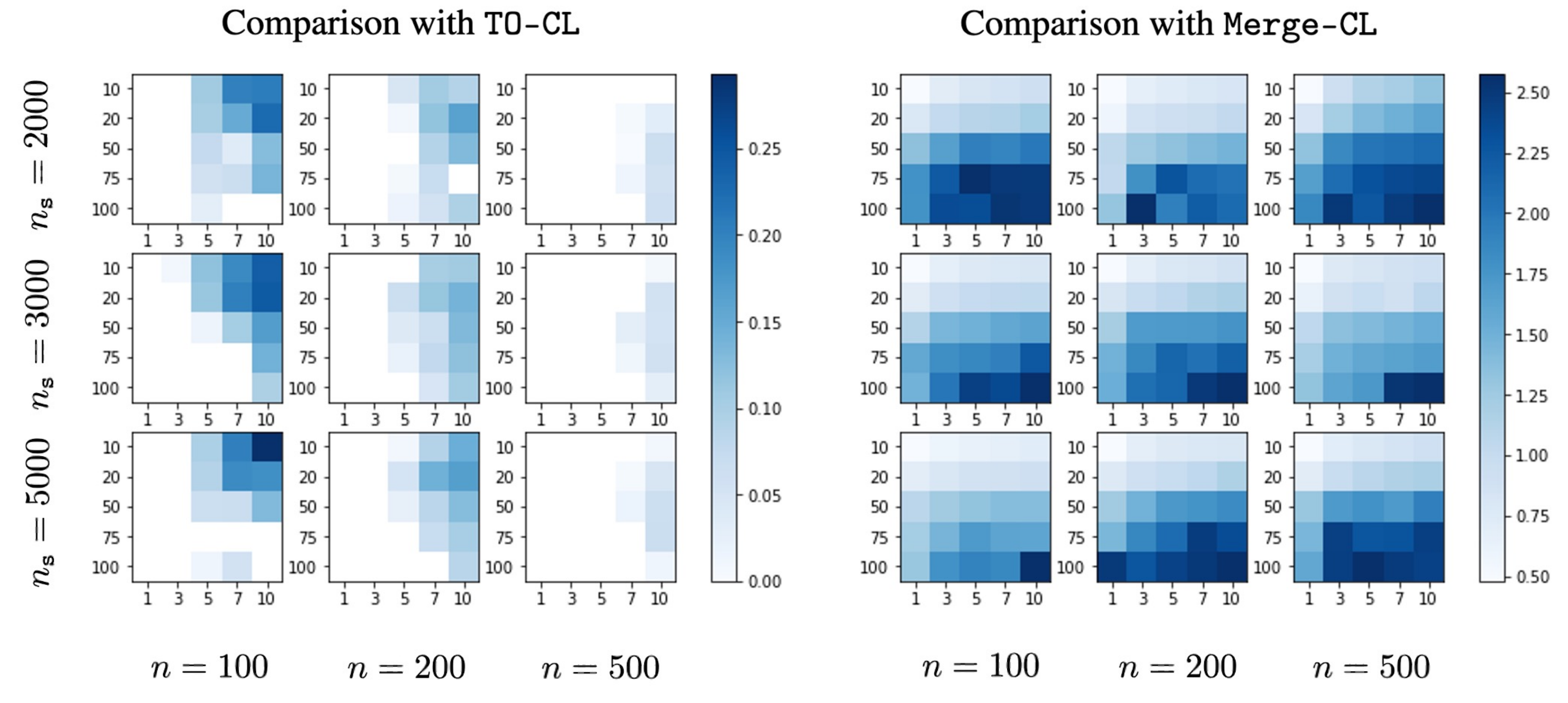}
}
%%\vspace{-0.1in}
\caption{Comparison between our proposed \texttt{$\ell_1$-TCL} with \texttt{TO-CL} (left) and \texttt{Merge-CL} (right) baseline learning frameworks. In each sub-heatmap, the x-axis represents the sparsity $s$, and the y-axis represents the dimensionality $d$. We report the difference between the average ACE estimation errors of our proposed and the baseline frameworks: positive values indicate improved accuracy, whereas negative values are all truncated to zeros for better visualization.}
\label{fig:exp_syn}
%%\vspace{-0.05in}
\end{figure}

Figure~\ref{fig:exp_syn} reports the difference between average (over independent $100$ trials) ACE estimation errors of our proposed and the baseline frameworks: positive values indicate improved accuracy whereas negative values are all truncated to zeros for better visualization. 
From the comparison with \texttt{TO-CL} (left panel), we can observe that \texttt{$\ell_1$-TCL} yields more accurate ACEs for most considered experimental settings.
Most importantly, we can observe that the benefit from our TL approach is the most significant when we have a limited amount of target domain data, and this benefit gradually disappears when we have an increasing amount of target domain data.
From the comparison with \texttt{Merge-CL} (right panel), we can observe improved accuracy for almost all settings, verifying that \texttt{Merge-CL} is inherently flawed due to the different PS models between the target and source domain. In our semi-synthetic data (or pseudo-real-data) experiment, we will not consider \texttt{Merge-CL}.

\subsection{Additional details of the real-data example}\label{appendix:add_real_result}

Similar to our synthetic data experiments above, we again consider the GLM-based PS model and use the TLIPW estimator \eqref{eq:TLIPW} in our \texttt{$\ell_1$-TCL} framework;
the hyperparameter is selected via a 5-fold CV using target domain data based on maximum average treatment classification prediction AUC.
As mentioned previously, we use vanilla logistic regression for rough estimation using source domain data, and in the $\ell_1$ regularized bias correction step, we use gradient descent to optimize the objective function \eqref{eq:re-formulation} with respect to the (sparse) difference.
In our implementation of the bias correction step, we perform a grid search over \texttt{total number of iterations} $\in \{5000, 10000, 20000\}$, \texttt{initial learning rate} $\in \{0.05,0.02,0.01,0.005,0.001\}$, \texttt{learning rate decay ratio} $\in \{0.5,0.8,0.9,0.95\}$ and \texttt{$\ell_1$ regularization strength} $\log_{10} \lambda \in \{-2.5 , -2.25, -2 ,\dots, 0\}$. The learning rate decays every $1000$ iterations.

For the bootstrap uncertainty quantification, we re-fit the model with a hyperparameter re-selected for each bootstrap sample; then, $90\%$ confidence interval is constructed based on the $5\%$ and $95\%$ quantiles of the $200$ bootstrap ACE estimates. Additionally, the mean and median of the bootstrap ACE estimates are reported.

\section{Additional Details of NN-Based Experiments}\label{appendix:add_ihdp_result}

\subsection{Description of IHDP dataset}

The IHDP dataset was collected from an observational study in which the goal was to observe the impact of visits from a healthcare provider on children's cognitive development. Patients were placed in the treatment group if they received special care or home visits from a provider. A semi-synthetic dataset was created from the original dataset by removing a nonrandom amount of the treatment group in order to induce treatment imbalance. The final cohort consists of 747 subjects, with 139 in the treatment group and 608 in the control group. Each subject contains 25 covariates, 6 of which are continuous, 19 of which are categorical (with 18 being binary and 1 having labels 0, 1, 2). These variables were collected from the child's measurements, the mother's behaviors during pregnancy, and the mother's measurements at the time of birth. The variables are detailed in Table \ref{table:ihdp_dataset}. Complete details can be found in the original papers by \citet{brooks1992effects,hill2011bayesian}, and the used dataset can be found in the \texorpdfstring{\protect\hyperlink{gitlink}{open source implementation}},.

\begin{table*}[!htp]
    %\tiny
    \centering
    \caption{List of covariates in the IHDP dataset.}
    \label{table:ihdp_dataset}
    \vspace{.1in}
    \resizebox{\textwidth}{!}{
        \begin{tabular}{ll}
         \toprule
          {Type} & {Name}   \\
          \cmidrule(l){1-2}
          \multirow{2}{*}{Child's measurements} & Birth Weight, Head Circumference, Weeks Born Preterm,  \\
          & Birth Order, First Born, Neonatal Health Index, Sex, Twin Status. \\
          \cmidrule(l){2-2}
          {Behaviors observed during pregnancy} & Smoked Cigarettes, Drank Alcohol, Took Drugs \\
          \cmidrule(l){2-2}
          \multirow{3}{*}{Mother's measurements at time of birth} & Age, Marital Status, Educational Attainment, \\
          & Worked During Pregnancy, Received Prenatal Care, \\
          & Resident Site at Start of Intervention. \\
          \bottomrule
        \end{tabular}
        }
\end{table*}

\subsection{Experimental configurations and training details}\label{appendix:NN_training_detail}
The source-target domain partition is based on the $4$-th categorical covariate, which yields $n_\mys = 546$ source domain sample (with labels 0) and $n = 201$ target domain samples (with labels 1). 
The rough estimation step in the nuisance parameter estimation stage is simply done by applying Dragonnet or TARNet fitting on the source domain data; subsequently, in the bias correction step, we randomly select $100$ target domain samples as in-sample data, which are (randomly) decomposed into $70$ training samples and $30$ validation samples, and $101$ out-of-sample testing data. 
Formally, the nuisance parameter estimation stage of, for example, Dragonnet-based \texttt{$\ell_1$-TCL} is done by:
\begin{changemargin}{0.8cm}{.5cm}
\underline{{Rough estimation for Dragonnet}}: $
\hat \theta_\mys = \arg \min_{\theta} \ \cL_{\rm Dragon}(\theta;\cD_\mys),$ 
\end{changemargin}
%\vspace{-0.13in}
\begin{changemargin}{1cm}{.5cm} 
\underline{{Bias correction for Dragonnet}}: $ \ \
\hat \theta_\myt = \hat \theta_\mys + \arg \min_{\Delta} \ \cL_{\rm Dragon}(\Delta + \hat \theta_\mys;\cD_\myt) + \lambda \norm{\Delta}_1,$
\end{changemargin}
where the objective function $\cL_{\rm Dragon}$ is defined in eq.~\eqref{eq:dragon_obj}.

The primary goal is to compare three learning frameworks: \texttt{TO-CL}, \texttt{WS-TCL}, and our proposed \texttt{$\ell_1$-TCL}. Additionally, we use the IPW, OR, and DR estimators for the downstream ACE estimation, which results in a total of 9 {\it estimation procedures}; here, we call a specific learning framework coupled with a specific ACE estimator an estimation procedure (recall that an ACE estimator is defined as a specific nuisance model coupled with a specific plug-in estimator for ACE).

For \texttt{WS-TCL} as well as our proposed \texttt{$\ell_1$-TCL} frameworks, the rough estimation step uses default NN hyperparameters in the \texorpdfstring{\protect\hyperlink{gitlink}{open source implementation}},. Notice that those hyperparameters may not be optimal for the rough estimation step using source domain data since the default hyperparameters are optimized using the full data; however, due to limited computational resources, we only consider the following hyperparameter selection based on grid-search in the bias correction step:
We consider \texttt{learning rate} $\in \{1\text{\rm e-}6,2\text{\rm e-}6,5\text{\rm e-}6,1\text{\rm e-}5,2\text{\rm e-}5,5\text{\rm e-}5,1\text{\rm e-}4,1\text{\rm e-}3\}$
and \texttt{batch size} $\in \{1,3,6,16,32,64\}$; in particular, for our proposed \texttt{$\ell_1$-TCL}, we also consider \texttt{regularization strength} $\lambda \in  \{1\text{\rm e-}6,5\text{\rm e-}6,1\text{\rm e-}5,$ $5\text{\rm e-}5,1\text{\rm e-}4,1\text{\rm e-}3,1\text{\rm e-}2,1\text{\rm e-}1,5\text{\rm e-}1,1\text{\rm e}1\}$.
For a fair comparison, we additionally optimize the hyperparameter for the baseline \texttt{TO-CL} framework by considering \texttt{learning rate} $\in \{1\text{\rm e-}6,2\text{\rm e-}6,5\text{\rm e-}6,$ $1\text{\rm e-}5,2\text{\rm e-}5,5\text{\rm e-}5,1\text{\rm e-}4,1\text{\rm e-}3\}$
and \texttt{batch size} $\in \{1,3,6,16,32,64\}$. We will show that, even with sub-optimal NN hyperparameters in the rough estimation steps, the TCL frameworks outperform the baseline \texttt{TO-CL} framework, verifying the necessity of using source domain data for accurate ACE estimation.

The hyperparameter selection criteria are NN regression loss, NN classification CE loss, and MSE. Again let us take Dragonnet as an example, slightly abuse the notation, and let $(\boldsymbol{x}_{i}, z_{i}, y_{i}), \ i = 1,\dots,n$ denote the validation target dataset, then the three aforementioned hyperparameter selection metrics are:
$\frac{1}{n} \sum_{i=1}^n \left(m_{z_i}(\theta; \boldsymbol{x}_{i}) - y_i\right)^2, \ \frac{1}{n} \sum_{i=1}^n \operatorname{CE}\left(e(\theta; \boldsymbol{x}_{i}),z_i\right)$ and $\frac{1}{n} \sum_{i=1}^n \left(e(\theta; \boldsymbol{x}_{i}) - z_i\right)^2$.

\subsection{Additional results}

We report additional results using NN classification-based criteria in Table~\ref{tab:idx4}, where the column-wise best results are highlighted with a green background color, indicating the best learning framework for the corresponding ACE estimator, and the smallest error is highlighted in bold font, indicating the overall best estimation procedure. 
Results in Table~\ref{tab:idx4} exhibit similar patterns as shown in Table~\ref{tab:idx4} that TL, especially our proposed \texttt{$\ell_1$-TCL}, generally helps improve ACE estimation accuracy, and IPW estimators do not perform well. Furthermore, as we can see from Table~\ref{tab:idx4} In the case where \texttt{WS-TCL} outperforms our proposed \texttt{$\ell_1$-TCL}, such as the out-of-sample accuracy for TARNet-DR ACE estimator in the ``NN classification CE loss'' sub-table, the improvement of \texttt{WS-TCL} is typically marginal; in contrast, when \texttt{$\ell_1$-TCL} performs the best, the improvement is significant, see, e.g., the rest TARNet-OR and TARNet-DR ACE estimators in both sub-tables of Table~\ref{tab:idx4} as well as Table~\ref{tab:idx4}.

\begin{table}[!htp]
\centering
\caption{Mean and standard deviation table using other hyperparameter selection criteria (specified on top of each sub-table) with the same source-target domain partition as in Table~\ref{tab:idx4_1000} but with fewer trials (to be precise, 50 trials) due to computing time consideration.}
\label{tab:idx4}
\vspace{.1in}
\resizebox{.85\textwidth}{!}{%
\begin{tabular}{lcccccc} 
\multicolumn{7}{c}{{Hyperparameter selected based on minimum NN regression loss on validation dataset}} \\ 
\toprule
\multirow{2}{*}{In-sample}    & \multicolumn{3}{c}{Dragonnet} & \multicolumn{3}{c}{3-Headed Variant of TARNet} \\
    & IPW       & OR      & DR      &      IPW       & OR      & DR\\
\cmidrule(l){2-4}
\cmidrule(l){5-7}
\texttt{TO-CL}  & $12.479_{(26.993)}$ & $0.868_{(1.47)}$ & $0.654_{(0.702)}$ & $6.85_{(6.192)}$ & $0.567_{(0.446)}$ & $0.468_{(0.364)}$ \\ 
\texttt{WS-TCL} & $6.414_{(9.667)}$ & \cellcolor[HTML]{C6EFCE}{\color[HTML]{006100}  $0.534_{(0.552)}$ } & \cellcolor[HTML]{C6EFCE}{\color[HTML]{006100}  $0.572_{(0.636)}$ } & $3.502_{(4.101)}$ & $0.413_{(0.313)}$ & $0.359_{(0.22)}$ \\ 
\texttt{$\ell_1$-TCL}  & \cellcolor[HTML]{C6EFCE}{\color[HTML]{006100}  $6.412_{(9.664)}$ } & $0.543_{(0.557)}$ & $0.58_{(0.634)}$ & \cellcolor[HTML]{C6EFCE}{\color[HTML]{006100}  $3.326_{(3.626)}$ } & \cellcolor[HTML]{C6EFCE}{\color[HTML]{006100}  $0.36_{(0.312)}$ } & \cellcolor[HTML]{C6EFCE}{\color[HTML]{006100}  $\boldsymbol{0.293}_{(0.222)}$ } \\ 
\cmidrule(l){1-7}
  \multirow{2}{*}{Out-of-sample}    & \multicolumn{3}{c}{Dragonnet} & \multicolumn{3}{c}{3-Headed Variant of TARNet} \\
    & IPW       & OR      & DR      &      IPW       & OR      & DR\\
\cmidrule(l){2-4}
\cmidrule(l){5-7}
\texttt{TO-CL}  & $35.352_{(75.125)}$ & $0.826_{(1.248)}$ & $2.009_{(3.397)}$ & $5.664_{(6.884)}$ & $0.671_{(0.56)}$ & $0.367_{(0.318)}$ \\ 
\texttt{WS-TCL} & $17.684_{(20.993)}$ & \cellcolor[HTML]{C6EFCE}{\color[HTML]{006100}  $0.512_{(0.586)}$ } & \cellcolor[HTML]{C6EFCE}{\color[HTML]{006100}  $1.324_{(1.492)}$ } & $4.204_{(6.092)}$ & $0.476_{(0.399)}$ & $0.339_{(0.289)}$ \\ 
\texttt{$\ell_1$-TCL}  & \cellcolor[HTML]{C6EFCE}{\color[HTML]{006100}  $17.682_{(20.996)}$ } & $0.519_{(0.615)}$ & $1.337_{(1.494)}$ & \cellcolor[HTML]{C6EFCE}{\color[HTML]{006100}  $4.039_{(4.762)}$ } & \cellcolor[HTML]{C6EFCE}{\color[HTML]{006100}  $0.418_{(0.353)}$ } & \cellcolor[HTML]{C6EFCE}{\color[HTML]{006100}  $\boldsymbol{0.308}_{(0.251)}$ } \\ 
\bottomrule
\rule{0pt}{0.5ex} \\ 
\multicolumn{7}{c}{{Hyperparameter selected based on minimum validation NN classification CE loss}} \\ 
\toprule
\multirow{2}{*}{In-sample}    & \multicolumn{3}{c}{Dragonnet} & \multicolumn{3}{c}{3-Headed Variant of TARNet} \\
    & IPW       & OR      & DR      &      IPW       & OR      & DR\\
\cmidrule(l){2-4}
\cmidrule(l){5-7}
\texttt{TO-CL}  & $11.411_{(30.657)}$ & \cellcolor[HTML]{C6EFCE}{\color[HTML]{006100}  $0.615_{(0.773)}$ } & \cellcolor[HTML]{C6EFCE}{\color[HTML]{006100}  $0.675_{(0.649)}$ } & $3.926_{(4.234)}$ & $0.768_{(0.545)}$ & $0.464_{(0.365)}$ \\ 
\texttt{WS-TCL} & \cellcolor[HTML]{C6EFCE}{\color[HTML]{006100}  $10.455_{(18.907)}$ } & $0.682_{(0.75)}$ & $0.851_{(0.948)}$ & \cellcolor[HTML]{C6EFCE}{\color[HTML]{006100}  $3.418_{(3.428)}$ } & $0.469_{(0.334)}$ & $0.383_{(0.262)}$ \\ 
\texttt{$\ell_1$-TCL}  & $11.781_{(27.023)}$ & $0.813_{(1.123)}$ & $0.907_{(1.135)}$ & $4.012_{(6.241)}$ & \cellcolor[HTML]{C6EFCE}{\color[HTML]{006100}  $0.328_{(0.245)}$ } & \cellcolor[HTML]{C6EFCE}{\color[HTML]{006100}  $\boldsymbol{0.326}_{(0.334)}$ } \\ 
\cmidrule(l){1-7}
  \multirow{2}{*}{Out-of-sample}    & \multicolumn{3}{c}{Dragonnet} & \multicolumn{3}{c}{3-Headed Variant of TARNet} \\
    & IPW       & OR      & DR      &      IPW       & OR      & DR\\
\cmidrule(l){2-4}
\cmidrule(l){5-7}
\texttt{TO-CL}  & $32.228_{(45.735)}$ & \cellcolor[HTML]{C6EFCE}{\color[HTML]{006100}  $0.629_{(0.548)}$ } & \cellcolor[HTML]{C6EFCE}{\color[HTML]{006100}  $1.513_{(1.627)}$ } & \cellcolor[HTML]{C6EFCE}{\color[HTML]{006100}  $4.144_{(5.009)}$ } & $0.806_{(0.641)}$ & $0.433_{(0.482)}$ \\ 
\texttt{WS-TCL} & \cellcolor[HTML]{C6EFCE}{\color[HTML]{006100}  $29.121_{(34.475)}$ } & $0.629_{(0.816)}$ & $2.535_{(4.217)}$ & $4.371_{(5.18)}$ & $0.529_{(0.484)}$ & \cellcolor[HTML]{C6EFCE}{\color[HTML]{006100}  $\boldsymbol{0.349}_{(0.339)}$ } \\ 
\texttt{$\ell_1$-TCL}  & $33.303_{(54.411)}$ & $0.746_{(0.951)}$ & $2.639_{(3.806)}$ & $5.779_{(9.668)}$ & \cellcolor[HTML]{C6EFCE}{\color[HTML]{006100}  $0.426_{(0.33)}$ } & $0.35_{(0.429)}$ \\ 
\bottomrule
\rule{0pt}{0.5ex} \\ 
\multicolumn{7}{c}{{Hyperparameter selected based on minimum validation NN classification MSE}} \\ 
\toprule
\multirow{2}{*}{In-sample}    & \multicolumn{3}{c}{Dragonnet} & \multicolumn{3}{c}{3-Headed Variant of TARNet} \\
    & IPW       & OR      & DR      &      IPW       & OR      & DR\\
\cmidrule(l){2-4}
\cmidrule(l){5-7}
\texttt{TO-CL}  & $10.116_{(24.513)}$ & $0.767_{(1.077)}$ & $1.033_{(1.075)}$ & $4.647_{(5.418)}$ & $0.607_{(0.539)}$ & $0.474_{(0.322)}$ \\ 
\texttt{WS-TCL} & $8.342_{(13.545)}$ & $0.664_{(0.868)}$ & $0.719_{(0.789)}$ & \cellcolor[HTML]{C6EFCE}{\color[HTML]{006100}  $3.77_{(5.429)}$ } & $0.589_{(0.431)}$ & $0.408_{(0.296)}$ \\ 
\texttt{$\ell_1$-TCL}  & \cellcolor[HTML]{C6EFCE}{\color[HTML]{006100}  $8.249_{(13.454)}$ } & \cellcolor[HTML]{C6EFCE}{\color[HTML]{006100}  $0.623_{(0.848)}$ } & \cellcolor[HTML]{C6EFCE}{\color[HTML]{006100}  $0.704_{(0.759)}$ } & $4.012_{(6.241)}$ & \cellcolor[HTML]{C6EFCE}{\color[HTML]{006100}  $0.328_{(0.245)}$ } & \cellcolor[HTML]{C6EFCE}{\color[HTML]{006100}  $\boldsymbol{0.326}_{(0.334)}$ } \\ 
\cmidrule(l){1-7}
  \multirow{2}{*}{Out-of-sample}    & \multicolumn{3}{c}{Dragonnet} & \multicolumn{3}{c}{3-Headed Variant of TARNet} \\
    & IPW       & OR      & DR      &      IPW       & OR      & DR\\
\cmidrule(l){2-4}
\cmidrule(l){5-7}
\texttt{TO-CL}  & $31.608_{(55.768)}$ & $0.743_{(0.892)}$ & $2.261_{(3.283)}$ & \cellcolor[HTML]{C6EFCE}{\color[HTML]{006100}  $4.764_{(6.154)}$ } & $0.739_{(0.715)}$ & $0.383_{(0.384)}$ \\ 
\texttt{WS-TCL} & $26.398_{(47.011)}$ & $0.663_{(1.014)}$ & $1.711_{(2.249)}$ & $5.437_{(8.29)}$ & $0.73_{(0.698)}$ & $0.451_{(0.569)}$ \\ 
\texttt{$\ell_1$-TCL}  & \cellcolor[HTML]{C6EFCE}{\color[HTML]{006100}  $25.786_{(46.738)}$ } & \cellcolor[HTML]{C6EFCE}{\color[HTML]{006100}  $0.661_{(0.947)}$ } & \cellcolor[HTML]{C6EFCE}{\color[HTML]{006100}  $1.666_{(2.185)}$ } & $5.779_{(9.668)}$ & \cellcolor[HTML]{C6EFCE}{\color[HTML]{006100}  $0.426_{(0.33)}$ } & \cellcolor[HTML]{C6EFCE}{\color[HTML]{006100}  $\boldsymbol{0.35}_{(0.429)}$ } \\ 
\bottomrule
\rule{0pt}{0.5ex} \\ \end{tabular}%
}
\end{table}

In addition, we consider another (randomly selected) binary covariate for the source-target domain partition, which is the $8$-th categorical covariate and yields $n_\mys = 642$ source domain sample (with labels 0) and $n = 105$ target domain samples (with labels 1). The (random) train-validation split gives 73 training samples (that is why we choose the largest batch size grid to be 64) and 32 validation samples. We repeat the same experiments and report the results in Table~\ref{tab:idx8}.

\begin{table}[!htp]
\centering
\caption{Additional absolute error mean and standard deviation table for all aforementioned hyperparameter selection criteria (specified on top of each sub-table) using the different (from that of Table~\ref{tab:idx4}) source-target domain partition.}
\label{tab:idx8}
\vspace{.1in}
\resizebox{.85\textwidth}{!}{%
\begin{tabular}{lcccccc} 
\multicolumn{7}{c}{{Hyperparameter selected based on minimum validation NN regression loss}} \\ 
\toprule
\multirow{2}{*}{In-sample}    & \multicolumn{3}{c}{Dragonnet} & \multicolumn{3}{c}{3-Headed Variant of TARNet} \\
    & IPW       & OR      & DR      &      IPW       & OR      & DR\\
\cmidrule(l){2-4}
\cmidrule(l){5-7}
\texttt{TO-CL}  & $5.752_{(14.189)}$ & $0.572_{(0.483)}$ & $0.592_{(0.775)}$ & $5.713_{(9.2)}$ & $0.466_{(0.406)}$ & $0.457_{(0.434)}$ \\ 
\texttt{WS-TCL} & $6.448_{(10.771)}$ & $0.572_{(0.463)}$ & $0.675_{(1.08)}$ & $1.703_{(2.018)}$ & \cellcolor[HTML]{C6EFCE}{\color[HTML]{006100}  $0.44_{(0.362)}$ } & \cellcolor[HTML]{C6EFCE}{\color[HTML]{006100}  $\boldsymbol{0.375}_{(0.291)}$ } \\ 
\texttt{$\ell_1$-TCL}  & \cellcolor[HTML]{C6EFCE}{\color[HTML]{006100}  $5.376_{(9.682)}$ } & \cellcolor[HTML]{C6EFCE}{\color[HTML]{006100}  $0.477_{(0.526)}$ } & \cellcolor[HTML]{C6EFCE}{\color[HTML]{006100}  $0.526_{(0.884)}$ } & \cellcolor[HTML]{C6EFCE}{\color[HTML]{006100}  $1.697_{(2.017)}$ } & $0.448_{(0.362)}$ & $0.376_{(0.292)}$ \\ 
\cmidrule(l){1-7}
  \multirow{2}{*}{Out-of-sample}    & \multicolumn{3}{c}{Dragonnet} & \multicolumn{3}{c}{3-Headed Variant of TARNet} \\
    & IPW       & OR      & DR      &      IPW       & OR      & DR\\
\cmidrule(l){2-4}
\cmidrule(l){5-7}
\texttt{TO-CL}  & \cellcolor[HTML]{C6EFCE}{\color[HTML]{006100}  $16.704_{(24.684)}$ } & $0.89_{(1.05)}$ & $1.927_{(3.917)}$ & $9.916_{(12.285)}$ & $0.736_{(0.811)}$ & \cellcolor[HTML]{C6EFCE}{\color[HTML]{006100}  $0.685_{(0.851)}$ } \\ 
\texttt{WS-TCL} & $21.43_{(22.895)}$ & $0.821_{(1.105)}$ & $2.252_{(2.92)}$ & \cellcolor[HTML]{C6EFCE}{\color[HTML]{006100}  $7.699_{(12.458)}$ } & \cellcolor[HTML]{C6EFCE}{\color[HTML]{006100}  $\boldsymbol{0.623}_{(0.908)}$ } & $0.723_{(1.089)}$ \\ 
\texttt{$\ell_1$-TCL}  & $18.03_{(21.948)}$ & \cellcolor[HTML]{C6EFCE}{\color[HTML]{006100}  $0.731_{(1.318)}$ } & \cellcolor[HTML]{C6EFCE}{\color[HTML]{006100}  $1.661_{(2.259)}$ } & $7.703_{(12.46)}$ & $0.627_{(0.906)}$ & $0.724_{(1.084)}$ \\ 
\bottomrule
\rule{0pt}{0.5ex} \\ 
\multicolumn{7}{c}{{Hyperparameter selected based on minimum validation NN classification CE loss}} \\ 
\toprule
\multirow{2}{*}{In-sample}    & \multicolumn{3}{c}{Dragonnet} & \multicolumn{3}{c}{3-Headed Variant of TARNet} \\
    & IPW       & OR      & DR      &      IPW       & OR      & DR\\
\cmidrule(l){2-4}
\cmidrule(l){5-7}
\texttt{TO-CL}  & \cellcolor[HTML]{C6EFCE}{\color[HTML]{006100}  $5.752_{(14.189)}$ } & \cellcolor[HTML]{C6EFCE}{\color[HTML]{006100}  $0.572_{(0.483)}$ } & \cellcolor[HTML]{C6EFCE}{\color[HTML]{006100}  $0.592_{(0.775)}$ } & $6.302_{(12.42)}$ & $0.516_{(0.47)}$ & $0.496_{(0.703)}$ \\ 
\texttt{WS-TCL} & $6.448_{(10.771)}$ & $0.572_{(0.463)}$ & $0.675_{(1.08)}$ & \cellcolor[HTML]{C6EFCE}{\color[HTML]{006100}  $1.703_{(2.018)}$ } & \cellcolor[HTML]{C6EFCE}{\color[HTML]{006100}  $0.44_{(0.362)}$ } & $0.375_{(0.291)}$ \\ 
\texttt{$\ell_1$-TCL}  & $6.442_{(10.761)}$ & $0.577_{(0.466)}$ & $0.677_{(1.139)}$ & $2.056_{(2.331)}$ & $0.446_{(0.449)}$ & \cellcolor[HTML]{C6EFCE}{\color[HTML]{006100}  $\boldsymbol{0.352}_{(0.314)}$ } \\ 
\cmidrule(l){1-7}
  \multirow{2}{*}{Out-of-sample}    & \multicolumn{3}{c}{Dragonnet} & \multicolumn{3}{c}{3-Headed Variant of TARNet} \\
    & IPW       & OR      & DR      &      IPW       & OR      & DR\\
\cmidrule(l){2-4}
\cmidrule(l){5-7}
\texttt{TO-CL}  & \cellcolor[HTML]{C6EFCE}{\color[HTML]{006100}  $16.704_{(24.684)}$ } & $0.89_{(1.05)}$ & \cellcolor[HTML]{C6EFCE}{\color[HTML]{006100}  $1.927_{(3.917)}$ } & $10.353_{(15.414)}$ & $0.878_{(1.185)}$ & $0.782_{(1.099)}$ \\ 
\texttt{WS-TCL} & $21.43_{(22.895)}$ & \cellcolor[HTML]{C6EFCE}{\color[HTML]{006100}  $0.821_{(1.105)}$ } & $2.252_{(2.92)}$ & \cellcolor[HTML]{C6EFCE}{\color[HTML]{006100}  $7.699_{(12.458)}$ } & \cellcolor[HTML]{C6EFCE}{\color[HTML]{006100}  $\boldsymbol{0.623}_{(0.908)}$ } & $0.723_{(1.089)}$ \\ 
\texttt{$\ell_1$-TCL}  & $21.41_{(22.875)}$ & $0.822_{(1.114)}$ & $2.272_{(2.948)}$ & $8.048_{(13.393)}$ & $0.632_{(0.809)}$ & \cellcolor[HTML]{C6EFCE}{\color[HTML]{006100}  $0.696_{(0.917)}$ } \\ 
\bottomrule
\rule{0pt}{0.5ex} \\ 
\multicolumn{7}{c}{{Hyperparameter selected based on minimum validation NN classification MSE}} \\ 
\toprule
\multirow{2}{*}{In-sample}    & \multicolumn{3}{c}{Dragonnet} & \multicolumn{3}{c}{3-Headed Variant of TARNet} \\
    & IPW       & OR      & DR      &      IPW       & OR      & DR\\
\cmidrule(l){2-4}
\cmidrule(l){5-7}
\texttt{TO-CL}  & \cellcolor[HTML]{C6EFCE}{\color[HTML]{006100}  $5.752_{(14.189)}$ } & \cellcolor[HTML]{C6EFCE}{\color[HTML]{006100}  $0.572_{(0.483)}$ } & \cellcolor[HTML]{C6EFCE}{\color[HTML]{006100}  $0.592_{(0.775)}$ } & $6.302_{(12.42)}$ & $0.516_{(0.47)}$ & $0.496_{(0.703)}$ \\ 
\texttt{WS-TCL} & $6.448_{(10.771)}$ & $0.572_{(0.463)}$ & $0.675_{(1.08)}$ & \cellcolor[HTML]{C6EFCE}{\color[HTML]{006100}  $1.703_{(2.018)}$ } & \cellcolor[HTML]{C6EFCE}{\color[HTML]{006100}  $0.44_{(0.362)}$ } & $0.375_{(0.291)}$ \\ 
\texttt{$\ell_1$-TCL}  & $6.441_{(10.761)}$ & $0.576_{(0.465)}$ & $0.677_{(1.134)}$ & $2.056_{(2.331)}$ & $0.446_{(0.449)}$ & \cellcolor[HTML]{C6EFCE}{\color[HTML]{006100}  $\boldsymbol{0.352}_{(0.314)}$ } \\ 
\cmidrule(l){1-7}
  \multirow{2}{*}{Out-of-sample}    & \multicolumn{3}{c}{Dragonnet} & \multicolumn{3}{c}{3-Headed Variant of TARNet} \\
    & IPW       & OR      & DR      &      IPW       & OR      & DR\\
\cmidrule(l){2-4}
\cmidrule(l){5-7}
\texttt{TO-CL}  & \cellcolor[HTML]{C6EFCE}{\color[HTML]{006100}  $16.704_{(24.684)}$ } & $0.89_{(1.05)}$ & \cellcolor[HTML]{C6EFCE}{\color[HTML]{006100}  $1.927_{(3.917)}$ } & $10.353_{(15.414)}$ & $0.878_{(1.185)}$ & $0.782_{(1.099)}$ \\ 
\texttt{WS-TCL} & $21.43_{(22.895)}$ & \cellcolor[HTML]{C6EFCE}{\color[HTML]{006100}  $0.821_{(1.105)}$ } & $2.252_{(2.92)}$ & \cellcolor[HTML]{C6EFCE}{\color[HTML]{006100}  $7.699_{(12.458)}$ } & \cellcolor[HTML]{C6EFCE}{\color[HTML]{006100}  $\boldsymbol{0.623}_{(0.908)}$ } & $0.723_{(1.089)}$ \\ 
\texttt{$\ell_1$-TCL}  & $21.408_{(22.875)}$ & $0.822_{(1.112)}$ & $2.271_{(2.946)}$ & $8.048_{(13.393)}$ & $0.632_{(0.809)}$ & \cellcolor[HTML]{C6EFCE}{\color[HTML]{006100}  $0.696_{(0.917)}$ } \\ 
\bottomrule
\rule{0pt}{0.5ex} \\ \end{tabular}%
}
\end{table}

In Table~\ref{tab:idx8}, we can observe similar patterns as mentioned before: The best in-sample performance is $0.352$, which is given by TARNet-DR estimator using proposed \texttt{$\ell_1$-TCL}, and it is much better than the best \texttt{WS-TCL} in-sample performance, which is $0.375$ also given by TARNet-DR estimator. In contrast, even though \texttt{WS-TCL} yields the best out-of-sample performance via TARNet-OR estimator ($0.623$), there is only a marginal increment compared to that of \texttt{$\ell_1$-TCL} ($0.627$). Additionally, the best out-of-sample performance of \texttt{$\ell_1$-TCL} is still given using the NN regression loss as the hyperparameter selection criterion, which is consistent with previous findings.

\begin{table}[!htp]
\centering
\caption{Median and interquartile range ([Q1, Q3], where Q1 and Q3 stand for the $25\%$ and $75\%$ quantiles) of absolute errors of estimated causal effects ovee 50 IHDP datasets for all aforementioned hyperparameter selection criteria (specified on top of each sub-table). The source-target domain partition is the same with that of Table~\ref{tab:idx8}.}
\label{tab:idx8_median}
\vspace{.1in}
\resizebox{\textwidth}{!}{%
\begin{tabular}{lcccccc} 
\multicolumn{7}{c}{{Hyperparameter selected based on minimum validation NN regression loss}} \\ 
\toprule
\multirow{2}{*}{In-sample}    & \multicolumn{3}{c}{Dragonnet} & \multicolumn{3}{c}{3-Headed Variant of TARNet} \\
    & IPW       & OR      & DR      &      IPW       & OR      & DR\\
\cmidrule(l){2-4}
\cmidrule(l){5-7}
\texttt{TO-CL}  & $2.46_{[1.175,5.219]}$ & $0.481_{[0.244,0.657]}$ & $0.303_{[0.185,0.676]}$ & $3.097_{[2.196,5.591]}$ & $0.383_{[0.149,0.611]}$ & $0.345_{[0.163,0.617]}$ \\ 
\texttt{WS-TCL} & $3.212_{[1.411,6.828]}$ & $0.46_{[0.291,0.717]}$ & $0.433_{[0.192,0.731]}$ & \cellcolor[HTML]{C6EFCE}{\color[HTML]{006100}  $1.086_{[0.512,1.851]}$ } & \cellcolor[HTML]{C6EFCE}{\color[HTML]{006100}  $0.338_{[0.165,0.662]}$ } & $0.289_{[0.141,0.547]}$ \\ 
\texttt{$\ell_1$-TCL}  & \cellcolor[HTML]{C6EFCE}{\color[HTML]{006100}  $2.048_{[1.251,5.817]}$ } & \cellcolor[HTML]{C6EFCE}{\color[HTML]{006100}  $0.339_{[0.155,0.589]}$ } & \cellcolor[HTML]{C6EFCE}{\color[HTML]{006100}  $\boldsymbol{0.288}_{[0.112,0.593]}$ } & $1.086_{[0.511,1.784]}$ & $0.353_{[0.159,0.659]}$ & \cellcolor[HTML]{C6EFCE}{\color[HTML]{006100}  $0.289_{[0.142,0.542]}$ } \\ 
\cmidrule(l){1-7}
  \multirow{2}{*}{Out-of-sample}    & \multicolumn{3}{c}{Dragonnet} & \multicolumn{3}{c}{3-Headed Variant of TARNet} \\
    & IPW       & OR      & DR      &      IPW       & OR      & DR\\
\cmidrule(l){2-4}
\cmidrule(l){5-7}
\texttt{TO-CL}  & \cellcolor[HTML]{C6EFCE}{\color[HTML]{006100}  $9.297_{[3.413,14.861]}$ } & $0.587_{[0.296,0.979]}$ & \cellcolor[HTML]{C6EFCE}{\color[HTML]{006100}  $0.835_{[0.277,1.876]}$ } & $5.068_{[2.957,12.122]}$ & $0.451_{[0.22,0.87]}$ & $0.452_{[0.199,0.825]}$ \\ 
\texttt{WS-TCL} & $15.119_{[4.856,27.383]}$ & $0.513_{[0.286,0.901]}$ & $0.979_{[0.547,2.246]}$ & \cellcolor[HTML]{C6EFCE}{\color[HTML]{006100}  $3.636_{[0.973,10.543]}$ } & \cellcolor[HTML]{C6EFCE}{\color[HTML]{006100}  $0.425_{[0.121,0.764]}$ } & $0.399_{[0.157,0.897]}$ \\ 
\texttt{$\ell_1$-TCL}  & $12.953_{[5.219,19.517]}$ & \cellcolor[HTML]{C6EFCE}{\color[HTML]{006100}  $\boldsymbol{0.368}_{[0.183,0.737]}$ } & $0.862_{[0.457,1.982]}$ & $3.636_{[0.97,10.543]}$ & $0.426_{[0.127,0.762]}$ & \cellcolor[HTML]{C6EFCE}{\color[HTML]{006100}  $0.395_{[0.163,0.958]}$ } \\ 
\bottomrule
\rule{0pt}{0.5ex} \\ 
\multicolumn{7}{c}{{Hyperparameter selected based on minimum validation NN classification CE loss}} \\ 
\toprule
\multirow{2}{*}{In-sample}    & \multicolumn{3}{c}{Dragonnet} & \multicolumn{3}{c}{3-Headed Variant of TARNet} \\
    & IPW       & OR      & DR      &      IPW       & OR      & DR\\
\cmidrule(l){2-4}
\cmidrule(l){5-7}
\texttt{TO-CL}  & \cellcolor[HTML]{C6EFCE}{\color[HTML]{006100}  $2.46_{[1.175,5.219]}$ } & $0.481_{[0.244,0.657]}$ & \cellcolor[HTML]{C6EFCE}{\color[HTML]{006100}  $0.303_{[0.185,0.676]}$ } & $2.699_{[2.089,4.958]}$ & $0.359_{[0.245,0.586]}$ & $0.315_{[0.162,0.58]}$ \\ 
\texttt{WS-TCL} & $3.212_{[1.411,6.828]}$ & \cellcolor[HTML]{C6EFCE}{\color[HTML]{006100}  $0.46_{[0.291,0.717]}$ } & $0.433_{[0.192,0.731]}$ & \cellcolor[HTML]{C6EFCE}{\color[HTML]{006100}  $1.086_{[0.512,1.851]}$ } & $0.338_{[0.165,0.662]}$ & $0.289_{[0.141,0.547]}$ \\ 
\texttt{$\ell_1$-TCL}  & $3.215_{[1.406,6.861]}$ & $0.461_{[0.298,0.719]}$ & $0.43_{[0.195,0.702]}$ & $1.187_{[0.705,2.253]}$ & \cellcolor[HTML]{C6EFCE}{\color[HTML]{006100}  $0.309_{[0.196,0.55]}$ } & \cellcolor[HTML]{C6EFCE}{\color[HTML]{006100}  $\boldsymbol{0.27}_{[0.096,0.477]}$ } \\ 
\cmidrule(l){1-7}
  \multirow{2}{*}{Out-of-sample}    & \multicolumn{3}{c}{Dragonnet} & \multicolumn{3}{c}{3-Headed Variant of TARNet} \\
    & IPW       & OR      & DR      &      IPW       & OR      & DR\\
\cmidrule(l){2-4}
\cmidrule(l){5-7}
\texttt{TO-CL}  & \cellcolor[HTML]{C6EFCE}{\color[HTML]{006100}  $9.297_{[3.413,14.861]}$ } & $0.587_{[0.296,0.979]}$ & \cellcolor[HTML]{C6EFCE}{\color[HTML]{006100}  $0.835_{[0.277,1.876]}$ } & $5.103_{[2.116,11.363]}$ & $0.541_{[0.362,0.813]}$ & $0.51_{[0.258,0.719]}$ \\ 
\texttt{WS-TCL} & $15.119_{[4.856,27.383]}$ & \cellcolor[HTML]{C6EFCE}{\color[HTML]{006100}  $0.513_{[0.286,0.901]}$ } & $0.979_{[0.547,2.246]}$ & \cellcolor[HTML]{C6EFCE}{\color[HTML]{006100}  $3.636_{[0.973,10.543]}$ } & $0.425_{[0.121,0.764]}$ & \cellcolor[HTML]{C6EFCE}{\color[HTML]{006100}  $0.399_{[0.157,0.897]}$ } \\ 
\texttt{$\ell_1$-TCL}  & $15.117_{[4.856,27.358]}$ & $0.514_{[0.266,0.891]}$ & $0.976_{[0.593,2.275]}$ & $3.781_{[0.832,11.022]}$ & \cellcolor[HTML]{C6EFCE}{\color[HTML]{006100}  $\boldsymbol{0.361}_{[0.193,0.76]}$ } & $0.429_{[0.201,0.748]}$ \\ 
\bottomrule
\rule{0pt}{0.5ex} \\ 
\multicolumn{7}{c}{{Hyperparameter selected based on minimum validation NN classification MSE}} \\ 
\toprule
\multirow{2}{*}{In-sample}    & \multicolumn{3}{c}{Dragonnet} & \multicolumn{3}{c}{3-Headed Variant of TARNet} \\
    & IPW       & OR      & DR      &      IPW       & OR      & DR\\
\cmidrule(l){2-4}
\cmidrule(l){5-7}
\texttt{TO-CL}  & \cellcolor[HTML]{C6EFCE}{\color[HTML]{006100}  $2.46_{[1.175,5.219]}$ } & $0.481_{[0.244,0.657]}$ & \cellcolor[HTML]{C6EFCE}{\color[HTML]{006100}  $0.303_{[0.185,0.676]}$ } & $2.699_{[2.089,4.958]}$ & $0.359_{[0.245,0.586]}$ & $0.315_{[0.162,0.58]}$ \\ 
\texttt{WS-TCL} & $3.212_{[1.411,6.828]}$ & \cellcolor[HTML]{C6EFCE}{\color[HTML]{006100}  $0.46_{[0.291,0.717]}$ } & $0.433_{[0.192,0.731]}$ & \cellcolor[HTML]{C6EFCE}{\color[HTML]{006100}  $1.086_{[0.512,1.851]}$ } & $0.338_{[0.165,0.662]}$ & $0.289_{[0.141,0.547]}$ \\ 
\texttt{$\ell_1$-TCL}  & $3.215_{[1.405,6.859]}$ & $0.461_{[0.299,0.719]}$ & $0.43_{[0.194,0.702]}$ & $1.187_{[0.705,2.253]}$ & \cellcolor[HTML]{C6EFCE}{\color[HTML]{006100}  $0.309_{[0.196,0.55]}$ } & \cellcolor[HTML]{C6EFCE}{\color[HTML]{006100}  $\boldsymbol{0.27}_{[0.096,0.477]}$ } \\ 
\cmidrule(l){1-7}
  \multirow{2}{*}{Out-of-sample}    & \multicolumn{3}{c}{Dragonnet} & \multicolumn{3}{c}{3-Headed Variant of TARNet} \\
    & IPW       & OR      & DR      &      IPW       & OR      & DR\\
\cmidrule(l){2-4}
\cmidrule(l){5-7}
\texttt{TO-CL}  & \cellcolor[HTML]{C6EFCE}{\color[HTML]{006100}  $9.297_{[3.413,14.861]}$ } & $0.587_{[0.296,0.979]}$ & \cellcolor[HTML]{C6EFCE}{\color[HTML]{006100}  $0.835_{[0.277,1.876]}$ } & $5.103_{[2.116,11.363]}$ & $0.541_{[0.362,0.813]}$ & $0.51_{[0.258,0.719]}$ \\ 
\texttt{WS-TCL} & $15.119_{[4.856,27.383]}$ & \cellcolor[HTML]{C6EFCE}{\color[HTML]{006100}  $0.513_{[0.286,0.901]}$ } & $0.979_{[0.547,2.246]}$ & \cellcolor[HTML]{C6EFCE}{\color[HTML]{006100}  $3.636_{[0.973,10.543]}$ } & $0.425_{[0.121,0.764]}$ & \cellcolor[HTML]{C6EFCE}{\color[HTML]{006100}  $0.399_{[0.157,0.897]}$ } \\ 
\texttt{$\ell_1$-TCL}  & $15.116_{[4.852,27.351]}$ & $0.514_{[0.265,0.892]}$ & $0.974_{[0.593,2.275]}$ & $3.781_{[0.832,11.022]}$ & \cellcolor[HTML]{C6EFCE}{\color[HTML]{006100}  $\boldsymbol{0.361}_{[0.193,0.76]}$ } & $0.429_{[0.201,0.748]}$ \\ 
\bottomrule
\rule{0pt}{0.5ex} \\ \end{tabular}%
}
\end{table}
%\vspace*{\fill}

Lastly, since both \texttt{WS-TCL} and \texttt{$\ell_1$-TCL} are sensitive to the choice of hyperparameters, there will be cases/trials where the optimal empirical option of not covered by the pre-defined grid, leading to unfavorable results. Since our \texttt{$\ell_1$-TCL} has one additional hyperparameter, i.e., the regularization strength, such effect will be amplified for \texttt{$\ell_1$-TCL} given the limited computation resources when we perform grid search for hyperparameter selections. Therefore, we report the the median and interquartile range of the aforementioned experiments using $8$-th categorical covariate for source-target domain partition in Table~\ref{tab:idx8_median}, which shows that both the in-sample and out-of-sample best (in terms of median) estimates are given by our \texttt{$\ell_1$-TCL}. Overall all results above support the effectiveness of our proposed \texttt{$\ell_1$-TCL} framework.

\subsection{Definition of standard mean difference}\label{appendix:SMD}

\begin{definition}[Cohen's \texttt{d}]
    Given two sets of samples $\cA = \{a_{i}, \ i = 1,\dots,m\}$ and $\cB = \{b_{j}, \ j = 1,\dots,n\}$, the Cohen's \texttt{d} is defined as follows: 
\begin{equation*}%\label{eq:SMD}
    \texttt{d}_{\rm Cohen}\big(\cA,\cB\big) =\left(\bar{a}-\bar{b}\right) \Bigg/ {\sqrt{\frac{S_{\cA}^2+S_{\cB}^2}{2}}},
\end{equation*}
where \(\bar{a} = \sum_{i=1}^m a_i/m, \ \bar{b} = \sum_{i=1}^n b_j/n\), and the pooled sample variance can be calculated as:
\[S_{\cA} = \frac{1}{m} \sum_{i=1}^m \left(a_i - \bar{a}\right)^2, \quad S_{\cB} = \frac{1}{n} \sum_{j=1}^n \left(b_j - \bar{b}\right)^2.\]
\end{definition}

When the Cohen's \texttt{d}'s absolute value is close to zero, the standardized means of two distributions are similar to each other, i.e., the covariates' distributions are balanced.
In our multivariate setting, we can simply use the average of the absolute Cohen's \texttt{d}'s as the selection criterion, i.e.,
\[\text{SMD} = \frac{1}{d}\sum_{j=1}^{d} \Big|\texttt{d}_{\rm Cohen}\big(\{\boldsymbol{x}(j): \boldsymbol{x} \in \cD_{\boldsymbol{x},\rm trt}\}, \{\boldsymbol{x}(j): \boldsymbol{x} \in \cD_{\boldsymbol{x},\rm ctrl}\}\big)\Big|, \]
where $\boldsymbol{x}(j)$ is the $j$-th element of the vector $\boldsymbol{x} \in \RR^d$, and
\[\cD_{\boldsymbol{x},\rm trt} = \left\{ \frac{\boldsymbol{x}_{i}}{\hat e(\boldsymbol{x}_{i})}: {z}_{i} = 1\right\},\quad \cD_{\boldsymbol{x},\rm ctrl} = \left\{ \frac{\boldsymbol{x}_{i}}{1-\hat e(\boldsymbol{x}_{i})}: {z}_{i} = 0\right\}.\]
This above metric tells us the average standardized mean difference after propensity score weighting.

\end{document}